\documentclass[acmtog, authorversion, nonacm]{acmart}
  \renewcommand\footnotetextcopyrightpermission[1]{}
\AtBeginDocument{%
  }

\usepackage{makecell}
\usepackage{tabularray}
\usepackage{tabularx}
\usepackage{colortbl}
\usepackage{multirow}
\usepackage[export]{adjustbox}
\usepackage{subcaption}
\usepackage{xcolor}
  \usepackage{longtable}
  \usepackage{placeins}

\AtBeginDocument{\hypersetup{hidelinks}}

\begin{document}

\title{Kineo: Calibration-Free Metric Motion Capture From Sparse RGB Cameras}

\author{Charles Javerliat}
\email{charles.javerliat@ec-lyon.fr}
\affiliation{%
  \institution{École Centrale de Lyon, CNRS, INSA Lyon, Université Claude Bernard Lyon 1, Université Lumière Lyon 2, LIRIS, UMR5205, ENISE}
  \city{Saint-Etienne}
  \country{France}
}

\author{Pierre Raimbaud}
\email{pierre.raimbaud@ec-lyon.fr}
\affiliation{%
  \institution{École Centrale de Lyon, CNRS, INSA Lyon, Université Claude Bernard Lyon 1, Université Lumière Lyon 2, LIRIS, UMR5205, ENISE}
  \city{Saint-Etienne}
  \country{France}
}

\author{Guillaume Lavoué}
\email{guillaume.lavoue@enise.ec-lyon.fr}
\affiliation{%
  \institution{École Centrale de Lyon, CNRS, INSA Lyon, Université Claude Bernard Lyon 1, Université Lumière Lyon 2, LIRIS, UMR5205, ENISE}
  \city{Saint-Etienne}
  \country{France}
}

\renewcommand{\shortauthors}{Javerliat et al.}

\begin{abstract}
Markerless multi-view motion capture is often constrained by the need for precise camera calibration, limiting accessibility for non-experts and in-the-wild captures. Existing calibration-free approaches mitigate this requirement but suffer from high computational cost and reduced reconstruction accuracy.
We present Kineo, a fully automatic, calibration-free pipeline for markerless motion capture from videos captured by unsynchronized, uncalibrated, consumer-grade RGB cameras. Kineo leverages 2D keypoints from off-the-shelf detectors to simultaneously calibrate cameras, including Brown-Conrady distortion coefficients, and reconstruct 3D keypoints and dense scene pointmaps at metric scale. A confidence-aware spatio-temporal keypoint subsampling strategy, combined with graph-based global optimization, enables robust calibration at a fixed computational cost independent of sequence length. We further introduce a pairwise reprojection consistency score, a heuristic measure of the agreement between views, which we use to weight bone lengths in metric scale recovery.
Evaluations on EgoHumans and Human3.6M demonstrate substantial improvements over prior calibration-free methods. Compared to previous state-of-the-art approaches, Kineo reduces camera translation error by approximately 83--85\%, camera angular error by 86--92\%, and world mean-per-joint position error (W-MPJPE) by 83--91\%.
Kineo is also efficient in real-world scenarios, processing multi-view sequences faster than their duration in a specific configuration (e.g., 36\,min to process 1\,h\,23\,min of footage). The full pipeline and evaluation code are openly released to promote reproducibility and practical adoption at: 
\href{https://liris-xr.github.io/kineo/}{liris-xr.github.io/kineo}.
\end{abstract}

\maketitle

\section{Introduction}
Motion capture technology is an essential tool for converting physical movement into digital representations. Traditional optical systems (e.g., OptiTrack~\cite{optitrack2025}, Vicon~\cite{vicon2025}, Qualisys~\cite{qualisys2025}) rely on multiple calibrated infrared cameras that track reflective markers placed on key locations of the subject's body. An animated skeletal model is then reconstructed for the motion to be analyzed or re-targeted on a virtual character for a realistic animation. While these systems achieve sub-millimeter accuracy in controlled studio conditions, they require subjects to wear optical markers, which can induce behavioral biases and make their use challenging or impractical. Moreover, deployment for in-the-wild scenarios (e.g., outdoor sport, theatrical performances or concerts, craftsmanship), or for casual users (e.g., home fitness, motion-based gaming) is difficult: the setup requires careful calibration, controlled lighting with limited infrared perturbations, and precise marker placement on the body. All cameras are wired together for temporal synchronization and connected to power outlets, limiting on-location recordings in uncontrolled environments. Finally, the specialized hardware and studio requirements make these systems costly and inaccessible to most users.

Recent advances in computer vision and deep learning have led to markerless motion capture methods, which improve accessibility by eliminating optical markers and using consumer-grade RGB cameras. Still, monocular methods suffer from depth ambiguities and occlusions, and multi-view methods require precisely calibrated and synchronized cameras, a limiting assumption in non-studio settings where cameras are often moved to accommodate different scenes or subjects. Calibration is typically performed with a known pattern (a checkerboard or a wand), and temporal synchronization with dedicated hardware or software during capture. Although effective, this process is tedious and error-prone, especially for non-experts, and improper calibration or synchronization can render the footage unusable for methods that assume known camera parameters.

Calibration-free methods address these challenges by recovering camera parameters post-capture from video and audio information. Substantial progress has been made recently, especially in reconstructing cameras, humans, and scene at metric scale~\cite{mullerReconstructingPeoplePlaces2025,rojasHamst3r2025}. Yet several limitations remain: a notable accuracy gap with calibrated methods persists, current methods are not yet well suited to long sequences, and they are difficult to run on consumer hardware. State-of-the-art methods rely on several computationally intensive networks and on non-linear optimization performed independently on each frame. As a result, cost grows linearly with sequence length, and the calibration of each frame does not benefit from evidence gathered in the others.

Our key observation is that the precision and confidence scores of modern 2D keypoint detectors provide 2D cues reliable enough to feed directly into a classical Structure-from-Motion (SfM) pipeline to calibrate the cameras. Camera calibration then needs neither a parametric body model nor any per-frame network beyond the 2D keypoint detector, and this design gives two system-level properties. First, calibration scales well to long sequences. Since keypoints are not constrained by a body model and can be filtered individually, observations from any frame can contribute to a single optimization. Unreliable correspondences can thus be filtered out across all frames and views, and the remaining ones subsampled into a fixed-size set from which the camera parameters are estimated. Second, the pipeline is modular: since its stages (2D keypoint detection, camera calibration, 3D reconstruction, and metric scale recovery) communicate only through explicit, interpretable data such as keypoints and camera parameters, each stage can be independently replaced by a stronger alternative without retraining an end-to-end model.

We build on this observation to introduce Kineo (Fig.~\ref{fig:method_overview}), a fully automatic, calibration-free motion capture system for unsynchronized, uncalibrated, consumer-grade RGB cameras. From the 2D keypoints of an off-the-shelf detector, Kineo jointly calibrates the cameras, including lens distortion, and reconstructs 3D keypoints and dense scene pointmaps at metric scale. It runs end-to-end from raw video and audio (the latter used for temporal alignment), in an offline mode for long captures and an online mode for interactive feedback. Our contribution lies at the system level: we deliberately build on well-understood geometric principles, whose behavior is predictable and whose interfaces are explicit, and adapt and integrate them into a pipeline whose calibration scales well to long sequences and whose stages are individually replaceable. The pipeline consists of five modules: 2D detection, calibration initialization, bundle adjustment, 3D consistency scoring, and metric scale recovery. The first uses an off-the-shelf detector; for each of the other four we propose a default instantiation:
\begin{itemize}
    \item \textbf{Confidence-Aware Calibration Initialization.} We select keypoint correspondences according to 2D detector confidences, which yields high-quality matches for robust and efficient RANSAC-based essential-matrix estimation. We then model the camera network as a weighted graph, weight each edge by a pairwise calibration-quality metric (average Sampson distance), and extract a minimum spanning tree (MST) to initialize a globally consistent configuration of camera extrinsics, so that a poor pair cannot corrupt the whole network.
    \item \textbf{Distortion-Aware Bundle Adjustment.} We estimate lens distortion coefficients during bundle adjustment, which prior calibration-free pipelines omit; on the wide-angle cameras of EgoHumans this reduces W-MPJPE from 0.41 to 0.17\,m over the same pipeline without it (Table~\ref{tab:human_metrics}).
    \item \textbf{3D Consistency Score.} We lift 2D confidences into 3D by defining a \textit{pairwise reprojection consistency score} for each triangulated point, which weights the bone lengths in our SMPL-based scale recovery and, qualitatively, flags the joints on which the views disagree.
    \item \textbf{Metric Scale Recovery.} To address global scale ambiguity, we introduce and evaluate two independent approaches: (1) a human-prior-based method that leverages the SMPL model~\cite{SMPL:2015} when people are present in the scene, and (2) a model-agnostic method using monocular metric depth estimation, less accurate but applicable when no human prior is available.
\end{itemize}

We evaluate the full system on the EgoHumans and Human3.6M datasets, demonstrating substantial improvements over prior cal\-i\-bra\-tion-free methods: camera translation error decreases by approximately 83--85\%, camera angular error by 86--92\%, and world mean-per-joint position error (W-MPJPE) by 83--91\%. We further validate Kineo in real-world settings, in both offline and online (real-time) scenarios, where it processes extended sequences faster than their duration: six five-minute views at 50\,Hz (30 minutes of footage) are processed in 11 minutes on an NVIDIA RTX 3080 Ti, excluding the cross-view re-identification of subjects (Sec.~\ref{sec:use_case_offline}). Our pipeline and evaluation code are released as open source at \href{https://liris-xr.github.io/kineo/}{liris-xr.github.io/kineo}.

\section{Related Work}
Achieving markerless motion capture in uncontrolled environments, particularly for non-expert users, requires addressing several challenges to enable a fully automated pipeline. These challenges include synchronizing battery-powered cameras, estimating camera parameters without manual calibration, resolving global metric scale, and maintaining computational efficiency suitable for consumer-grade hardware. Although recent advances in computer vision and deep learning have significantly enhanced the accessibility of markerless motion capture, especially in monocular settings, many multi-view methods still rely on pre-calibrated, synchronized camera setups or assume known intrinsics and extrinsics. These constraints limit their practicality in real-world or casual scenarios where such prerequisites are unavailable without manual calibration. In contrast, our work targets the calibration-free setting, aiming to operate directly on raw, unsynchronized video streams without requiring specialized hardware or manual setup.

In the following, we review related work across five key dimensions central to our method: (1) temporal synchronization, (2) automatic camera calibration, (3) metric scale recovery, (4) 3D uncertainty and consistency estimation, and (5) performance and system-level considerations. We cover both classical geometric pipelines and recent learning-based approaches, emphasizing the assumptions they make, their applicability in uncontrolled environments, and their relevance to the broader goal of general-purpose calibration-free motion capture.

\subsection{Temporal Synchronization}

Temporal synchronization refers to the process of recovering the time offset between the camera internal clock and a global time reference to express the timestamps of each frame in a shared world time reference. Temporal synchronization is essential in multi-view motion capture to synchronize camera frames, as even small temporal offsets between cameras can lead to inaccuracies in the reconstruction due to visual differences in the observed poses across views, especially when dealing with fast movements~\cite{shuaiNovelViewSynthesis2022,bragagnoloMultiviewPoseFusion2024,voSpatiotemporalBundleAdjustment2020}. Traditional studio camera systems rely on physical wiring to trigger all cameras simultaneously. However, such hardware-based synchronization is not available or impractical when using consumer-grade devices like smartphones or action cameras. Alternatively, jam syncing can be used. This technique synchronizes the internal clocks of multiple recording devices by transferring a timecode from a master device to slave devices. The synchronization must be performed at the start of the recording and periodically throughout the session to prevent clock drift. Once the slave devices receive the timecode, they operate independently while remaining aligned with the master device's time reference. Importantly, jam syncing requires dedicated hardware~\cite{tentacle_sync2025} or software~\cite{social_sync2014} at capture time.
When dedicated synchronization hardware or software is unavailable or was not used during capture, post-capture temporal synchronization methods can be employed. Among these, audio-based techniques have been used~\cite{shrsthaSynchronizationMulticameraVideo2007,haslerMarkerlessMotionCapture2009}. These methods leverage the sound captured by each camera to find the time offsets (temporal extrinsics). \citet{haslerMarkerlessMotionCapture2009} cross-correlate the audio waveforms recorded by the cameras, computing the cross-correlation with the Fast Fourier Transform, and take the lag at its peak as the time offset. When cameras are located at varying distances from the sound source, differences in audio propagation time can cause desynchronization. For instance, a 20-meter distance difference between two cameras recording at 60 Hz would lead to an estimated misalignment of around 4 frames. However, such large disparities are rare in practice, since the sound is typically emitted by the subject, who is usually positioned near the center of a circular or spherical camera setup. In these configurations, the propagation delay is negligible.
Another post-capture synchronization approach consists of estimating the time offset from the images rather than from the audio. \citet{sinhaCameraNetworkCalibration2010} and~\citet{albl2017} recover the time offset of each camera pair jointly with its epipolar geometry, from silhouettes of moving objects or people and from point trajectories, respectively. Other methods assume that a coarse time offset, accurate to within a few frames, is available and refine it jointly with the 3D motion. \citet{shuaiNovelViewSynthesis2022} include the time offset as a variable to be optimized within a joint estimation framework, assuming identical frame rates between all cameras. \citet{voSpatiotemporalBundleAdjustment2020} model the frame rate, time offset and rolling shutter pixel readout speed of each camera; assuming known frame rates, they optimize the time offsets jointly with the camera parameters and 3D trajectories, and estimate the readout speed with a separate self-calibration. To perform subframe alignment, \citet{shuaiNovelViewSynthesis2022} interpolate poses to a common time frame for their optimization process. They estimate joint velocities using finite differences, allowing them to interpolate poses using a first-order approximation. While computationally efficient, this approach is sensitive to noise. To improve robustness, \citet{voSpatiotemporalBundleAdjustment2020} fit a Discrete Cosine Transform (DCT) basis to the estimated trajectories to resample them as a post-processing step, which is less sensitive to noise. More recently, \citet{Choi_2025_ICCV} proposed using dynamic time warping (DTW) between every pair of videos to compute a constant time offset that minimizes the distance between 3D joint positions, extracted from estimated SMPL-H~\cite{romeroEmbodiedHandsModeling2017} models obtained via SLAHMR~\cite{SLAHMR2023} for each video individually. These integer offsets are then refined to subframe accuracy by jointly optimizing them with a dynamic NeRF~\cite{fridovichkeilKPlanesExplicitRadiance2023}. However, each pairwise DTW compares every frame of one video with every frame of the other, which can become a computational bottleneck for long sequences.

Ultimately, while image-based synchronization offers improvements in temporal precision, the practical balance of accuracy, speed, and ease of use tends to favor audio-based methods for most consumer-grade multi-view capture scenarios, where sound propagation delay is negligible. 
We therefore build on \citet{haslerMarkerlessMotionCapture2009} and estimate the time offsets from the cross-correlation of the audio tracks, but represent each track using Mel-Frequency Cepstral Coefficients (MFCCs)~\cite{davisComparisonParametricRepresentations1980} rather than raw audio waveforms, and derive conditions under which subframe synchronization is achievable (Sec.~\ref{sec:temporal_sync}).

\subsection{Automatic Camera Calibration}

Accurate estimation of camera parameters, both intrinsics (e.g., focal length, principal point, distortion coefficients) and extrinsics (position and orientation), is essential for multi-view 3D reconstruction. Examples include the use of such parameters in methods that first regress 2D keypoints or features and subsequently lift them into 3D space through techniques such as triangulation~\cite{zhangAdaFuseAdaptiveMultiview2021,XuUncalipose_2022_BMVC} or volumetric fusion~\cite{iskakovLearnableTriangulationHuman2019,VoxelPose2020,FasterVoxelPose2022,CrossViewFusion2019}. In optimization-based pipelines, the estimated camera parameters are used to reproject 3D structures (commonly a SMPL model or skeletal representation) into 2D space, minimizing the reprojection error with respect to the observed 2D cues~\cite{bragagnoloMultiviewPoseFusion2024}. While these methods exploit accurate camera parameters, most assume access to ground-truth values provided by datasets or obtained through manual calibration, rather than estimating them. This section focuses on existing methods for automatically estimating camera intrinsics and extrinsics.

\subsubsection{Camera Intrinsics}
Camera intrinsics refer to the internal characteristics of the camera, typically represented in the intrinsic matrix $K$. These include the focal length, principal point (the intersection of the optical axis with the image plane), and skew (often negligible). In the idealized pinhole camera model, these parameters are sufficient for projection. However, real-world lenses introduce optical distortions that cause straight lines in the scene to appear curved in the image.

High-fidelity intrinsic calibration is essential in both monocular and multi-view systems, as it directly impacts the accuracy of the 2D-3D mapping through the projection operation. In monocular regression-based approaches that predict 3D structure directly from 2D inputs, having access to precise intrinsics at inference time has been shown to improve the quality of the reconstruction~\cite{tokenHMR,kocabasSPECSeeingPeople2022,patel2024camerahmr,sarandiNeuralLocalizerFields2024,multiHMR}. Incorporating intrinsics into the network inputs enables more generalizable performance compared to methods using a simplified model of orthographic projection or fixed training-set intrinsics. These simplified models introduce a camera bias, resulting in a trade-off: achieving more accurate 3D pose estimates can come at the expense of poorer alignment with 2D image features, and vice versa~\cite{tokenHMR,kocabasSPECSeeingPeople2022}.

In multi-view reconstruction, intrinsics are equally critical. This holds both for methods where intrinsics are used to unproject the 2D keypoints or features into a shared 3D space~\cite{iskakovLearnableTriangulationHuman2019,VoxelPose2020,FasterVoxelPose2022,CrossViewFusion2019} and for optimization-based methods when computing the reprojection loss to align the 3D model with 2D cues~\cite{hewittLookMaNo2024,mullerReconstructingPeoplePlaces2025}.

Because camera intrinsics are fixed properties of the camera itself and independent of its viewpoint, they can be obtained through offline calibration after data capture. Consequently, some methods relax the constraints on obtaining these parameters by using ground-truth intrinsics, assuming the camera is accessible for later calibration and thus only focus on recovering the camera extrinsics~\cite{XuUncalipose_2022_BMVC,hewittLookMaNo2024,Choi_2025_ICCV}. However, this assumption limits applicability in scenarios involving online videos, unknown camera setups, or fully calibration-free approaches.

Several learning-based methods have been proposed to enable calibration-free inference by estimating camera parameters directly from image data. Within the human pose estimation domain, some monocular approaches focus on estimating the camera field of view from images containing humans~\cite{patel2024camerahmr,kocabasSPECSeeingPeople2022} to feed them to the body shape and pose regression model or for later optimization.

More recently, multi-view approaches have also incorporated intrinsic parameter estimation. For instance, HAMSt3R~\cite{rojasHamst3r2025} builds upon DUSt3R~\cite{wangDUSt3RGeometric3D2024} and MASt3R~\cite{leroy2024groundingimagematching3d}, recovering intrinsics from the predicted pointmaps but with a model specialized for human-centric reconstruction which yields better reconstruction quality. Similarly, \citet{mullerReconstructingPeoplePlaces2025} leverage DUSt3R~\cite{wangDUSt3RGeometric3D2024} to initialize camera intrinsics that are subsequently refined in a global optimization loop. Both methods remain tailored to human subjects and do not estimate distortion coefficients.
More broadly, other learning-based methods have been introduced for general-purpose applications, predicting both the focal length and lens distortion parameters~\cite{GeoCalib,AnyCalib,DeepCalib}, or just the camera focal length~\cite{wangMoGeUnlockingAccurate2025,leroy2024groundingimagematching3d,wangDUSt3RGeometric3D2024} from RGB images. In our method, we adopt the general model MoGe-2~\cite{wangMoGeUnlockingAccurate2025} to predict an initial camera focal length for each view (see Sec.~\ref{sec:intrinsics_init}). The focal length is subsequently optimized in the bundle adjustment step (see Sec.~\ref{sec:graph_bundle_adjustment}). To our knowledge, human pose estimation methods, whether monocular or multi-view, typically assume distortion-free inputs, relying on ground-truth calibration to undistort images as a preprocessing step prior to reconstruction or evaluation or simply neglecting distortion altogether. In contrast, our pipeline also estimates distortion coefficients, following the Brown-Conrady model~\cite{conradyDecentredLensSystems1919, brownDecenteringDistortionLenses1966, brownCloseRangeCameraCalibration1971}, by integrating them in the bundle adjustment (see Sec.~\ref{sec:graph_bundle_adjustment}), enabling the handling of lens distortion and improving multi-view consistency.

\subsubsection{Camera Extrinsics}
Camera extrinsics define the position and orientation of a camera within a shared world coordinate system. They are represented as a rigid transformation consisting of a rotation matrix $R$ and a translation vector $\mathbf{t}$, which map 3D points from world to camera coordinates. Accurate extrinsics are critical for multi-view motion capture, as they enable consistent integration of information across cameras. Misestimated extrinsics lead to geometric misalignment, resulting in inaccurate 3D reconstructions.

Various approaches have been proposed to estimate extrinsics in multi-camera setups, broadly falling into learning-based human-centric methods and classical geometric Structure-from-Motion (SfM) approaches.

Learning-based human-centric methods often leverage strong priors on human shape and pose. \citet{Choi_2025_ICCV} first estimate monocular human motion in each view using SLAHMR~\cite{SLAHMR2023}, which predicts SMPL-H~\cite{romeroEmbodiedHandsModeling2017} body meshes per frame. They compute an initial extrinsic estimate by minimizing per-joint distances to align motions across views, followed by refinement using a NeRF-based photometric optimization~\cite{mildenhallNeRFRepresentingScenes2020, fridovichkeilKPlanesExplicitRadiance2023}. This achieves sub-degree rotational and sub-centimeter translational errors on the CMU Panoptic Studio dataset~\cite{jooPanopticStudioMassively2015}.

\citet{hewittLookMaNo2024} adopt a PnP-based strategy, aligning 3D head meshes with 2D facial landmarks, following the method introduced by \citet{szymanowiczPhotorealistic360Head2022}. The head serves as a stable reference as its overall shape remains relatively stable across facial expressions and head poses. Initial estimates are refined through direct optimization of camera rotations and translations.

\citet{mullerReconstructingPeoplePlaces2025} use DUSt3R~\cite{wangDUSt3RGeometric3D2024} to predict dense 3D pointmaps from image pairs, recover relative poses, and integrate them into a global optimization that simultaneously updates SMPL-X~\cite{pavlakosExpressiveBodyCapture2019} models, scene pointmaps, and camera parameters. Similarly, HAMSt3R~\cite{rojasHamst3r2025}, trained specifically on human-centric scenes, aligns multiple views using predicted pairwise pointmaps within a common global frame, based on the method from MASt3R~\cite{leroy2024groundingimagematching3d}.

Despite differences, these methods share a reliance on parametric body models (e.g., SMPL~\cite{SMPL:2015} or SOMA~\cite{hewittLookMaNo2024}) or human-specific scene training. Such priors provide strong constraints that help resolve global scale ambiguities, as body shape enforces a consistent metric scale, but are computationally intensive, especially when optimization to refine initial camera poses and body poses is involved, like in HSfM~\cite{mullerReconstructingPeoplePlaces2025}.

Classical SfM approaches estimate extrinsics by exploiting geometric correspondences between views. Frameworks like COLMAP~\cite{schoenberger2016sfm} perform incremental bundle adjustment using traditional feature matching. While computationally efficient, these methods are sensitive to noise, outliers, and unreliable correspondences.

Outlier rejection is commonly implemented by embedding the estimation algorithm within a RANSAC~\cite{fischlerRandomSampleConsensus1981} framework~\cite{puweinJointCameraPose2015,XuUncalipose_2022_BMVC}, which robustly estimates relative camera poses from noisy keypoint correspondences. However, the number of correspondences grows with the length of the sequence, and occluded or poorly localized keypoints lower the inlier ratio. We therefore discard low-confidence correspondences and draw a fixed number of the remaining ones at random across all frames.

Early motion-capture SfM approaches also exploited semantic cues from the human body, including silhouettes~\cite{tsuhanchenAccurateSelfcalibrationTwo2007,lvCameraCalibrationVideo2006,krahnstoeverBayesianAutocalibrationSurveillance2005,sinhaCameraNetworkCalibration2010}, bounding box centers~\cite{xuWideBaselineMultiCameraCalibration2021}, and 2D keypoints~\cite{puweinJointCameraPose2015,XuUncalipose_2022_BMVC}, with keypoints proving particularly robust for establishing correspondences.

UnCaliPose~\cite{XuUncalipose_2022_BMVC} follows a similar principle, using 2D keypoints to estimate initial camera poses that are later refined through joint bundle adjustment with 3D human poses. Although effective, the method depends on known intrinsics, either provided as ground truth or obtained from an external estimator, does not reconstruct the scene, and shows limited accuracy in metric-scale outputs. To improve calibration, UnCaliPose adopts a two-step approach, first calibrating cameras across multiple frames and then solving for human poses, but the paper does not specify how these frames are selected. Without a selection strategy, incorporating all frames makes joint bundle adjustment computationally infeasible for long sequences. Our approach builds upon this framework by also leveraging 2D keypoints, but extends it to estimate camera intrinsics and distortion parameters, reconstruct a scene pointmap (as in HSfM and HAMSt3R), and produce metric-scale results with per-keypoint consistency scores. Our confidence-aware spatio-temporal keypoint subsampling restricts calibration and bundle adjustment to reliable detections, at a computational cost that remains fixed as sequences grow longer.

Once pairwise relative poses are estimated using SfM, the next step is to recover the global camera calibration by composing these pairwise transformations into a consistent network. Greedy strategies, such as selecting the next camera based on maximum inliers~\cite{puweinJointCameraPose2015}, offer efficiency, but accumulate errors and provide a suboptimal solution. \citet{XuUncalipose_2022_BMVC} instead validate each pairwise pose against indirect estimates obtained through a third camera, keeping the one with the smallest average alignment angle to the others. In addition, relative extrinsics computed via SfM are only determined up to an unknown scale factor, which must be resolved to properly compose them into the absolute extrinsics. \citet{XuUncalipose_2022_BMVC} address this by leveraging the fixed length of bones in 3D space to resolve scale ambiguities between camera pairs. However, this approach requires that the bone is visible and well triangulated by all camera pairs involved, which may not always be the case in challenging capture scenarios.

We formulate the calibration as a graph problem: cameras are nodes, pairwise calibrations forming the edges are weighted by a pairwise calibration score, and a minimum spanning tree selects the set of pairwise calibrations of minimum total cost. This formulation, which can be found in the literature as the epipolar graph~\cite{moulonGlobalFusionRelative2013, wilsonRobustGlobalTranslations2014}, allows us to solve for the relative scales via loop closure constraints.
To recover the global scale, we explore two approaches. The first draws inspiration from prior work using human body models; we estimate global scale as a downstream step using the shape priors of the SMPL model. The second leverages a monocular metric geometry estimator, MoGe-2~\cite{wangMoGeUnlockingAccurate2025} that we use to predict per-view depth maps in real-world units. This allows the reconstruction to be rescaled without any category-specific parametric model, but less accurately than the first approach.
Both strategies preserve the efficiency of a classical Structure-from-Motion pipeline while enabling metric-scale reconstruction. As a result, our method produces calibrated cameras, 3D keypoints, and a point cloud of the surrounding scene, matching the outputs of previous works.

\subsection{3D Uncertainty and Consistency}

Although the literature in 3D reconstruction focuses on recovering structure from monocular or multi-view input, most methods do not provide an explicit measure of uncertainty or consistency for their 3D estimates. Yet, such measures are useful for downstream tasks such as hypothesis fusion, temporal smoothing, and filtering, where occlusions and uncertain predictions must be handled appropriately.

Deterministic methods, which produce a single 3D estimate from one or more images, typically do not model uncertainty. Their output is unique, making it difficult to assess or compare reliability across views or time. This limitation is particularly problematic in multi-view scenarios, where certain views may be significantly more informative than others due to occlusion or pose ambiguity. In the absence of confidence scores, all views are treated equally during fusion, which can lead to suboptimal results.

To overcome this, probabilistic approaches have been introduced to model uncertainty by predicting distributions over plausible 3D poses. For example, ProHMR~\cite{prohmr} and HumaniFlow~\cite{humaniflow} sample multiple hypotheses from learned distributions, allowing per-joint or per-vertex uncertainty to be estimated from the variance of the samples. These uncertainty estimates enable view-consistent fusion either by optimizing the joint probability across views or by weighting individual predictions in a fusion step according to their confidence. However, this expressiveness comes at the cost of computational efficiency, as accurate uncertainty estimation typically requires drawing a large number of samples.

More efficient alternatives such as POCO~\cite{dwivediPOCO3DPose2023} have been proposed to address this trade-off. POCO extends deterministic models by directly predicting uncertainty alongside the 3D output, eliminating the need for sampling. While significantly faster, such methods remain tightly coupled to parametric models and are not readily applicable to non-parametric 3D representations such as lifted keypoints.

In this work, we introduce a consistency score tailored to lifted 3D keypoints in multi-view settings, independent of any body model. Our approach defines this score as the weighted mean reprojection agreement across all views, where each view's contribution is modulated by its 2D keypoint confidence. This formulation reflects how well the 3D reconstruction aligns with available 2D evidence, prioritizing views with reliable observations. The resulting score captures both geometric consistency and detection confidence, while remaining lightweight and applicable to a broad range of scenarios. Unlike the uncertainty estimates above, this heuristic measures agreement with the 2D evidence rather than 3D error; we use it to weight the bone lengths in metric scale recovery. We provide a detailed description in Sec.~\ref{sec:output_and_confidence}.

\subsection{Output Representation and Performance Considerations}

Despite substantial progress in markerless motion capture, many recent methods remain computationally intensive, rendering them impractical for long video sequences or deployment on consumer-grade hardware. For example, on an NVIDIA V100 GPU, HSfM requires about 118 seconds to process a single 4-view frame with three subjects, while HAMSt3R takes 14 seconds without its optional SMPL fitting~\cite{rojasHamst3r2025}. To put this into perspective, processing a 20-minute video at 25 frames per second would require approximately 41 days for HSfM and 5 days for HAMSt3R.

This limitation stems from the use of computationally heavy models applied to each frame individually and, in the case of HSfM, from an additional optimization loop involving numerous parameters, which further increases computational complexity.

A major source of complexity in this optimization arises from the type of 3D representation used. Parametric models~\cite{SMPL:2015,hewittLookMaNo2024,pavlakosExpressiveBodyCapture2019} are widely adopted due to their compactness and strong structural priors. These models encode pose and shape in low-dimensional spaces, enabling them to represent plausible pose and shape configurations even under occlusions or depth ambiguities. Such priors are particularly valuable in single-view settings, where geometric constraints are crucial to resolve ambiguity. However, in multi-view scenarios, where ambiguities are naturally reduced through additional viewpoints, these structural constraints become less critical. The computational cost of parametric models mainly comes from the iterative fitting needed to align model projections with 2D image evidence for each video frame. Each iteration evaluates the transformation stages of the model (e.g., pose decoding from latent space~\cite{pavlakosExpressiveBodyCapture2019}, blendshape evaluation, linear blend skinning). Although restricting skinning to the vertices from which the joints are regressed, as done during training in the code released with GVHMR~\cite{shenWorldGroundedHumanMotion2024}, makes each evaluation cheaper, the fitting must still be repeated for every frame. Moreover, these models are typically constrained to specific subject categories, such as humans, and are subject to biases like mean-shape bias~\cite{senguptaSyntheticTrainingAccurate2020, sarandiNeuralLocalizerFields2024}.

By contrast, non-parametric models, including keypoints, voxels, point clouds, etc., offer a more direct and computationally efficient alternative but do not provide inherent constraints. Keypoint-based representations, in particular, are widely used in multi-view and optimization-based pipelines due to their lightweight nature. While they lack explicit structural constraints, soft priors can be introduced to guide the optimization toward plausible solutions~\cite{bragagnoloMultiviewPoseFusion2024}. In cases where such priors are omitted, we refer to the models as~\textit{free-form} non-parametric representations. A hybrid strategy has also emerged, wherein non-parametric predictions are converted to parametric model parameters for compactness and use in downstream tasks. In particular, \citet{sarandiNeuralLocalizerFields2024} propose a fast method to fit SMPL meshes to dense non-parametric 3D keypoints.

Hybrid pipelines that combine learning-based regressors with subsequent optimization have shown strong performance and enhanced robustness in challenging scenarios. A representative example is HSfM, a calibration-free method that achieves good accuracy by leveraging multiple specialized models to jointly reconstruct cameras, scenes and human subjects. Specifically, HSfM employs HMR2.0~\cite{hmr2} for pose regression, DUSt3R~\cite{wangDUSt3RGeometric3D2024} to estimate the scene and cameras, ViTPose~\cite{xu2022vitpose} for keypoint detection, and SMPL-X~\cite{pavlakosExpressiveBodyCapture2019} for body modeling, and its released code additionally uses SAM2~\cite{sam2} to detect and associate people across views. These components provide complementary information that, when refined through a global optimization stage, leads to SOTA quality reconstructions. However, this architecture also increases computational complexity and results in longer runtimes due to the integration of several heavyweight models.

Our method achieves superior performance both in speed and accuracy through a more frugal architecture, leveraging only a monocular geometry estimator and a 2D keypoint detector. This minimal yet effective combination reduces inference time significantly, while improving reconstruction accuracy. Our approach enables faster and more scalable processing, making it suitable for long video sequences and consumer-grade hardware.

Beyond algorithmic complexity, existing pipelines are limited by system-level inefficiencies. Most existing methods operate on frames that are pre-extracted from video streams, introducing preprocessing delays and significant disk usage, especially for high-resolution or long-duration videos. To eliminate this overhead, we incorporate hardware-accelerated GPU decoding directly into our pipeline, avoiding the need to store or read individual frames. Furthermore, we extend MMPose~\cite{mmpose2020} to support fully GPU-resident, batched keypoint inference, removing redundant CPU-GPU transfers. These system-level improvements enable real-time performance with minimal memory and disk overhead. The resulting architecture is modular, facilitating the integration of alternative keypoint detectors and strategies with minimal effort.

\section{Method Overview}

\begin{figure*}[!htbp]
   \centering
   \includegraphics[width=\linewidth]{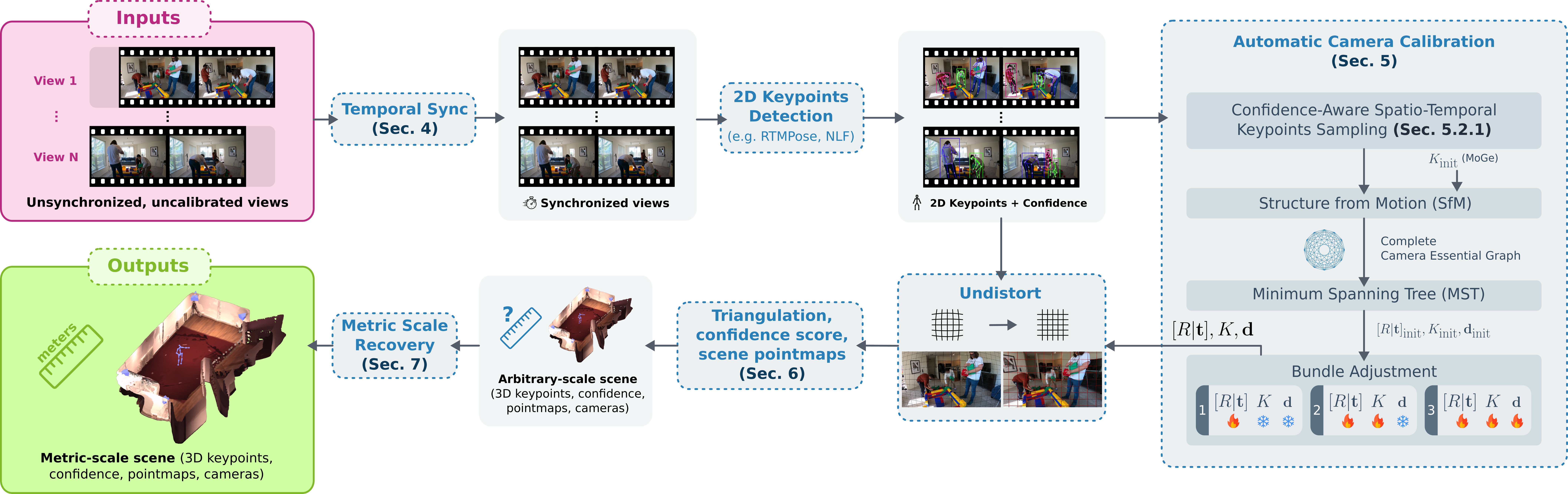}
   \caption{Overview of the Kineo pipeline for markerless motion capture from uncalibrated and unsynchronized multi-camera videos. Starting from raw video inputs, the system first performs audio-based temporal synchronization to align unsynchronized streams on a shared timeline. Next, automatic camera calibration estimates extrinsic and intrinsic parameters, including Brown-Conrady lens distortion, via a graph-based optimization over 2D keypoint correspondences selected by a confidence-aware subsampling strategy. Using the recovered cameras, 3D keypoints and scene pointmaps are reconstructed, and each triangulated keypoint is assigned a pairwise reprojection consistency score to measure the agreement between views. Finally, metric-scale recovery is achieved either through a human body prior using the SMPL model or a monocular metric depth estimator for subject-agnostic scaling. The modular design enables robust and scalable reconstruction of human subjects across long, multi-view sequences, and can extend to other subjects given a suitable 2D keypoint detector.}
   \label{fig:method_overview}
\end{figure*}

We propose a pipeline for performing markerless motion capture from uncalibrated and unsynchronized cameras. The pipeline is designed to handle any subject for which a 2D keypoint detector is available, which we evaluate on humans, including scenes with multiple individuals. It outputs metric-space 3D keypoints, scene pointmaps, and complete camera parameters, including extrinsics, intrinsics, and Brown-Conrady distortion coefficients. The pipeline is designed for modularity and efficiency, enabling scalable processing of long video sequences. An overview of the pipeline is shown in Figure~\ref{fig:method_overview}.

Given videos from unsynchronized and uncalibrated cameras, the pipeline first estimates temporal offsets between video streams via audio-based synchronization, aligning keypoint detections on a shared global timeline (see Sec.~\ref{sec:temporal_sync}).

Next, automatic camera calibration is performed using 2D keypoints detected by a replaceable off-the-shelf detector. A graph-based formulation is used to recover a global camera configuration through minimum spanning tree selection. Central to the robustness and efficiency of this stage is our keypoint subsampling strategy proposed in Sec.~\ref{sec:sampling}, which selects high-confidence correspondences to reduce noise and outliers while ensuring computational efficiency. This stage also estimates lens distortion parameters, which are often neglected by other approaches (see Sec.~\ref{sec:automatic_camera_calibration}).

Using the estimated cameras, scene pointmaps and skeletons are reconstructed in arbitrary units. We assign a consistency score to the triangulated keypoints, the pairwise reprojection consistency score (see Sec.~\ref{sec:pairwise_reprojection_confidence_score}), which weights the bone lengths during the metric-scale recovery stage (see Sec.~\ref{sec:global_scale}).

To obtain metric-scale reconstructions, we propose two strategies. The first leverages a human body prior, using the SMPL model when people are present in the scene, similar in concept to HSfM~\cite{mullerReconstructingPeoplePlaces2025}. The second is subject-agnostic, employing a monocular metric depth estimator to recover scale even in scenes without humans, at the cost of a lower accuracy (see Sec.~\ref{sec:global_scale}).

\section{Temporal Synchronization}
\label{sec:temporal_sync}

To relate 2D cues from different viewpoints, we must estimate temporal offsets between the videos. This is particularly important for battery-powered, unsynchronized cameras, where hardware-based synchronization is unavailable or impractical for easy-to-deploy setups. Unlike image-based methods~\cite{sinhaCameraNetworkCalibration2010,albl2017,voSpatiotemporalBundleAdjustment2020,shuaiNovelViewSynthesis2022,Choi_2025_ICCV}, we synchronize the cameras from audio, which keeps this stage agnostic to the tracked subject and to the keypoint detector. We take the temporal lag $\gamma_k$ of each camera as the peak of the cross-correlation between its audio track and that of a reference camera. To indicate when audio alone can give subframe accuracy, we state below a simple error budget relating the hop length of the audio features to the differences in camera-to-source distance.

This approach relies on three assumptions. First, all cameras record a common sound, clearly audible in each audio track. Second, audio and video share the same timeline within each device, so that the lag estimated on the audio tracks transfers directly to the video frames. This generally holds for smartphones and action cameras, which record both streams into a single container, but not when audio is captured by an external recorder or a wireless microphone. Third, the lag $\gamma_k$ corresponds to a constant frame offset over the whole sequence. This requires a constant frame rate, whereas smartphones commonly record at a variable one, and no clock drift between devices, whose oscillators tick at slightly different rates. In this work, we assume these effects to be negligible.

Rather than relying on raw waveform cross-correlation~\cite{haslerMarkerlessMotionCapture2009}, which is sensitive to differences in microphone response, we use compact audio descriptors to improve efficiency and robustness. Specifically, we represent each audio track using Mel-Frequency Cepstral Coefficients (MFCCs), a perceptually motivated and widely adopted feature in audio analysis. MFCCs capture the spectral envelope of the signal and are robust to noise. The reduced set of features allows for efficient and accurate cross-correlation, making them a suitable choice for estimating temporal lag in multi-camera settings.

The accuracy of this method depends on two main factors: the temporal resolution of the MFCC representation and the physical distance between the sound source and each camera. The former is determined by the hop length $H$ used during the Short-Time Fourier Transform (STFT), and the latter by the time delay resulting from different distances to the sound source. We formalize these two sources of error as $\Delta t_1$ and $\Delta t_2$, respectively, the latter having been identified by \citet{haslerMarkerlessMotionCapture2009}. We derive conditions under which subframe synchronization is achievable. Let $f_s$ be the audio sample rate, $f_r$ the video frame rate, $H$ the hop length, $d_1$ and $d_2$ the distances from each camera to the sound source respectively, and $c~\approx 343\,\text{m/s}$ the speed of sound in air. Then:
$$\Delta t_1=\frac{H}{2f_s}~\quad~\text{and}~\quad~\Delta t_2 =~\frac{|d_1 - d_2|}{c}$$
Subframe accuracy is ensured when the combined error is below a video frame duration:
$$\Delta t_1+\Delta t_2 <~\frac{1}{f_r}$$
which yields the following constraints:
\begin{equation*}
   H <~\frac{2f_s}{f_r}-\frac{2f_s|d_1-d_2|}{c}\quad\text{and}\quad|d_1-d_2| < c(\frac{1}{f_r}-\frac{H}{2f_s})
\end{equation*}

These inequalities guide the choice of hop length $H$ given the approximate scene geometry, or conversely, define the spatial constraints required for accurate synchronization when $H$ is fixed.

In typical multi-view camera arrangements (e.g., circular rigs), microphone distances to the sound source vary by only a few meters. For instance, with $|d_1 - d_2| < 5$\,m, $f_s = 48$\,kHz, and $f_r = 60$\,Hz, the hop length $H$ must be below 200 samples to ensure subframe accuracy. Conversely, with $H = 1$, the method remains accurate for inter-microphone distances below approximately 5.7\,m. Note that this bound only accounts for the temporal resolution of the audio features and the sound propagation delay, and should therefore be read as a best-case condition. It omits several other error sources: the per-device audio/video offset and the variable frame rate discussed above, clock drift between devices, ambiguities in the cross-correlation peak caused by background noise, reverberation, or multiple sound sources, and changes in the distances $d_1$ and $d_2$ when the sound source moves during the sequence. In all experiments, all audio tracks are resampled to the sample rate of the reference camera ($f_s = 48$\,kHz for AIST++). We compute $13$ MFCCs over $128$ HTK mel bands, with a $2048$-point FFT, a $40$\,ms window and a $5$\,ms hop ($H = 240$ samples). The lag is the maximum of the full cross-correlation, with no restricted search range or peak interpolation, so offsets are resolved to one hop ($5$\,ms). In case finer alignment is required, image-based synchronization techniques~\cite{voSpatiotemporalBundleAdjustment2020, shuaiNovelViewSynthesis2022} can be used to refine an initial audio-based estimate. We did not incorporate such image-based refinement in our pipeline.

\section{Graph-based Automatic Camera Calibration}
\label{sec:automatic_camera_calibration}

Recovering accurate camera extrinsics and intrinsics is essential for leveraging multi-view setups: without proper calibration, 3D reconstruction suffers from misalignments and degraded accuracy in downstream tasks. At the same time, manual calibration is impractical, especially for in-the-wild deployments.

We address the problem of automatic and efficient calibration of camera networks by modeling the multi-view system as a graph. Each of the $N$ nodes corresponds to a camera $C_i$, and edges represent pairwise relative geometry between two cameras $C_i$ and $C_j$, encoded by the essential matrix $E_{ij}$. These essential matrices are estimated via robust RANSAC-based Structure-from-Motion (SfM) on a subsampled high-confidence keypoint correspondence set (see Sec.~\ref{sec:sampling}). Edge weights reflect calibration quality, quantified by the average Sampson distance~\cite{sampsonFittingConicSections1982} over the sampled correspondences.

This graph formulation, as in other SfM methods~\cite[e.g.,][]{moulonGlobalFusionRelative2013, wilsonRobustGlobalTranslations2014}, enables global reasoning about pairwise relationships and allows us to select an optimal subset of calibrations through a minimum spanning tree (MST). This contrasts with prior incremental approaches that add cameras greedily to a global frame, at the risk of yielding a suboptimal global configuration~\cite{puweinJointCameraPose2015}. By composing relative poses along the MST, we recover global camera extrinsics efficiently with a guarantee of minimizing the total pairwise calibration error (Sec.~\ref{sec:graph_absolute_extrinsics}). Table~\ref{supp-tab:notation} of the supplementary materials lists the notation used in this section. The recovered cameras are expressed in arbitrary units, and metric scaling to real-world dimensions is applied in Sec.~\ref{sec:global_scale}.

\subsection{Intrinsics Initialization}
\label{sec:intrinsics_init}

To construct the camera graph, we need to recover the relative poses between camera pairs, which requires decomposing the essential matrices into their corresponding rotations and translation directions. Computing the essential matrix, however, depends on knowledge of the camera intrinsics $K$. In our scenario, where cameras are assumed to be static and the subject is moving, Kruppa’s formulation~\cite{kruppa1913,hartleyKruppaEquations} cannot be applied. One alternative is to compute fundamental matrices, which do not require intrinsics, and then optimize the intrinsics for each camera by minimizing the Sampson distance over all edges. However, full-graph optimization is computationally expensive, growing quadratically with the number of cameras. To maintain a linear cost with respect to the number of cameras, we instead use MoGe-2~\cite{wangMoGeUnlockingAccurate2025} to obtain an initial estimate of each camera’s intrinsics. This model is later reused to predict scene pointmaps. These intrinsics are later refined during bundle adjustment (Sec.~\ref{sec:graph_bundle_adjustment}), into which we incorporate lens distortion coefficients that follow the Brown-Conrady model~\cite{conradyDecentredLensSystems1919, brownDecenteringDistortionLenses1966, brownCloseRangeCameraCalibration1971}, a step omitted in prior works.

\subsection{Efficient Pairwise Calibration}
\label{sec:graph_relative_extrinsics}

We propose an approach for pairwise camera calibration that is both robust and scalable to long sequences due to its fixed computational cost. Our method builds upon the work of \citet{puweinJointCameraPose2015} by introducing confidence-aware spatio-temporal subsampling of keypoint correspondences, which enhances robustness while maintaining efficiency. Additionally, we leverage a graph-based formulation to compute relative scales reliably, incorporating the quality of each pairwise calibration to produce geometrically consistent relative extrinsics.

The edges of the graph represent pairwise camera calibrations, encoded by an essential matrix $E_{ij}$ that captures the relative rotation $R_{ij}$ and translation direction $\hat{\mathbf{t}}_{ij}$ between cameras $C_i$ and $C_j$. Classical approaches estimate this matrix using the Structure-from-Motion (SfM) five-point algorithm~\cite{nisterEfficientSolutionFivePoint} within a RANSAC loop applied either to all raw keypoint correspondences~\cite{XuUncalipose_2022_BMVC} or to a subset of correspondences~\cite{puweinJointCameraPose2015}. Both strategies can be unstable when low-quality correspondences are present, and processing all correspondences becomes computationally prohibitive for long sequences.

Learning-based methods such as DUSt3R~\cite{wangDUSt3RGeometric3D2024} predict relative poses in a single-frame multi-view setting and can be more robust when high-confidence keypoints are unavailable. However, they do not exploit temporal consistency, which can limit accuracy, and their reliance on an additional neural network increases computational overhead; for example, DUSt3R requires approximately 40 ms for inference on a single image pair using an H100 GPU. In contrast, our method is more computationally frugal because it fully exploits the same 2D keypoints that are already required for triangulating the final skeleton output. By selectively subsampling these keypoints based on their detection confidence, we feed RANSAC cleaner correspondences, so that it converges in fewer iterations, while preserving the efficiency of classical SfM and ensuring scalability to long sequences.

\subsubsection{Confidence-Aware Spatio-Temporal Subsampling}
\label{sec:sampling}

\begin{figure}
    \centering
\maxsizebox{\linewidth}{0.3\textheight}{%
    \includegraphics[width=\linewidth]{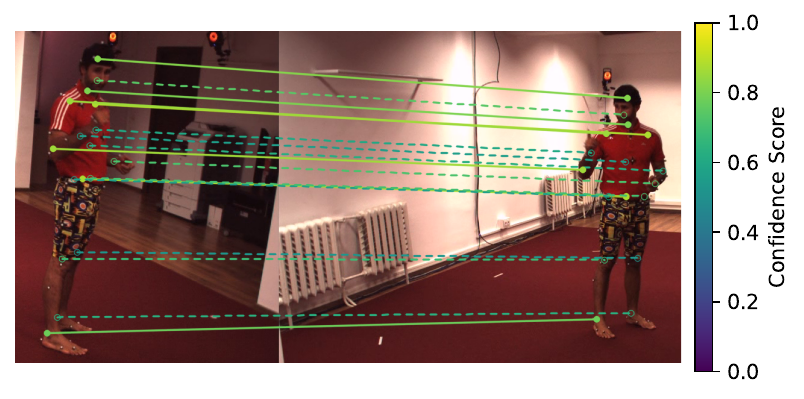}%
}
\caption{Example subsampling for a pair of views from Human3.6M. Colors indicate the pair confidence score $w_{ij,k}$, solid lines represent selected correspondences, and dashed lines represent correspondences that were not selected. Only correspondences whose pair score exceeds $\tau$ can be selected, which discards the keypoints the 2D keypoint detector does not locate confidently (e.g., occluded or poorly localized keypoints).}
    \label{fig:subsampling}
\end{figure}

An effective spatio-temporal subsampling strategy for keypoint correspondences is important for three reasons: (1) it gives RANSAC more reliable correspondences by filtering out low-quality ones, which reduces the number of iterations as shown in Sec.~\ref{sec:ablation_study_confidence}, and it leverages information across multiple frames, (2) it reduces computational complexity by limiting the number of correspondences, which is particularly important for scalability on long sequences, and (3) it preserves sufficient geometric diversity to ensure accurate essential matrix estimation. Traditional RANSAC samples correspondences uniformly, ignoring the varying reliability of detections. As a result, if low-confidence correspondences constitute a majority, they can disproportionately influence hypothesis generation, leading to inaccurate relative pose estimates and requiring more iterations to converge on a valid model. \citet{puweinJointCameraPose2015} addressed this by sampling correspondences with higher probabilities for temporally distant correspondences and for those involving different keypoints. However, their strategy does not leverage keypoint confidence scores, which can lead to the inclusion of unreliable correspondences. PROSAC~\cite{chumMatchingPROSACProgressive2005} also exploits a quality ranking, by drawing RANSAC hypotheses progressively from the highest-ranked correspondences first, which reduces the number of hypotheses needed to reach a solution. Our use of the detector confidence differs: rather than ordering hypothesis generation over all correspondences, we use it to filter which correspondences are eligible, then draw at most $P$ of them at random across the whole sequence.

Building on the method of \citet{puweinJointCameraPose2015}, we improve the spatio-temporal subsampling strategy by incorporating the 2D keypoint detector confidence scores $w_{i,k}~\in [0,1]$, with $i$ the view index and $k$ the keypoint index, into the selection process.
Each point correspondence between cameras $C_i$ and $C_j$ is denoted as $(\mathbf{x}_{i,k},\mathbf{x}_{j,k})$. Its score is computed as the geometric mean of the detector confidences:
\begin{equation}
\label{eq:pair_score}
w_{ij,k}=\sqrt{w_{i,k}~\cdot w_{j,k}}
\end{equation}
Only correspondences with $w_{ij,k}>\tau$ are retained, where the threshold $\tau$ is set empirically to 0.6 for the NLF detector~\cite{sarandiNeuralLocalizerFields2024}. From this filtered pool, a final set of at most $P=3000$ correspondences is randomly selected to implicitly promote temporal diversity and geometric coverage: since the pool gathers the confident correspondences of all frames, the selected correspondences are spread over the whole sequence and over the image regions where the subjects are detected.
An example of this confidence-aware subsampling is shown in Fig.~\ref{fig:subsampling}. This final set of correspondences is then used in the standard RANSAC loop to generate hypotheses and remove any remaining outliers while estimating the essential matrices. To verify that this implicit coverage is sufficient, Sec.~\ref{sec:ablation_study_sampling} compares random selection with farthest-point sampling~\cite{eldarFarthestPointStrategy1997}, which explicitly enforces spatio-temporal coverage, and with confidence-only sampling, which keeps the most confident correspondences: farthest-point sampling reaches nearly the same accuracy at a higher computational cost, and confidence-only sampling degrades calibration. We therefore keep random selection.

\subsection{Graph Edge Reliability Estimation}
\label{sec:graph_edge_reliability}

To assess the reliability of pairwise camera calibration, we measure the geometric consistency between its corresponding point sets. To make this measure independent of camera intrinsics, the points are first normalized by their intrinsic calibration matrix estimated in Sec.~\ref{sec:intrinsics_init}:
\begin{equation*}
\hat{\mathbf{x}}_{i,p} = K_i^{-1}\mathbf{x}_{i,p}~\quad~\text{and}~\quad~\hat{\mathbf{x}}_{j,p} = K_j^{-1}\mathbf{x}_{j,p}
\end{equation*}
The normalized Sampson distance for correspondence $p$ is:
\begin{equation*}
\maxsizebox{\linewidth}{!}{%
$\mathcal{S}(E_{ij}, \hat{\mathbf{x}}_{i,p}, \hat{\mathbf{x}}_{j,p}) =
\frac{(\hat{\mathbf{x}}_{j,p}^\top E_{ij} \hat{\mathbf{x}}_{i,p})^2}
{(E_{ij} \hat{\mathbf{x}}_{i,p})_x^2 + (E_{ij} \hat{\mathbf{x}}_{i,p})_y^2
 + (E_{ij}^\top \hat{\mathbf{x}}_{j,p})_x^2 + (E_{ij}^\top \hat{\mathbf{x}}_{j,p})_y^2}$
}
\end{equation*}
Finally, the calibration score for the camera pair is computed as the average Sampson distance, evaluated over the $P$ sampled correspondences:
\begin{equation}
e_{ij} = \frac{1}{P}\sum_{p=1}^{P}\mathcal{S}(E_{ij}, \hat{\mathbf{x}}_{i,p}, \hat{\mathbf{x}}_{j,p})
\end{equation}

This score is used as the weight of the edge connecting cameras $C_i$ and $C_j$ in the camera graph, reflecting the reliability of their relative pose.

\subsection{Solving Relative Scales}
\label{sec:graph_relative_scales}

The essential matrix $E_{ij}$ is only defined up to a relative scale factor $\lambda_{ij}$. Formally, given its decomposition into a relative rotation $R_{ij}\in SO(3)$ and a translation direction $\hat{\mathbf{t}}_{ij}\in \mathbb{R}^3$, the relation between two camera poses is expressed as:
\begin{equation*}
R_j=R_{ij}R_i~\quad~\text{and}~\quad~\mathbf{t}_j=R_{ij}\mathbf{t}_i+\lambda_{ij}\hat{\mathbf{t}}_{ij}
\end{equation*}

In previous work, finding this relative scale $\lambda_{ij}$ is addressed by aligning 3D structures in a common space. \citet{XuUncalipose_2022_BMVC} rescale and align the triangulated bones from each camera pair into a shared coordinate system. In their method, the quality of the pairwise triangulation directly affects the result when computing the relative scales. Methods like DUSt3R~\cite{wangDUSt3RGeometric3D2024} incorporate the scaling factors into a global alignment loop. The predicted pairwise pointmaps are then scaled and aligned into a globally consistent 3D reconstruction.

\begin{figure}[ht]
    \centering
\maxsizebox{!}{0.2\textheight}{%
    \includegraphics[width=\linewidth]{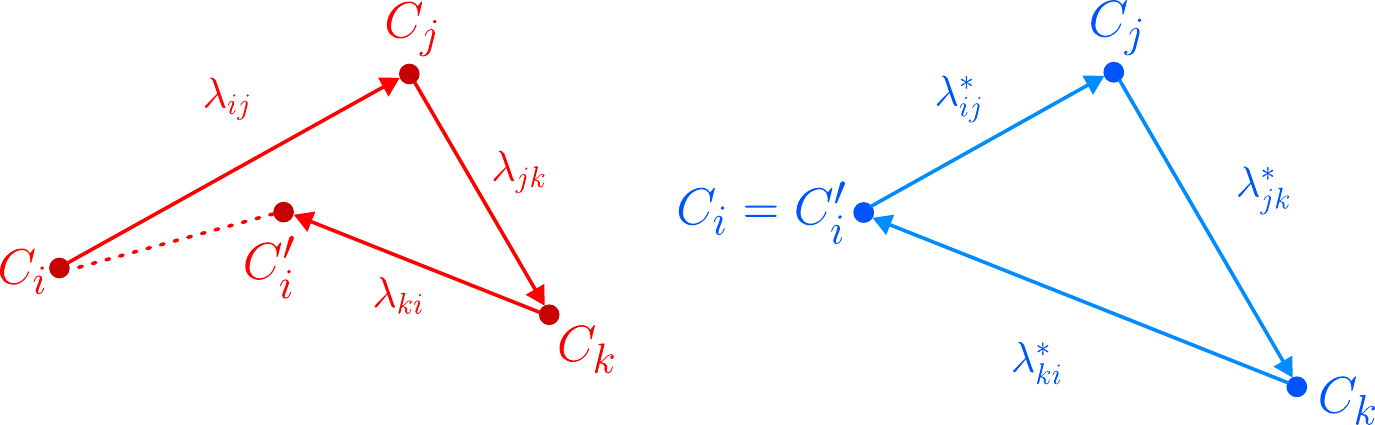}%
}
    \caption{Example of camera loop closure with incorrect (left) and correct (right) relative scale factors. 
    When the relative scales $\lambda_{ij}$ are incorrect, traversing the cycle $C_i \!\to\! C_j \!\to\! C_k \!\to\! C_i'$ does not return to the original camera position, resulting in $C_i \neq C_i'$. In contrast, when the correct scales are used, the composition of the relative transformations yields the identity, and the loop closes consistently with $C_i = C_i'$.}
    \label{fig:relative_scales}
\end{figure}

Instead, we solve for the relative scales $\lambda_{ij}$ by exploiting loop closure constraints~\cite{arrigoniComputingTranslationsNorm}. The key idea is that if we follow the relative transformations along a closed cycle of cameras, for example moving from $C_i$ to $C_j$ to $C_k$ and back to $C_i$, we should return to the starting point. This concept is illustrated in Fig.~\ref{fig:relative_scales}. Formally, each triplet of cameras yields one linear constraint of the form:
\begin{equation}
   ~\lambda_{ij} R_{ki} R_{jk}~\hat{\mathbf{t}}_{ij} +~\lambda_{jk} R_{ki}~\hat{\mathbf{t}}_{jk} +~\lambda_{ki}~\hat{\mathbf{t}}_{ki} =~\mathbf{0}
\end{equation}

There is one unknown relative scale factor per camera pair, which we gather into $\boldsymbol{\lambda}\in\mathbb{R}^{Q}$, with $Q$ the number of camera pairs. Collecting all $L$ loop constraints yields a homogeneous linear system $A\boldsymbol{\lambda}=\mathbf{0}$, where $A\in\mathbb{R}^{3L\times Q}$ contains the coefficients from the constraints.

Not all loop closure constraints are equally reliable. Some camera pairs (edges in the graph) may have less accurate essential matrices. Consequently, loops that include these less reliable edges are also less trustworthy. To account for this, we leverage the edge weights computed in Sec.~\ref{sec:graph_edge_reliability} to weight the loop closure constraints. Specifically, since $e_{ij}$ is a mean Sampson distance (lower is better), we define a loop cost from the scores of its three edges and weight each loop by its min-max normalized cost, so that the most and least reliable loops receive weights 1 and 0:
\begin{equation}
    c_{ijk}=\left(e_{ij}\,e_{jk}\,e_{ki}\right)^{\frac{1}{3}}
    \quad\text{and}\quad
    \omega_{ijk}=\frac{\max c-c_{ijk}}{\max c-\min c}
\end{equation}
Loops with an edge having fewer than the five correspondences required to estimate $E_{ij}$ are discarded ($\omega_{ijk}=0$). Repeating each $\omega_{ijk}$ over the three rows of its constraint in a diagonal matrix $W\in\mathbb{R}^{3L\times 3L}$, we formulate the weighted homogeneous system:
\begin{equation}
    W^{\frac{1}{2}}A\boldsymbol{\lambda}=\mathbf{0}
\end{equation}

We solve this system by nonlinear optimization. This formulation admits the trivial solution $\boldsymbol{\lambda}=\mathbf{0}$, and otherwise the solution is only determined up to a global scale factor. A common way to remove this ambiguity is to fix one variable, e.g., set $\lambda_{12}=1$. However, this can bias the solution in systems with many loop constraints where $\lambda_{12}$ is underrepresented. In such cases, all $\boldsymbol{\lambda}$ values tend to shrink toward zero to minimize the loss, except for the fixed $\lambda_{12}=1$. When composing the relative extrinsics into absolute extrinsics (see Sec.~\ref{sec:graph_absolute_extrinsics}), this often leads to inconsistent solutions, such as two clusters of cameras separated by the long translation scaled by $\lambda_{12}$. To avoid this, we instead add a regularization term that encourages all $\boldsymbol{\lambda}$ values to stay close to 1. Formally, we solve the following optimization problem:
\begin{equation}
\min_{\boldsymbol{\lambda}}\frac{1}{3L}\|W^{\frac{1}{2}}A\boldsymbol{\lambda}\|_2^2 + \frac{\mu}{Q}\|\boldsymbol{\lambda} - \mathbf{1}\|_1
\end{equation}
where $\mu$ is the regularizer weight, set to $10^{-3}$; averaging both terms keeps their balance from depending on the numbers of loops and camera pairs. In practice, we use the L-BFGS algorithm~\cite{liuLimitedMemoryBFGS1989} and solve for the logarithm of $\boldsymbol{\lambda}$ to ensure the exponentiated values remain positive.

\subsection{MST-Based Absolute Extrinsics}
\label{sec:graph_absolute_extrinsics}

Having estimated relative extrinsics, the next step is to recover the absolute poses of all cameras in a common global frame. Naively chaining relative transformations along arbitrary paths in the camera graph can lead to error accumulation, especially when some edges are less reliable. To limit this accumulation, we use the calibration reliability scores computed in Sec.~\ref{sec:graph_edge_reliability} to guide edge selection.

We define the first camera $C_{1}$ as the origin, whose pose is $[R\mid\mathbf{t}]_{1}=[I\mid\mathbf{0}]$, where $I$ is the identity rotation matrix and $\mathbf{0}$ the zero translation vector. Since multiple paths may connect $C_{1}$ to a camera $C_k$, each path accumulates errors differently depending on the quality of the relative transforms. To minimize the total pairwise calibration error, we construct a minimum spanning tree (MST) of the camera graph using Kruskal’s algorithm~\cite{kruskalShortestSpanningSubtree1956}, with edge weights corresponding to pairwise calibration reliability computed in Sec.~\ref{sec:graph_edge_reliability}. Absolute extrinsics are then computed by composing relative transformations along the unique MST path from $C_{1}$ to each camera $C_k$. Formally, if $\mathcal{P}^*_k$ denotes this MST path, the absolute pose $[R\mid\mathbf{t}]_{k}$ is obtained as

\begin{equation}
[R\mid\mathbf{t}]_{k}=~\prod_{(i,j)~\in \mathcal{P}^*_k}^{\leftarrow}[R\mid\mathbf{t}]_{ij}=\prod_{(i,j)~\in \mathcal{P}^*_k}^{\leftarrow}~\begin{bmatrix}R_{ij} &~\lambda_{ij}\hat{\mathbf{t}}_{ij}~\\~\mathbf{0}^\top & 1\end{bmatrix},
\end{equation}
where $R_{ij}$ and $\hat{\mathbf{t}}_{ij}$ are the relative rotation and unit translation vector from camera $C_i$ to $C_j$, and $\lambda_{ij}$ is the relative scale factor computed in Sec.~\ref{sec:graph_relative_scales}. This MST-based initialization provides a robust starting point for the subsequent bundle adjustment, which further refines camera poses and reduces residual errors.

\subsection{Bundle Adjustment}
\label{sec:graph_bundle_adjustment}

The MST-based initialization described in Sec.~\ref{sec:graph_absolute_extrinsics} provides a solid starting point for absolute camera extrinsics. However, residual errors remain due to the linear method used to estimate the pairwise essential matrices. Similarly, the intrinsics initialized in Sec.~\ref{sec:intrinsics_init} are only approximate and assume a simplified pinhole model without accounting for lens distortions. While other methods overlook distortion parameters, often undistorting images using the ground-truth parameters, we propose to integrate the distortion coefficients as parameters to be optimized in the bundle adjustment~\cite{triggsBundleAdjustmentModern2000}.

Optimizing over all detected keypoints across the sequence is computationally prohibitive. Instead, we refine camera parameters using a carefully selected subset of points, deferring triangulation of the full sequence to a later stage. Following the strategy in Sec.~\ref{sec:sampling}, we subsample keypoints to maintain a fixed computational cost and ensure scalability for long sequences. We select keypoints that are confidently detected in at least two views, which is the minimum requirement for triangulation:
\begin{equation}
\sum_{i=1}^{N} \mathbf{1}\!\left[w_{i,k} > \kappa \right] \;\geq\; 2,
\end{equation}
where $N$ is the number of views, $\kappa$ is a minimum keypoint confidence threshold (0.6 for NLF) and $\mathbf{1}\!\left[\,\cdot\,\right]$ the indicator function (equals 1 if the condition holds, 0 otherwise). The selected keypoints are then uniformly subsampled to a maximum of $M=500$ keypoints, and are triangulated to initialize the optimization, denoted as $\mathbf{X}_m \in \mathbb{R}^3$. The values of $\tau$, $P$, $\kappa$ and $M$ given here are those used with NLF; as the confidence scores of different detectors are not on the same scale, they are set empirically per detector, and Table~\ref{supp-tab:sampling_settings} of the supplementary materials lists their values in each experiment.

Bundle adjustment proceeds in three passes, with optimized parameters gradually added at each stage. In the first pass, only camera poses (extrinsics) and 3D points are optimized to reduce misalignments, while intrinsic parameters remain fixed. In the second pass, we jointly optimize camera poses, 3D points and focal lengths. In the third and final pass, distortion coefficients are added, following the Brown-Conrady model, which accounts for both radial and tangential distortions.
The principal point is kept fixed in all passes, and the focal length is optimized as a single value $f_x=f_y$. On our datasets, we share one set of intrinsics across views ($K_i=K$, $\mathbf{d}_i=\mathbf{d}$), initialized by averaging the per-view estimates of Sec.~\ref{sec:intrinsics_init}. This assumes that the cameras of a rig have the same intrinsics; for heterogeneous rigs, the intrinsics are optimized per view.
In every pass, the optimization goal is to minimize the weighted average reprojection error of the 3D points across all views:
\begin{equation}
\label{eq:reproj_residuals}
\epsilon_{i,m}=\|\pi(K_i,\mathbf{d}_i,[R|\mathbf{t}]_i,\mathbf{X}_m)-\mathbf{x}_{i,m}\|_2
\end{equation}
The residuals are then aggregated into a robust, weighted loss function using a Huber loss~\cite{huberRobustEstimationLocation1964} $\mathcal{H}_\delta(\cdot)$, where each residual is weighted by the 2D detector confidence $w_{i,m}\in[0,1]$ for the corresponding keypoint:
\begin{equation}
\label{eq:ba_objective}
\mathcal{L}_\text{reproj}=\frac{\sum_{i=1}^N \sum_{m=1}^M w_{i,m} \mathcal{H}_\delta \left(\epsilon_{i,m} \right)}{\sum_{i=1}^N \sum_{m=1}^M w_{i,m}}
\end{equation}
Here, $N$ is the number of views, $M$ the number of 3D points, $\epsilon_{i,m}$ is the reprojection residual, expressed in pixels, $\pi(\cdot)$ denotes the projection function, $K_i$ is the intrinsic matrix, $\mathbf{d}_i$ the Brown-Conrady distortion coefficients, and $[R \mid \mathbf{t}]_i$ the camera extrinsics. This formulation ensures that the optimization focuses on geometric misalignment while being robust to outliers, with the Huber threshold $\delta$ set to 1\,pixel. In the third pass, we add to $\mathcal{L}_\text{reproj}$ an L2 regularizer on the distortion coefficients, $\frac{\mu_\mathbf{d}}{5}\|\mathbf{d}\|_2^2$ averaged over the five Brown-Conrady coefficients, with $\mu_\mathbf{d}=1$, which keeps them small when the keypoints constrain them weakly.

Upon completion of bundle adjustment, the method yields refined camera extrinsics and intrinsics, as well as estimated lens distortion coefficients, which are typically not obtained by alternative approaches. The resulting camera extrinsics, however, are expressed in an arbitrary scale, and global scale still needs to be estimated to convert the reconstructed scene into metric units.

\section{3D Output and Consistency Score}
\label{sec:output_and_confidence}

Given accurate camera extrinsics and intrinsics, lifting 2D keypoints to 3D and reconstructing scene pointmaps can be formulated as downstream tasks. In this section, we describe the methods used for scene pointmap estimation and for triangulating and scoring the consistency of 3D keypoints.

\subsection{Scene Pointmaps}
\label{sec:scene_pointmaps}

We evaluated two models to reconstruct the scene pointmaps: MoGe-2~\cite{wangMoGeUnlockingAccurate2025} and VGGT~\cite{VGGT}. For both methods, each input image is first undistorted using the estimated distortion coefficients.

MoGe-2, being a monocular method, predicts a dense pointmap in camera space for each view, which is then merged into a global pointmap in world space using the camera extrinsics. To further improve prediction quality, we also provide MoGe-2 with the camera focal length estimated in earlier stages (see Sec.~\ref{sec:intrinsics_init} and Sec.~\ref{sec:graph_bundle_adjustment}).

In contrast, VGGT takes multiple views as input to jointly predict global pointmaps and camera poses. Since VGGT outputs its own camera parameters, we compute a similarity transform to align VGGT's predicted cameras with ours and apply this transform to the scene pointmap. Although this alignment is approximate due to VGGT’s cameras being less accurate than ours, this approach yields pointmaps that are more spatially consistent across views than those obtained with MoGe-2. However, VGGT is considerably more computationally intensive, and its complexity increases with the number of input cameras.

With either model, the pointmaps are tied to the skeletons only through the cameras and the global scale (Sec.~\ref{sec:global_scale}). No constraint relates the two, so contact is not enforced and feet can penetrate the floor.

\begin{figure}[ht]
    \centering
    \begin{subfigure}[b]{0.49\linewidth}
        \centering
        \includegraphics[width=\textwidth,trim=120 350 220 270,clip]{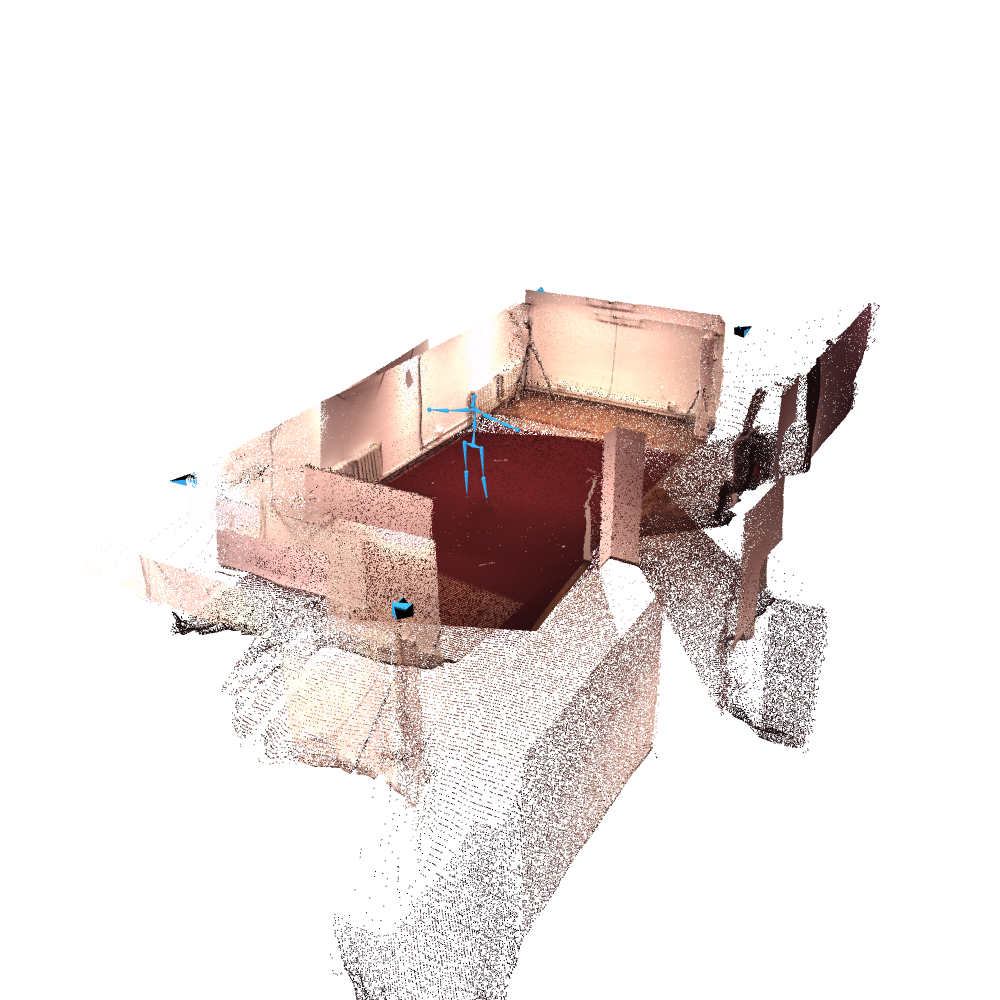}
        \caption{Scene pointmap generated with MoGe-2.}
        \label{fig:moge_scene_reconstruction}
    \end{subfigure}
    \hfill
    \begin{subfigure}[b]{0.49\linewidth}
        \centering
        \includegraphics[width=\textwidth,trim=120 350 220 270,clip]{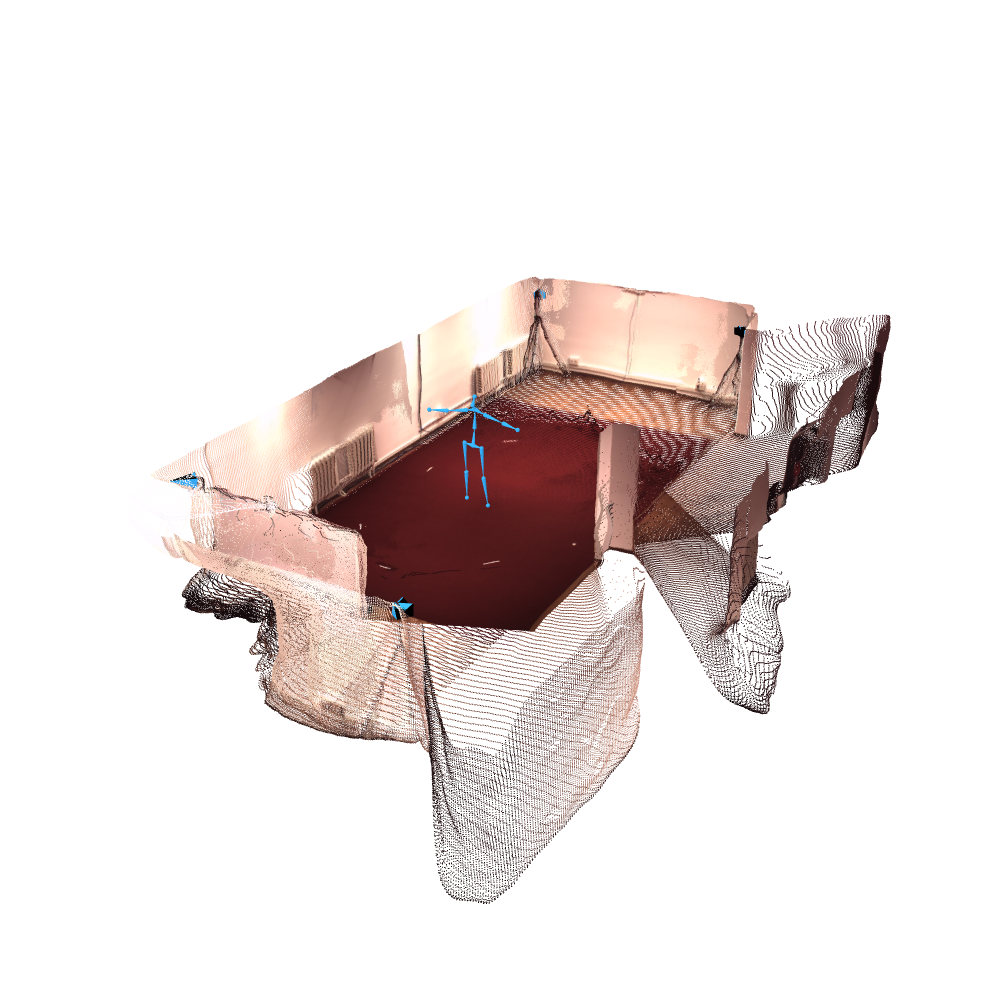}
        \caption{Scene pointmap generated with VGGT.}
        \label{fig:vggt_scene_reconstruction}
    \end{subfigure}
    \caption{Comparison of scene pointmaps produced by two geometry estimation models. (a) shows the pointmap generated with MoGe-2, which uses single-view images and camera intrinsics to produce metric pointmaps. (b) shows the pointmap generated with VGGT, which leverages multiple views to predict a globally consistent pointmap and camera poses, aligned to our camera system. VGGT provides more spatially consistent pointmaps across views, while MoGe-2 relies on single-view depth estimation.}
    \label{fig:both_subfigs}
\end{figure}

\subsection{Triangulation and Consistency Score}
\label{sec:triangulation}

\begin{figure}[h]
    \centering
    \begin{subfigure}{0.22\columnwidth}
        \centering
\maxsizebox{!}{0.2\textheight}{%
        \includegraphics[width=\linewidth]{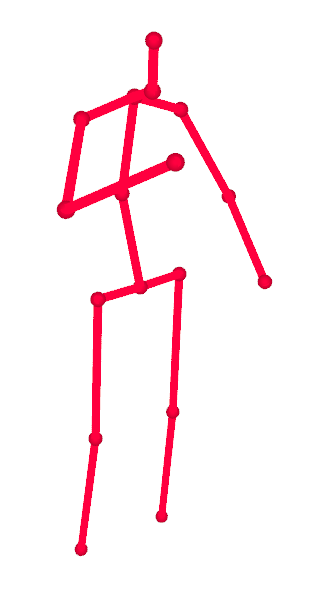}%
}
        \caption{Human3.6M}
    \end{subfigure}%
    \begin{subfigure}{0.22\columnwidth}
        \centering
\maxsizebox{!}{0.2\textheight}{%
        \includegraphics[width=\linewidth]{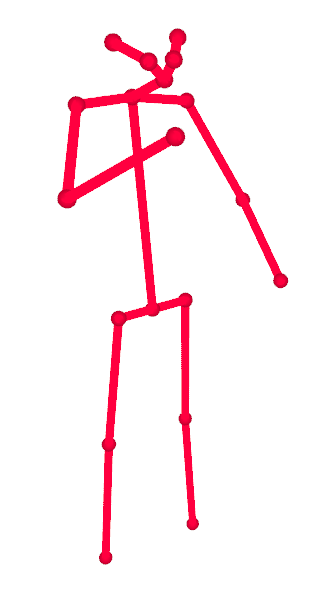}%
}
        \caption{COCO-19}
    \end{subfigure}%
    \begin{subfigure}{0.22\columnwidth}
        \centering
\maxsizebox{!}{0.2\textheight}{%
        \includegraphics[width=\linewidth]{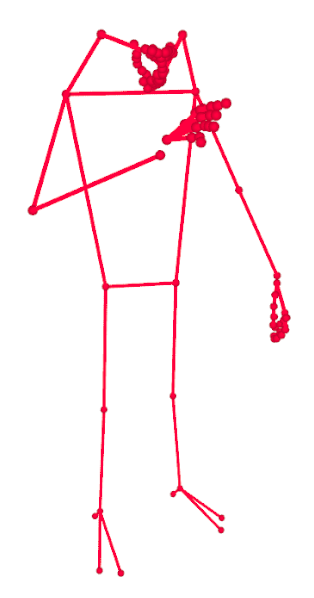}%
        }
        \caption{COCO-133}
    \end{subfigure}%
    \begin{subfigure}{0.22\columnwidth}
        \centering
\maxsizebox{!}{0.2\textheight}{%
        \includegraphics[width=\linewidth]{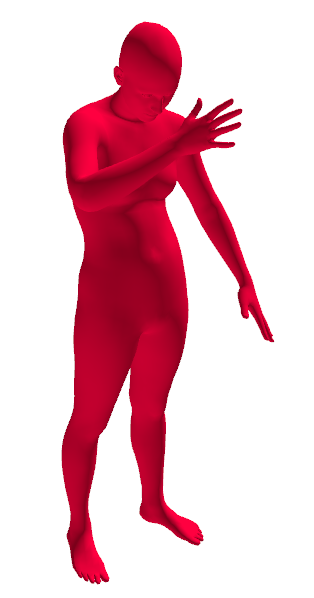}%
}
        \caption{SMPL}
    \end{subfigure}%
    \caption{Examples of different output formats with different 2D keypoint detectors used by Kineo. (a--c) show keypoint-based representations in various formats, while (d) illustrates a full parametric SMPL body reconstruction. For the SMPL model, NLF surface keypoints are first detected and subsequently fitted using the \textit{SMPLFitter} module from NLF.}
    \label{fig:kineo_outputs}
\end{figure}

The 2D keypoints are lifted to 3D space through a triangulation process using a weighted Direct Linear Transform (DLT) method. This process begins by first undistorting the 2D keypoints using the estimated distortion parameters. Subsequently, the undistorted keypoints are triangulated to determine their 3D positions, leveraging information from all $N$ views. Crucially, the triangulation is weighted by the 2D confidence score of each individual keypoint, $w_{i,m}$. This weighting allows the reconstruction to account for varying levels of uncertainty across different views, thereby improving the accuracy and robustness of the final 3D keypoint coordinates. Each frame is triangulated independently: temporal information only enters through the calibration, whose correspondences are drawn across the sequence, so the 3D keypoints inherit the frame-to-frame noise of the 2D detections, which can appear as jitter. As the pipeline takes any 2D keypoints as input, the system can generate multiple output formats to suit diverse downstream applications. Figure~\ref{fig:kineo_outputs} illustrates this flexibility, showing examples of 3D reconstructions in COCO-19~\cite{linMicrosoftCOCOCommon2014,jooPanopticStudioMassively2015}, COCO-133~\cite{jinWholeBodyHumanPose2020}, Human3.6M keypoint formats, as well as a full parametric SMPL body model. The SMPL reconstruction is obtained by first detecting NLF~\cite{sarandiNeuralLocalizerFields2024} surface keypoints and subsequently fitting them using the \textit{SMPLFitter} module from NLF.

\subsection{Pairwise Reprojection Consistency Score}
\label{sec:pairwise_reprojection_confidence_score}
Associating a consistency score to the lifted 3D keypoints is useful for downstream tasks that weight reconstructed points (e.g., filtering, weighted metrics). When reconstructing a 3D point from multiple camera views, not all points are equally reliable. Some may be consistent across views, while others may vary due to noise or occlusions. We propose a score that captures how well a 3D point aligns with the 2D detections across all cameras. Simply put, the score is high when most views agree on where the point should be.

\begin{figure}
\centering
\maxsizebox{\linewidth}{0.3\textheight}{%
\includegraphics[width=0.75\linewidth]{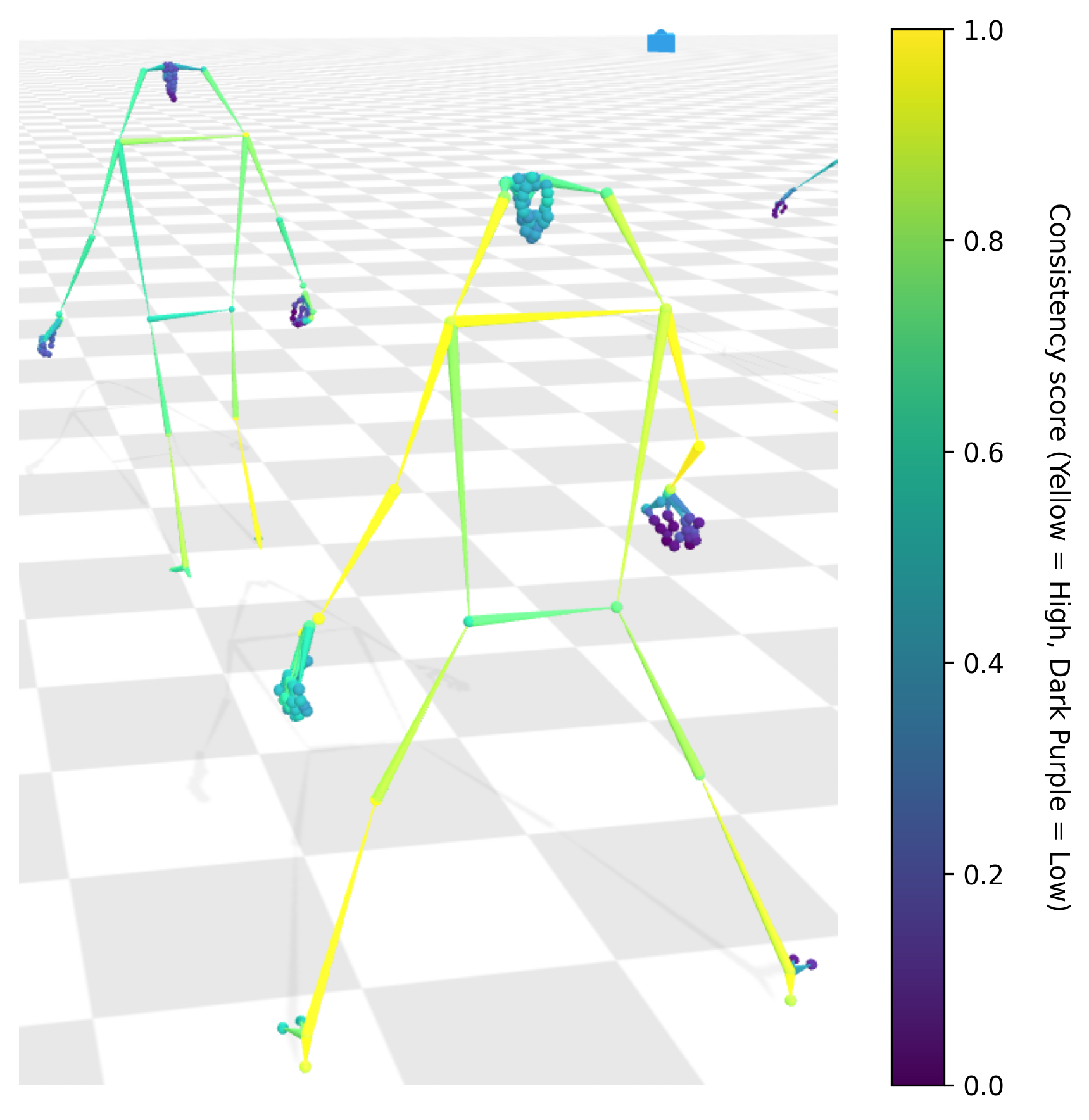}%
}
\caption{Visualization of the proposed \textit{pairwise reprojection consistency score} on 3D human skeleton reconstructions from the \textit{EgoHumans} dataset, obtained using Kineo. The low score on the hands shows that it flags the joints on which the views disagree. Beyond the bone weighting of Sec.~\ref{sec:smpl_scale}, it could be leveraged for downstream filtering, which we leave for future work.}
\label{fig:confidence_score_skeleton}
\end{figure}

For a 3D point $\mathbf{X}_m$, we first measure how well it projects back into each camera $i$ compared to the observed 2D detection $\mathbf{x}_{i,m}$. We normalize the reprojection residual by the camera’s focal length to remove the effect of varying pixel scales, allowing errors to be compared across cameras with different intrinsics:
\begin{equation}
\label{eq:reproj_normalized_residuals}
\tilde{\epsilon}_{i,m}=\frac{\|\pi(K_i,\mathbf{d}_i,[R|\mathbf{t}]_i,\mathbf{X}_m)-\mathbf{x}_{i,m}\|_2}{f_i}
\end{equation}
To transform the residual into a score within the range $[0, 1]$, where higher scores indicate lower residuals, we apply an exponential decay function with rate $\eta$, set to 1:
\begin{equation}
s_{i,m}=\exp(-\eta\tilde{\epsilon}_{i,m})
\end{equation}
We measure agreement between each pair of views using the geometric mean of their scores, weighted by the 2D detection confidences. This ensures that pairs with more reliable 2D points contribute more to the score, reflecting the importance of accurate reprojection for confident detections:
\begin{equation}
    s_{ij,m}=\sqrt{w_{i,m}\cdot w_{j,m}}\sqrt{s_{i,m}\cdot s_{j,m}}
\end{equation}
Finally, we compute the overall consistency score $s_m$ by averaging the scores of all view pairs:
\begin{equation}
s_m=\frac{2}{N(N-1)}\sum^{N-1}_{i=1}\sum^N_{j=i+1}s_{ij,m}
\end{equation}
This gives a final score between 0 and 1: points with high consistency across multiple confident views get a high score, while points on which the views disagree, or with low detection confidence, get a low score. Because the score is based on pairs of views, at least one valid pair is required for the score to be greater than 0. A non-zero score thus implies that the 3D point is triangulated from at least two views, which is a necessary condition for the reconstruction to be meaningful. Qualitative examples of these consistency scores on reconstructed keypoints are shown in Figure~\ref{fig:confidence_score_skeleton}. As a reprojection-consistency heuristic, this score measures how well a 3D point agrees with the confident 2D detections, rather than its 3D error. In our pipeline, it weights the bone lengths of the SMPL-based scale recovery (Sec.~\ref{sec:smpl_scale}). Temporal smoothing of the per-frame keypoints (Sec.~\ref{sec:triangulation}), for instance weighted by this score, is left for future work.

\section{Metric Scale Recovery}
\label{sec:global_scale}

The absolute extrinsics we estimated are defined only up to a global scale, which limits their applicability in downstream tasks requiring metric scale. Unlike methods relying solely on parametric body models to resolve scale ambiguity, which depend on the presence of humans in the scene, we propose two different strategies for global scale recovery: one specific to scenes involving humans and one that is generic and broadly applicable, enabling metric-scale reconstruction even for non-human subjects, at the cost of a lower accuracy.

\subsection{SMPL-based Scale Estimation}
\label{sec:smpl_scale}

\begin{figure*}[t]
    \centering
    \includegraphics[width=\linewidth]{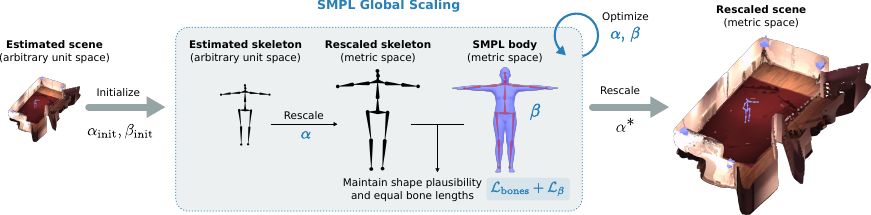}
     \caption{Overview of SMPL-based global scaling. The SMPL body model serves as a prior for metric scale, providing plausible human body proportions that guide the estimation of the global scale factor $\alpha$, which converts the scene from arbitrary units to metric units. Given an estimated skeleton in arbitrary units, we optimize $\alpha$ while updating the SMPL shape parameters $\beta$ so that the SMPL skeleton closely matches the bone lengths of the scaled input skeleton, enforced by $\mathcal{L}_\text{bones}$ while remaining close to the mean shape, enforced by $\mathcal{L}_\beta$. The resulting optimized scale $\alpha^*$ is then applied to rescale the entire scene, including the skeletons, cameras, and scene pointmaps, producing a metric-scale representation.}
    \label{fig:smpl_scaling}
\end{figure*}

The first approach estimates the global scale factor by leveraging the SMPL parametric body model. Conceptually, this approach is related to the method of \citet{mullerReconstructingPeoplePlaces2025}, which incorporates the SMPL model within a global optimization loop to constrain the solution space to metric-consistent configurations. However, our method differs in how the SMPL body prior is used to extract the global scale. Specifically, we treat scale estimation as a downstream task rather than integrating it into a global optimization. This avoids potential conflicts between camera and body shape objectives, reducing the risk of getting stuck in local minima.
Given a skeleton in arbitrary units, we formulate an optimization problem to recover the correct global scale. The goal is to determine the scale factor $\alpha$ that maps the arbitrary-unit skeleton to metric space while matching the bone lengths of the two skeletons by updating the SMPL shape parameters $\beta$, regularized towards the mean shape. This optimization is computationally efficient, as it only requires estimating body shape, ignoring pose, and thus eliminating the need for linear blend skinning. This approach is illustrated in Fig.~\ref{fig:smpl_scaling}.

Formally, the unposed SMPL model is a function ${M(\beta)\in\mathbb{R}^{6890\times3}}$ that maps the shape parameters ${\beta\in\mathbb{R}^{10}}$ to the body vertices. Skeleton keypoints are obtained from the SMPL model using a regressor ${\mathcal{J}\in\mathbb{R}^{K\times6890}}$, yielding ${\hat{\mathbf{j}}(\beta)\in\mathbb{R}^{K\times3}}$:
\begin{equation}
    \hat{\mathbf{j}}(\beta)=\mathcal{J}\,M(\beta)
\end{equation}
Let ${\mathbf{j}^{t}\in\mathbb{R}^{K\times3}}$ be the 3D skeleton keypoints in arbitrary units at frame $t$, and let $\mathcal{B}$ denote the set of bones defined by skeleton connectivity. The length of a bone $(k,l)\in\mathcal{B}$ is the distance between its two keypoints, in the SMPL and in the estimated skeletons:
\begin{equation}
\begin{aligned}
    \hat{\ell}_{kl}(\beta)&=\|\hat{\mathbf{j}}_k(\beta)-\hat{\mathbf{j}}_l(\beta)\|_2 &&\text{(SMPL)}\\
    \ell^t_{kl}&=\|\mathbf{j}^t_k-\mathbf{j}^t_l\|_2 &&\text{(estimated, frame $t$)}
\end{aligned}
\end{equation}

Leveraging the consistency scores of the keypoints estimated in Sec.~\ref{sec:pairwise_reprojection_confidence_score}, we define the confidence $s^t_{kl}$ of each bone as the geometric mean of the scores of its two endpoint keypoints. This formulation ensures that if either joint has a low score, the corresponding bone length is considered unreliable:

\begin{equation}
s^{t}_{kl}=\sqrt{s^{t}_k\cdot s^{t}_l}
\end{equation}

We aggregate the per-frame bone lengths and confidences over the $T$ frames of the sequence into a single length $\ell_{kl}$ and confidence $s_{kl}$ per bone, using a confidence-weighted median and a mean, respectively:
\begin{equation}
\ell_{kl}=\operatorname{med}_\text{w}\!\big(\ell^t_{kl}\big)_{t=1}^{T}
\quad
\text{and}
\quad
s_{kl}=\frac{1}{T}\sum_{t=1}^{T}s^t_{kl}
\end{equation}
where $\operatorname{med}_\text{w}$ is the weighted median, which reduces the influence of frames where the bone is occluded or poorly triangulated.

For a given global scale $\alpha$, these keypoints are mapped to the SMPL model space as $\alpha\mathbf{j}^{t}$, so that the bone lengths of the estimated skeleton become $\alpha\ell_{kl}$. We optimize $\alpha$ and $\beta$ using the L-BFGS optimizer by minimizing a loss that enforces both bone length consistency and plausible body shape:
\begin{equation}
\mathcal{L}_\text{bones}=\frac{\sum_{(k,l)\in\mathcal{B}} s_{kl} \,\mathcal{H}_\delta(\alpha\ell_{kl}-\hat{\ell}_{kl}(\beta))}{\sum_{(k,l)\in\mathcal{B}} s_{kl}}
\quad
\text{and}
\quad
\mathcal{L}_\beta=\|\beta\|_2^2
\end{equation}
where $\mathcal{H}_\delta$ is the Huber loss for robustness to outliers (e.g., bones with excessive length due to occlusion), with $\delta$ set to 0.5\,m. The total loss is expressed as:
\begin{equation}
\mathcal{L}_\text{tot}=\lambda_\text{bones}\mathcal{L}_\text{bones}+\lambda_\beta\mathcal{L}_\beta
\end{equation}
We set $\lambda_\text{bones}=\lambda_\beta=1$ and, as in Sec.~\ref{sec:graph_relative_scales}, solve for the logarithm of $\alpha$ to keep it positive. With multiple subjects, we optimize one $\beta$ per subject and a single $\alpha$ shared by all, as they lie in the same scene; $\mathcal{B}$ then contains the bones of all subjects, and $\mathcal{L}_\beta$ is averaged over subjects.

Since both $\alpha$ and $\beta$ affect the overall size of the skeleton, their joint optimization is ambiguous. The shape prior $\mathcal{L}_\beta$ resolves this ambiguity in favor of the unpenalized $\alpha$ by keeping the SMPL bone lengths close to those of the mean shape. The recovered scale is therefore biased towards the mean SMPL shape, a limitation of SMPL-based scaling that affects taller or shorter subjects~\cite{senguptaSyntheticTrainingAccurate2020, sarandiNeuralLocalizerFields2024}. We nevertheless keep $\beta$ in the optimization so that, in controlled settings, users can provide partial subject information through $\mathcal{L}_\beta$. A known height, for instance, fixes the overall size of the subject, while $\beta$ remains free to fit the observed bone lengths for the proportions that height alone does not determine (arm span, leg-to-torso ratio). The scaling ablation (Sec.~\ref{sec:ablation_study_global_scale}) gives an upper bound on the benefit of such information by fixing $\beta$ to an oracle shape fitted to the ground-truth bone lengths.

To ensure stable convergence, we initialize the SMPL shape parameters $\beta$ to the mean shape, i.e., the null vector, and set the initial scale factor $\alpha_\text{init}$ as the ratio between the SMPL and arbitrary-unit lengths of the most confident bone $(k^*,l^*)$:
\begin{equation}
\beta_\text{init} = \mathbf{0}
\quad
\text{and}
\quad
\alpha_\text{init} = \frac{\hat{\ell}_{k^*l^*}(\beta_\text{init})}{\ell_{k^*l^*}},
\quad
(k^*,l^*)=\operatorname*{arg\,max}_{(k,l)\in\mathcal{B}} s_{kl}
\end{equation}

The final optimized $\alpha^*$ is used to rescale the skeletons, scene pointmaps, and cameras.

\subsection{Metric Depth-based Scale Estimation}
\label{sec:depth_scale}

\begin{figure*}[t]
    \centering
    \includegraphics[width=\linewidth]{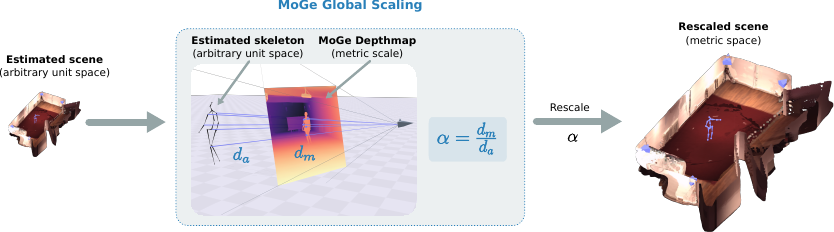}
    \caption{Overview of metric depth-based global scaling. Unlike SMPL-based scaling, which relies on a parametric human body model, metric depth-based scaling is subject-agnostic, as the metric depth map predicted directly from the RGB image by MoGe-2~\cite{wangMoGeUnlockingAccurate2025} is independent of the subjects present in the scene. The global scale factor $\alpha$ is computed as the ratio between the joint depths in metric units obtained from the depth map and the corresponding joint depths in arbitrary units. The resulting scale $\alpha$ is then applied to rescale the entire scene, including the skeletons, cameras, and scene pointmaps, producing a metric-scale representation.}
    \label{fig:moge_scaling}
\end{figure*}

While the previous SMPL-based approach is effective at recovering global scale, it is limited to footage including humans due to its reliance on a parametric body model. We overcome this restriction by using a monocular metric depth estimator, MoGe-2~\cite{wangMoGeUnlockingAccurate2025}, which can estimate a metric depth map from a monocular RGB image, at the cost of a lower accuracy than SMPL-based scaling (Sec.~\ref{sec:ablation_study_global_scale}). Since MoGe-2 recovers more accurate depth when given the horizontal field of view instead of estimating it, we feed it the field of view derived from the focal length estimated in previous stages (see Sec.~\ref{sec:intrinsics_init} and Sec.~\ref{sec:graph_bundle_adjustment}). The global scale $\alpha$ is computed as the ratio between the joint depths in metric units obtained from the depth map and the corresponding joint depths in arbitrary units. This approach is illustrated in Fig.~\ref{fig:moge_scaling}.

Formally, given the camera extrinsics $[R\mid\mathbf{t}]$, and the skeleton keypoints $\mathbf{j}\in\mathbb{R}^{K\times3}$, the keypoint depth corresponds to the $z$ coordinate of the joint in camera space:
\begin{equation}
    \begin{bmatrix}x_c\\y_c\\z_c\end{bmatrix}=[R\mid\mathbf{t}]\cdot\mathbf{j},\quad d_a=z_c
\end{equation}
Given the predicted depth map $D$, we sample the metric depth using the projected keypoint coordinates, using both the intrinsics $K$ and distortion coefficients $\mathbf{d}$:
\begin{equation}
    \begin{bmatrix}u \\ v\end{bmatrix}=\pi(K, \mathbf{d}, [R|\mathbf{t}], \mathbf{j}),\quad d_m=D_{v,u}
\end{equation}
The global scale factor is computed as the ratio between the distance in metric units and the distance in arbitrary units $\alpha=d_m/d_a$. In practice, we select a single frame for each view, choosing the one with the most reliable keypoints. Reliability is quantified by a heuristic score that evaluates how well the 3D keypoints reproject onto the given view:
\begin{equation}
    h_i=\frac{1}{K}\sum^K_{k=1}\exp(-\eta\epsilon_{i,k})
\end{equation}
where $K$ denotes the number of keypoints and $\epsilon_{i,k}$ represents the reprojection residuals as defined in Eq.~\ref{eq:reproj_residuals}. The exponential decay function, with the same rate $\eta$ as in Sec.~\ref{sec:pairwise_reprojection_confidence_score}, transforms the residuals into a score within the interval $[0, 1]$, such that higher values correspond to lower reprojection errors. The frame with the maximum score $h_i$ is selected for each view.

Next, we sample depth values at the keypoint locations in the selected frames, considering only those keypoints whose confidence $w_{i,k}$ exceeds a threshold $\zeta$, set to 0.3. This filtering avoids erroneous depth comparisons in regions where a joint is unlikely to be present. Finally, the global scale factor is estimated as the weighted average of the ratios between metric distances and distances in arbitrary units across all views:

\begin{equation}
\alpha=\frac{\sum_{i=1}^N\sum_{k=1}^K w_{i,k}\,\frac{d_{m,k}}{d_{a,k}}}{\sum_{i=1}^N\sum_{k=1}^K w_{i,k}}.
\end{equation}

The factor $\alpha$ is used to rescale the skeletons, scene pointmaps, and cameras.

\section{Experiments}
\def\usegt{\cellcolor{red!25}GT}
\def\useest{\cellcolor{green!25}Est.}
\def\notused{\cellcolor{orange!25}-}
\newcommand{\NA}{---}
\setlength{\heavyrulewidth}{1.5pt}
\setlength{\lightrulewidth}{0.75pt}
\setlength{\aboverulesep}{0pt}
\setlength{\belowrulesep}{0pt}
\newcommand{\tpad}[1]{\rule{0pt}{#1}}
\definecolor{breeze}{HTML}{dff9fb}

We conduct a comprehensive evaluation of our method through both comparative and ablation studies to assess audio-based temporal synchronization, automatic camera calibration accuracy, 3D reconstruction quality, and scalability to long sequences. We begin by establishing the experimental framework, including datasets and evaluation metrics. We then present quantitative comparisons against state-of-the-art methods, followed by systematic ablation studies to evaluate the contributions to performance of each component in our pipeline. Throughout our evaluation, we demonstrate that our approach achieves substantial improvements over existing calibration-free reconstruction methods while maintaining scalability for practical deployment.

\subsection{Experimental Setup}
\label{sec:experimental_setup}
\paragraph{Datasets} We evaluate our method on three complementary datasets that together provide comprehensive coverage of both controlled and in-the-wild scenarios. Human3.6M~\cite{human36m} provides highly accurate ground-truth annotations from a Vicon motion capture system, making it well-suited for evaluating joint localization accuracy and inference speed in controlled, single-person scenarios. EgoHumans~\cite{egohumans} presents more complex scenarios with both indoor and outdoor settings, multiple interacting subjects, frequent occlusions, and dynamic motions, enabling a thorough assessment of our method's robustness under realistic, unconstrained conditions. Notably, EgoHumans recordings are captured with GoPro cameras that exhibit significant lens distortion, making this dataset particularly valuable for evaluating the quality of our distortion coefficient estimation, a capability that distinguishes our method from prior approaches that typically ignore distortions or undistort the images with the ground-truth distortion coefficients. Finally, we use AIST++~\cite{li2021aistplusplus} to evaluate temporal synchronization. Unlike the other two, it provides unsynchronized multi-view videos with the audio of each camera. We compare our estimated offset $\gamma_k$ of each view $k$ to the reference view with its ground-truth offset $\bar{\gamma}_k$, obtained as detailed in Sec.~\ref{supp-sec:aistpp_gt_offsets} of the supplementary materials.

\paragraph{Evaluation Metrics} We assess both joint position accuracy and camera parameter quality using a comprehensive set of standard metrics. Following \citet{mullerReconstructingPeoplePlaces2025}, the predicted scene is first aligned with the ground truth based on camera centers prior to evaluation. For 3D joint reconstruction, we report the World Mean Per Joint Position Error (W-MPJPE), which quantifies the average Euclidean distance between predicted and ground-truth joint positions in the metric world coordinate system. To isolate local pose errors from global misalignment and scale variations, we also report the Procrustes-Aligned MPJPE (PA-MPJPE), computed per subject and per frame after applying a similarity transform from the predicted skeleton to the ground truth. For camera pose, we report the Translation Error (TE), in meters, and Angular Error (AE), in degrees, to quantify deviations in position and orientation from the ground truth. Complementing these error metrics, we report the Camera Center Accuracy (CCA)~\cite{relpose++}, which measures the proportion of estimated camera centers within predefined distance thresholds relative to the scene scale, and the Relative Rotation Accuracy (RRA), defined as the percentage of cameras whose angular error falls below specified thresholds. To decouple positional errors from global scale, we additionally compute scale-aligned variants by applying a similarity transformation that aligns and uniformly scales the estimated scene to the ground truth, resulting in the scale-aligned Translation Error (s-TE) and scale-aligned Camera Center Accuracy (s-CCA). Finally, we report the Field-of-View error (FoV), in degrees, to quantify deviations between the estimated and ground-truth camera field of view. For temporal synchronization, the offset error of view $k$ is $\gamma_k - \bar{\gamma}_k$, its estimated time offset to the reference view minus the ground-truth offset, in milliseconds. We report the median of its absolute value, and the fraction of views within one, two and four frames of the ground truth (17, 33 and 67\,ms at 59.94\,fps).

\paragraph{Implementation Details} Throughout all experiments, we assume known cross-view re-identification of subjects (as HSfM~\cite{mullerReconstructingPeoplePlaces2025} does) and, except in the runtime and online experiments, take person bounding boxes from the ground truth, focusing our evaluation on the core calibration and reconstruction components. For 2D keypoint detection, we primarily employ NLF as our detector due to its balance of accuracy and efficiency, though we also evaluate alternative detectors in our ablation studies. Unless otherwise specified, we use our SMPL-based global scale estimation and report results without post-processing or filtering applied to the reconstructed joints. Unless stated otherwise, experiments are conducted on a workstation equipped with an NVIDIA A100 GPU and an AMD Milan EPYC 7543 CPU. Table~\ref{supp-tab:implementation_details} of the supplementary materials lists the implementation details and hyperparameters of each stage.

\subsection{Quantitative Evaluation}
\label{sec:quantitative_evaluation}
Following the order of the pipeline, we first evaluate the temporal synchronization of Kineo against ground-truth offsets (Sec.~\ref{sec:quantitative_sync}), then its calibration and 3D reconstruction against state-of-the-art methods (Sec.~\ref{sec:quantitative_calibration}).

\subsubsection{Temporal Synchronization}
\label{sec:quantitative_sync}
We evaluate the temporal synchronization stage (Sec.~\ref{sec:temporal_sync}) on the test sequences of AIST++, using their raw recordings, in which the nine cameras start recording independently. To measure what the residual offset errors cost downstream, we run the full pipeline twice on the same videos, once with the estimated offsets and once with the ground-truth offsets, and compare the reconstruction accuracy (Table~\ref{tab:aistpp_sync}). Our audio-based synchronization estimates the time offsets with a median absolute error of 6.9\,ms, below the one-frame resolution of the ground truth (16.7\,ms at 59.94\,fps), and with 99\% of the views within two frames. These residual offset errors increase the PA-MPJPE by about 2\,mm, from 0.019\,m with the ground-truth offsets to 0.021\,m (Table~\ref{tab:aistpp_sync}). This accuracy also meets the initialization requirements of image-based refinement (Sec.~\ref{sec:temporal_sync}): 99\% of our offsets are within one frame at 10 to 30\,fps (compared to the two to three frames assumed by \citet{voSpatiotemporalBundleAdjustment2020}), indicating that our estimate could initialize such methods.

\begin{table}[!ht]
\centering
\caption{Audio-based temporal synchronization on AIST++ (59.94\,fps). Metrics include the median absolute offset error, the proportion of views within one, two and four frames of the ground-truth offset, and the PA-MPJPE of the full pipeline, in which all camera parameters are estimated. For each row, we indicate whether the time offsets $\gamma_k$ are estimated from the audio (Est.) or taken from the ground truth (GT).}
\maxsizebox{\linewidth}{0.3\textheight}{
\begin{tabular}{c|c|ccc|c}
  \toprule
  & \textbf{Median error} & \multicolumn{3}{c|}{\textbf{Views within (\%)}} & \textbf{PA-MPJPE} \\
  \textbf{Sync $\gamma_k$} & (ms)$\downarrow$ & 1 fr.$\uparrow$ & 2 fr.$\uparrow$ & 4 fr.$\uparrow$ & (m)$\downarrow$ \\
  \midrule
  \usegt & 0.0 & 100 & 100 & 100 & 0.019 \\
  \useest & 6.9 & 89 & 99 & 100 & 0.021 \\
  \bottomrule
\end{tabular}
}
\label{tab:aistpp_sync}
\end{table}

\label{sec:ablation_study_audio_sync}
Finally, we evaluate the robustness of the synchronization to degraded audio (Table~\ref{tab:audio_sync_robustness}; protocol and full results in Sec.~\ref{supp-sec:audio_sync_robustness} of the supplementary materials). We first remove the metronome count-in that opens the raw AIST++ recordings, to measure how much our method relies on its sharp clicks: the proportion of views within one frame drops from 89\% to 72\%, while all views remain within four frames. The remaining music then serves as the baseline for the other conditions. A periodic beep, inserted every 0.5\,s, tests whether a repetitive cue misleads the cross-correlation into an adjacent period. The results show that no estimate shifts to an adjacent beep, as the music disambiguates the beep, and the beep instead recovers most of the accuracy lost with the count-in (87\% of views within one frame). Lastly, we evaluate the impact of background noise on accuracy using recordings from DEMAND~\cite{thiemann2013demand}, park ambient noise for a calm environment and cafeteria babble for a noisier one, mixed at signal-to-noise ratios (SNR) of 20, 10 and 0\,dB with a different excerpt for each view. Accuracy decreases as the noise level increases, more so for babble, with 44\% of views within one frame at 0\,dB (compared to 52\% with park ambient noise). Overall, at least 99\% of the views remain within four frames (67\,ms) of the ground truth in all conditions, whereas the proportion within one frame depends on the presence of sharp transients and on the noise level, which suggests that a transient cue such as a clap within the capture can improve accuracy.

\begin{table}[!ht]
\centering
\caption{Robustness of audio-based synchronization on AIST++. \emph{Count-in removed} keeps only the music and is the baseline for the conditions below it.}
\maxsizebox{\linewidth}{!}{
\begin{tabular}{ll|c|ccc}
    \toprule
    & & \textbf{Median error} & \multicolumn{3}{c}{\textbf{Views within (\%)}} \\
    Audio condition & SNR & (ms)$\downarrow$ & 1 fr.$\uparrow$ & 2 fr.$\uparrow$ & 4 fr.$\uparrow$ \\
    \midrule
  Raw audio (with count-in) & -- & 6.9 & 89 & 99 & 100 \\
  Count-in removed & -- & 9.0 & 72 & 87 & 100 \\
  \quad + Periodic beep & -- & 7.8 & 87 & 96 & 100 \\
  \midrule
  \multirow{3}{*}{\quad + Park ambient} & 20\,dB & 9.0 & 68 & 80 & 100 \\
  & 10\,dB & 12.1 & 62 & 73 & 99 \\
  & 0\,dB & 14.0 & 52 & 63 & 99 \\
  \midrule
  \multirow{3}{*}{\quad + Cafeteria babble} & 20\,dB & 11.3 & 66 & 78 & 100 \\
  & 10\,dB & 12.9 & 58 & 68 & 99 \\
  & 0\,dB & 23.3 & 44 & 54 & 99 \\
  \bottomrule
\end{tabular}
}
\label{tab:audio_sync_robustness}
\end{table}

\subsubsection{Calibration and Reconstruction}
\label{sec:quantitative_calibration}
We benchmark our method against state-of-the-art multi-view reconstruction approaches that rely on different combinations of known and estimated camera parameters. UnCaliPose~\cite{XuUncalipose_2022_BMVC} estimates only the extrinsics and takes the intrinsics as input. We report the version evaluated by HSfM~\cite{mullerReconstructingPeoplePlaces2025}, in which the intrinsics are estimated by DUSt3R~\cite{wangDUSt3RGeometric3D2024} and the metric scale is given by ground-truth bone lengths. In contrast, HAMSt3R~\cite{rojasHamst3r2025} and HSfM~\cite{mullerReconstructingPeoplePlaces2025} initialize both intrinsics and extrinsics from the pointmaps predicted by MASt3R~\cite{leroy2024groundingimagematching3d} and DUSt3R~\cite{wangDUSt3RGeometric3D2024}, respectively, and then refine them by optimization. We also compare to methods evaluated with both known intrinsics and extrinsics on Human3.6M, including AdaFuse~\cite{zhangAdaFuseAdaptiveMultiview2021}, \citet{iskakovLearnableTriangulationHuman2019} and \citet{hewittLookMaNo2024}. Our evaluation considers multiple configurations in which camera parameters are either estimated or provided as ground truth. We also assess performance both with and without modeling lens distortions to quantify their impact on reconstruction accuracy. Tables~\ref{tab:human_metrics} and~\ref{tab:camera_metrics} report the human reconstruction and camera metrics, respectively. We evaluate HSfM accuracy on Human3.6M using the official code released on the authors' repository, ensuring evaluation conditions consistent with those used for EgoHumans. For full transparency, inference scripts are available at \url{https://github.com/cjaverliat/hsfm}, and the evaluation code is included in Kineo's repository.

\begin{table}[htbp]
\centering
\caption{Quantitative comparison of human pose estimation on EgoHumans~\cite{egohumans} and Human3.6M~\cite{human36m}. Metrics include mean-per-joint position error in metric world space (W-MPJPE, in meters) and its Procrustes-Aligned version (PA-MPJPE, in meters). For each method, we indicate whether camera parameters are estimated, taken from ground truth, or omitted, where $[R|\mathbf{t}]$ denotes the extrinsic camera parameters, $K$ the intrinsic camera matrix (focal length and principal point), and $\mathbf{d}$ the distortion coefficients. For each dataset, we highlight the \textbf{best} and \underline{second-best} \textit{calibration-free} methods. For both datasets, we use the temporal synchronization provided with the dataset rather than estimating it due to the lack of audio. \textbf{Note:} For HSfM on Human3.6M, results were obtained through our own evaluation, while other values are reported as in the respective papers. UnCaliPose values are those reported by \citet{mullerReconstructingPeoplePlaces2025}.}
\maxsizebox{\linewidth}{!}{
    \begin{tabular}{ll|cc|ccc}
        \toprule
        & \multirow{2}{*}{\textbf{Method}} & \multicolumn{2}{c|}{\textbf{Human Metrics}} & \multicolumn{3}{c}{\textbf{Parameters}} \\
        & & W-MPJPE$\downarrow$ & PA-MPJPE$\downarrow$ & $[R|\textbf{t}]$ & $K$ & $\mathbf{d}$ \\
        \midrule
        \multirow{6}{*}{\rotatebox{90}{\small EgoHumans}}
        & UnCaliPose~\cite{XuUncalipose_2022_BMVC} & 3.51 & 0.13 & \useest & \cellcolor{green!25}Est. & \cellcolor{orange!25}- \\
        & HSfM~\cite{mullerReconstructingPeoplePlaces2025} & \underline{1.04} & \underline{0.05} & \useest & \useest & \notused \\
        & HAMSt3R~\cite{rojasHamst3r2025} & 3.80 & 0.14 & \useest & \useest & \notused \\
& \cellcolor{breeze}{\textbf{Kineo}} & 0.16 & 0.02 & \useest & \usegt & \usegt \\
& \cellcolor{breeze}{\textbf{Kineo}} & 0.41 & 0.03 & \useest & \useest & \notused \\
& \cellcolor{breeze}{\textbf{Kineo}} & \textbf{0.17} & \textbf{0.02} & \useest & \useest & \useest \\      
        \midrule
        \multirow{7}{*}{\rotatebox{90}{Human3.6M}}
        & \citet{iskakovLearnableTriangulationHuman2019} & 0.02 & - & \usegt & \usegt & \usegt \\
        & AdaFuse~\cite{zhangAdaFuseAdaptiveMultiview2021} & 0.02 & - & \usegt & \usegt & \usegt \\
        & \citet{hewittLookMaNo2024} & 0.03 & - & \usegt & \usegt & \notused \\
& HSfM & \underline{0.47} & \underline{0.05} & \useest & \useest & \notused \\
& \cellcolor{breeze}{\textbf{Kineo}} & 0.02 & 0.02 & \useest & \usegt & \usegt \\
& \cellcolor{breeze}{\textbf{Kineo}} & 0.04 & 0.02 & \useest & \useest & \notused \\
& \cellcolor{breeze}{\textbf{Kineo}} & \textbf{0.04} & \textbf{0.02} & \useest & \useest & \useest \\
        \bottomrule
    \end{tabular}
}
\label{tab:human_metrics}
\end{table}
\begin{table*}[htbp]
\centering
\caption{Quantitative comparison of camera pose estimation on EgoHumans~\cite{egohumans} and Human3.6M~\cite{human36m}. Metrics include translation error (TE, in meters), scale-aligned translation error (s-TE, in meters), angular error (AE, in degrees), Camera Center Accuracy (CCA, as a proportion), scale-aligned CCA (s-CCA, as a proportion), and Relative Rotation Accuracy (RRA, as a proportion), at thresholds of 10 and 15 degrees for RRA and of 10 and 15\% of the scene scale for CCA, and the field-of-view error (FoV, in degrees). For each dataset we report the \textbf{best} and \underline{second-best} \textit{calibration-free} methods. For both datasets, we use the temporal synchronization provided with the dataset rather than estimating it due to the lack of audio. \textbf{Note:} For HSfM on Human3.6M, results were obtained through our own evaluation, while other values are reported as in the respective papers. UnCaliPose, DUSt3R and MASt3R values are those reported by \citet{mullerReconstructingPeoplePlaces2025}.}
\maxsizebox{\linewidth}{!}{
    \begin{tabular}{ll|cccccccccc|ccc}
        \toprule
        & \multirow{2}{*}{\textbf{Method}} & \multicolumn{10}{c|}{\textbf{Camera Metrics}} & \multicolumn{3}{c}{\textbf{Parameters}} \\
        & & TE$\downarrow$ & s-TE$\downarrow$ & AE$\downarrow$ & RRA@10$\uparrow$ & RRA@15$\uparrow$ & CCA@10$\uparrow$ & CCA@15$\uparrow$ & s-CCA@10$\uparrow$ & s-CCA@15$\uparrow$ & FoV$\downarrow$ & $[R|\textbf{t}]$ & $K$ & $\mathbf{d}$ \\
        \midrule
        \multirow{8}{*}{\rotatebox{90}{EgoHumans}}
        & UnCaliPose~\cite{XuUncalipose_2022_BMVC} & 2.63 & 2.63 & 60.90 & 0.28 & 0.39 & - & - & 0.33 & 0.44 & - & \useest & \cellcolor{green!25}Est. & \cellcolor{orange!25}- \\
        & DUSt3R~\cite{wangDUSt3RGeometric3D2024} & - & 1.15 & 11.00 & 0.61 & 0.74 & - & - & 0.49 & 0.74 & - & \useest & \useest & \notused \\
        & MASt3R~\cite{leroy2024groundingimagematching3d} & 4.97 & 0.92 & 10.42 & 0.61 & 0.74 & 0.06 & 0.07 & 0.65 & 0.86 & - & \useest & \useest & \notused \\
        & HSfM~\cite{mullerReconstructingPeoplePlaces2025} & \underline{2.09} & 0.75 & \underline{9.35} & 0.72 & \underline{0.89} & \underline{0.32} & \underline{0.46} & \underline{0.75} & \underline{0.91} & - & \useest & \useest & \notused \\
        & HAMSt3R~\cite{rojasHamst3r2025} & 2.33 & \underline{0.40} & 10.24 & \underline{0.77} & - & 0.06 & - & \underline{0.75} & - & - & \useest & \useest & \notused \\
& \cellcolor{breeze}{\textbf{Kineo}} & 0.29 & 0.05 & 0.34 & 1.00 & 1.00 & 0.99 & 1.00 & 0.99 & 1.00 & 0.00 & \useest & \usegt & \usegt \\
& \cellcolor{breeze}{\textbf{Kineo}} & 0.76 & 0.57 & 3.48 & 0.91 & 0.98 & 0.72 & 0.90 & 0.89 & 0.98 & 2.96 & \useest & \useest & \notused \\
& \cellcolor{breeze}{\textbf{Kineo}} & \textbf{0.34} & \textbf{0.15} & \textbf{0.69} & \textbf{1.00} & \textbf{1.00} & \textbf{0.98} & \textbf{0.99} & \textbf{0.99} & \textbf{1.00} & \textbf{1.03} & \useest & \useest & \useest \\
\midrule
\multirow{4}{*}{\rotatebox{90}{\small H3.6M}}
& HSfM~\cite{mullerReconstructingPeoplePlaces2025} & 0.83 & 0.33 & 6.44 & 0.95 & 0.96 & 0.46 & 0.74 & 0.95 & 0.95 & 36.97 & \useest & \useest & \notused \\
& \cellcolor{breeze}{\textbf{Kineo}} & 0.01 & 0.01 & 0.20 & 1.00 & 1.00 & 1.00 & 1.00 & 1.00 & 1.00 & 0.00 & \useest & \usegt & \usegt \\
& \cellcolor{breeze}{\textbf{Kineo}} & 0.13 & 0.03 & 0.90 & 1.00 & 1.00 & 1.00 & 1.00 & 1.00 & 1.00 & 0.57 & \useest & \useest & \notused \\
& \cellcolor{breeze}{\textbf{Kineo}} & \textbf{0.12} & \textbf{0.02} & \textbf{0.89} & \textbf{1.00} & \textbf{1.00} & \textbf{1.00} & \textbf{1.00} & \textbf{1.00} & \textbf{1.00} & \textbf{0.43} & \useest & \useest & \useest \\
        \bottomrule
    \end{tabular}
}
\label{tab:camera_metrics}
\end{table*}

\begin{figure}[!ht]
\centering
\maxsizebox{!}{0.4\textheight}{%
\includegraphics[width=\linewidth]{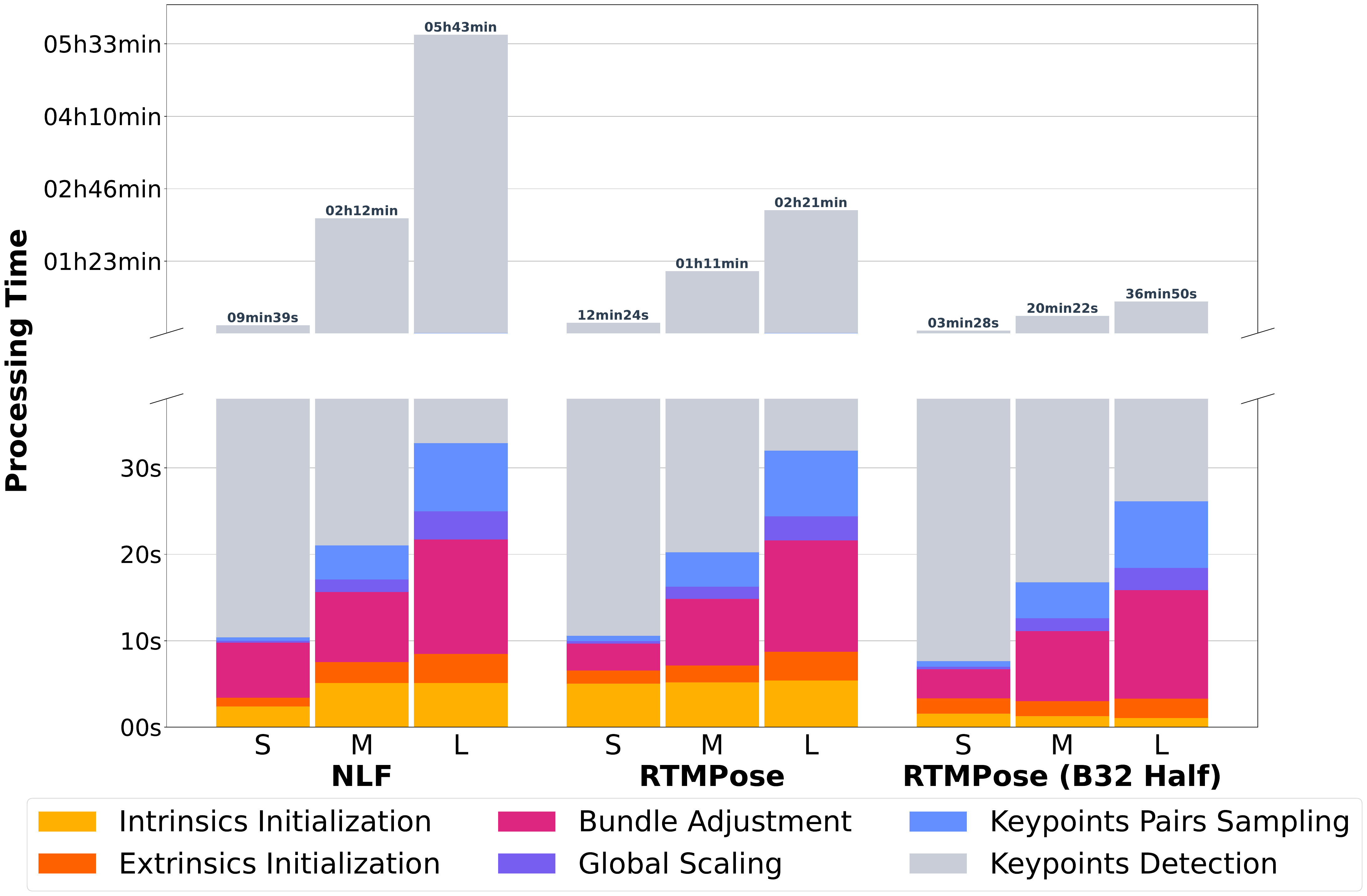}%
}
\caption{Automatic calibration runtimes for different keypoint detectors across sequences of varying durations. The comparison includes three keypoint sources: NLF (batch size 1, full precision), RTMPose (batch size 1, full precision), and RTMPose (batch size 32, half precision). Sequence sizes are small (S, 7\,min\,10\,s total across 4 views), medium (M, 43\,min\,03\,s total across 4 views), and long (L, 1\,h\,23\,min total across 4 views). Each stacked bar breaks down the runtime by processing stage: Intrinsics Initialization (yellow), Extrinsics Initialization (orange), Bundle Adjustment (pink), Global Scaling (purple), Keypoints Pairs Sampling (blue), and Keypoints Detection (gray). Total runtime is indicated above each bar. RTMPose (B32 Half) consistently achieves the fastest processing times, completing all stages in less time than the total duration of the original video sequences. Importantly, the calibration runtime scales efficiently, showing only moderate increases as sequence duration grows.}
\label{fig:runtime_metrics}
\end{figure}

Kineo achieves substantial improvements in both camera estimation and 3D human reconstruction across datasets. On EgoHumans, it surpasses the strongest prior calibration-free baseline, HSfM~\cite{mullerReconstructingPeoplePlaces2025}. In the fully calibration-free setting, where extrinsics, intrinsics, and distortion parameters are all estimated, Kineo achieves a camera translation error of 0.34 m (compared to 2.09 m for HSfM and 2.33 m for HAMSt3R) and an angular error of 0.69$^\circ$ (compared to 9.35$^\circ$ for HSfM and 10.24$^\circ$ for HAMSt3R), while achieving a vertical field-of-view error of 1.03$^\circ$. These improvements in camera estimation directly enhance human reconstruction quality, attaining a W-MPJPE of 0.17 m (compared to 1.04 m for HSfM and 3.80 m for HAMSt3R) and a PA-MPJPE of 2 cm (compared to 5 cm for HSfM and 14 cm for HAMSt3R).
A similar trend is observed on Human3.6M, where Kineo again delivers consistent gains over previous methods. Under the same calibration-free conditions, Kineo achieves a camera translation error of 0.12 m (compared to 0.83 m for HSfM) and an angular error of 0.89$^\circ$ (compared to 6.44$^\circ$ for HSfM). The field-of-view error is similarly reduced to 0.43$^\circ$ (compared to 36.97$^\circ$ for HSfM). These improvements in camera estimation directly translate to enhanced human reconstruction quality, yielding a W-MPJPE of 0.04 m (compared to 0.47 m for HSfM) and a PA-MPJPE of 2 cm (compared to 5 cm for HSfM).

To better understand the impact of estimating each camera parameter, we also examined intermediate configurations of Kineo. Providing ground-truth intrinsics, which can often be obtained from manufacturer specifications or through pre/post-capture calibration, consistently improves both camera and human metrics, as expected. On Human3.6M, for instance, access to known intrinsics reduces translation and angular errors from 0.12 m and 0.89$^\circ$ to 0.01 m and 0.20$^\circ$, and halves the W-MPJPE from 0.04 m to 0.02 m. This performance is competitive with that of fully calibrated methods such as AdaFuse~\cite{zhangAdaFuseAdaptiveMultiview2021} and \citet{iskakovLearnableTriangulationHuman2019}, indicating that Kineo attains accuracy on par with calibrated methods even in the absence of ground-truth extrinsics.
Most notably, modeling lens distortion proves critical, especially for EgoHumans, where the wide-angle GoPro cameras introduce significant distortions. When distortion parameters are estimated, Kineo's camera translation error decreases from 0.76 m to 0.34 m and angular error from 3.48$^\circ$ to 0.69$^\circ$, with corresponding improvements in reconstruction accuracy, W-MPJPE from 0.41 m to 0.17 m and PA-MPJPE from 0.03 m to 0.02 m. These results highlight that modeling distortions is essential for achieving high-fidelity calibration-free human reconstruction in challenging real-world capture conditions. We evaluate the accuracy of the estimated distortion in Sec.~\ref{sec:ablation_study_distortion}.

In addition to its accuracy gains, Kineo is computationally efficient and scalable. Figure~\ref{fig:runtime_metrics} reports automatic calibration runtimes for different pipeline stages across sequences of varying lengths and four views using multiple keypoint detectors. In this offline setting, the camera parameters are estimated from the full sequence before reconstruction. We evaluated three sequences of increasing length: a small sequence (S) typical of standard datasets, approximately 2 minutes in duration; a medium sequence (M) of roughly 11 minutes; and a long sequence (L) of approximately 21 minutes. Three keypoint detection configurations were tested: NLF~\cite{sarandiNeuralLocalizerFields2024}, RTMPose~\cite{rtmpose} with a batch size of 1 and full precision, and RTMPose with a batch size of 32 and half precision. In all configurations, person boxes are detected with RTMDet-tiny~\cite{lyuRTMDetEmpiricalStudy2022}. These calibration experiments were conducted with SMPL-based global scaling enabled on a single NVIDIA A100 GPU. Across all configurations, keypoint detection is the most time-consuming stage, while the remaining calibration stages scale moderately with sequence length. These results indicate that Kineo's calibration pipeline is both efficient and scalable, making it suitable for long-duration, real-world captures. For example, when processing a 21-minute, four-view recording using the RTMPose model for keypoint detection (half precision, batch size = 32), the total processing time is only $36$ minutes, substantially shorter than the cumulative duration of the input footage ($1~\text{h}~23~\text{min}$).

\subsubsection{Distortion Accuracy}
\label{sec:ablation_study_distortion}

\begin{figure*}[htbp]
  \centering
  \includegraphics[width=\textwidth]{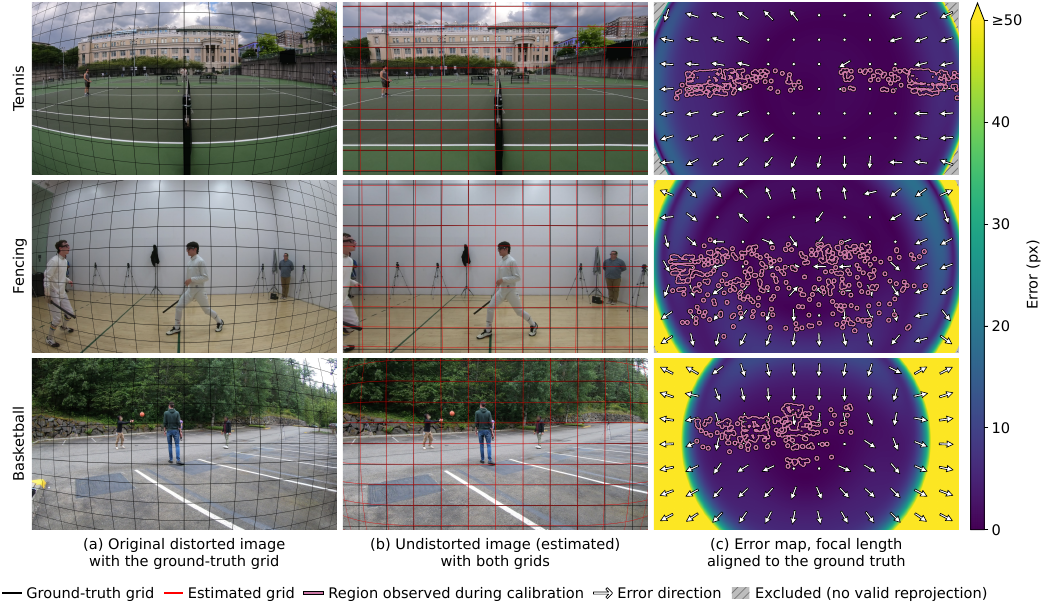}
  \caption{Estimated distortion on three EgoHumans~\cite{egohumans} views. (a)~Input frame with a grid distorted by the ground-truth camera. (b)~Frame and grid undistorted with the estimated camera (red), overlaid on the ground-truth grid (black). (c)~Error of the estimated distortion after focal length rescaling (Sec.~\ref{sec:ablation_study_distortion}), with arrows showing its direction and pink contours outlining the observed region.}
  \label{fig:distortion_error_maps}
\end{figure*}

We evaluate the estimated lens distortion against the ground-truth calibration of EgoHumans, whose wide-angle GoPro cameras exhibit significant distortion. As it is optimized during bundle adjustment (Sec.~\ref{sec:graph_bundle_adjustment}), the distortion is constrained by the sampled keypoints, which only cover the image region where the subjects move, i.e., the observed region, outlined in pink in Fig.~\ref{fig:distortion_error_maps}. Although the shared intrinsics let all views contribute, this region covers only 4.9\% of the image for the median view, which underconstrains the rest of the view. The observed region is, however, the one that matters for human reconstruction, since the detected keypoints lie within it and are undistorted prior to triangulation (Sec.~\ref{sec:triangulation}). We therefore report the error separately in the observed region and in the remainder of the image, where the distortion is extrapolated. Similarly to the reprojection error of AnyCalib~\cite{AnyCalib}, we measure it as the pixel displacement between the ground-truth and estimated cameras over a uniform grid, after rescaling the estimated focal length to separate the distortion error from the focal length error (details in Sec.~\ref{supp-sec:egohumans} of the supplementary materials). Over all views, the median error is 2.0\,px in the observed region. As Fig.~\ref{fig:distortion_error_maps} shows, it grows outside this region, especially towards the image borders beyond the largest distance from the image center that this region reaches, where it can exceed 50\,px (basketball). The estimated distortion is thus accurate where the keypoints are triangulated, but less so for tasks that undistort the whole image, such as our scene pointmaps (Sec.~\ref{sec:scene_pointmaps}).

\subsection{Ablation Study}
\label{sec:ablation_study}

\subsubsection{Impact of Camera Parameter Estimation}

We conduct an ablation study to assess the impact of estimating different subsets of camera parameters on human and camera metrics. The results are presented in Table~\ref{tab:ablation_study_camera_parameters}. Ground-truth 2D human keypoints and global scale are used to eliminate the influence of the 2D detector and the global scale estimation strategy, which are evaluated separately in Sec.~\ref{sec:ablation_study_detectors} and Sec.~\ref{sec:ablation_study_global_scale}.

\begin{table}[!ht]
\centering
\caption{Ablation study for different combinations of ground truth (GT), estimated (Est.) parameters and omitted parameters, evaluated on Human3.6M and EgoHumans datasets, with W-MPJPE, PA-MPJPE, TE and s-TE in meters, and AE and FoV in degrees. We isolate the effect of camera parameter estimation by using ground-truth keypoints to simulate an ideal detector and ground-truth global scale.}
\maxsizebox{\linewidth}{0.3\textheight}{
\begin{tabular}{lcc|cccc|ccc}
    \toprule
    & \multicolumn{2}{c|}{\textbf{Human Metrics}} & \multicolumn{4}{c|}{\textbf{Camera Metrics}} & \multicolumn{3}{c}{\textbf{Parameters}} \\
    & W-MPJPE$\downarrow$ & PA-MPJPE$\downarrow$ & TE$\downarrow$ & s-TE$\downarrow$ & AE$\downarrow$ & FoV$\downarrow$ & $[R\mid\textbf{t}]$ & $K$ & $\mathbf{d}$ \\
    \midrule
    \multirow{4}{*}{EgoHumans}
& 0.006 & 0.002 & 0.000 & 0.000 & 0.000 & 0.000 & \usegt & \usegt & \usegt \\
& 0.011 & 0.003 & 0.010 & 0.010 & 0.092 & 0.000 & \useest & \usegt & \usegt \\
& 0.311 & 0.024 & 0.658 & 0.658 & 3.813 & 2.776 & \useest & \useest & \notused \\
& 0.098 & 0.007 & 0.196 & 0.196 & 0.993 & 1.531 & \useest & \useest & \useest \\
    \midrule
    \multirow{4}{*}{Human3.6M}
& 0.000 & 0.000 & 0.000 & 0.000 & 0.000 & 0.000 & \usegt & \usegt & \usegt \\
& 0.000 & 0.000 & 0.000 & 0.000 & 0.002 & 0.000 & \useest & \usegt & \usegt \\
& 0.014 & 0.002 & 0.020 & 0.020 & 0.894 & 0.380 & \useest & \useest & \notused \\
& 0.006 & 0.001 & 0.011 & 0.011 & 0.863 & 0.243 & \useest & \useest & \useest \\
    \bottomrule
\end{tabular}
}
\label{tab:ablation_study_camera_parameters}
\end{table}

When all camera parameters are fixed to their ground-truth values, the reported metrics reflect only the residual errors arising from the triangulation process. In this configuration, the pipeline achieves minimal errors in both human and camera metrics, with PA-MPJPE and W-MPJPE approaching zero. It is important to note that our method relies on a simple DLT triangulation, which is not optimal due to its linear approximations; while these residual errors could potentially be reduced further by applying an additional optimization step on the keypoint reprojection error, we did not perform this step for efficiency, as we considered the residual error to be already sufficiently low.

When the intrinsics and distortion coefficients are ground-truth and only the extrinsic parameters are estimated, the resulting metrics reflect the residual errors introduced by the SfM initialization and the first bundle adjustment stage, which optimizes the camera poses. Expanding the estimation to include both extrinsic and intrinsic parameters, while fixing the distortion coefficients to zero (effectively omitting distortion modeling), reveals additional residual errors. These errors arise from the intrinsic parameter initialization and the second bundle adjustment stage, which optimizes the focal length. However, the most significant source of residual error in this configuration is the absence of distortion modeling. This becomes evident when distortions are introduced as an additional parameter to be estimated in the next configuration, leading to a substantial reduction in the overall reconstruction error. For instance, the world mean per joint position error (W-MPJPE) decreases by $\approx$68\% on the EgoHumans dataset and $\approx$57\% on the Human3.6M dataset.

\subsubsection{Impact of 2D Keypoint Detector}
\label{sec:ablation_study_detectors}

Since our method depends on 2D keypoint detections, we assess how the choice of detector affects human reconstruction accuracy through the PA-MPJPE. In this study, all camera parameters are estimated, creating a scenario in which no prior information is provided and the only variable is the selected 2D keypoint detector. The results of this evaluation are reported in Table~\ref{tab:ablation_study_detectors}.
\begin{table}[!ht]
\centering
\caption{Impact of 2D detector on PA-MPJPE (in meters, lower is better). All camera parameters are estimated (extrinsics, intrinsics, distortion coefficients).}
\maxsizebox{0.8\linewidth}{!}{
\begin{tabular}{ll|c}
    \toprule
    & \textbf{Keypoints} & \textbf{PA-MPJPE$\downarrow$} \\
    \midrule
    \multirow{3}{*}{EgoHumans}
& Ground-truth & 0.007 \\
& NLF~\cite{sarandiNeuralLocalizerFields2024} & 0.019 \\
& DWPose~\cite{dwpose} & 0.024 \\
    \midrule
    \multirow{3}{*}{Human3.6M}
& Ground-truth & 0.001 \\
& NLF~\cite{sarandiNeuralLocalizerFields2024} & 0.019 \\
& RTMPose~\cite{rtmpose} & 0.022 \\
    \bottomrule
\end{tabular}
}
\label{tab:ablation_study_detectors}
\end{table}

The choice of detector significantly influences reconstruction performance. As expected, ground-truth keypoints yield the lowest PA-MPJPE, establishing a performance upper bound. Notably, the PA-MPJPE error drops to 1 mm on Human3.6M and 7 mm on EgoHumans when ground-truth keypoints are used.

When comparing detectors, NLF~\cite{sarandiNeuralLocalizerFields2024} consistently outperforms DWPose~\cite{dwpose} on the EgoHumans dataset and RTMPose~\cite{rtmpose} on Human3.6M. This demonstrates that, while NLF incurs a higher computational cost, it yields lower errors in human pose estimation. This flexibility in the choice of the keypoint detector allows Kineo to balance the trade-off between accuracy and efficiency based on the specific use case: whether prioritizing maximal accuracy with a more resource-intensive model or opting for a lightweight, faster detector to achieve real-time inference of 2D keypoints.

These results highlight a key advantage of our modular approach: as 2D keypoint detection technology evolves, Kineo can seamlessly integrate advancements without requiring retraining. In contrast, end-to-end models would demand a complete retraining pipeline to incorporate new architectures or improvements. Thus, Kineo is well-positioned to leverage future advancements in 2D keypoint detection.

\subsubsection{Impact of Sampling Strategy}
\label{sec:ablation_study_sampling}
This ablation evaluates whether the spatio-temporal coverage that our subsampling obtains implicitly, by selecting at random among the confident candidates of all frames (Sec.~\ref{sec:sampling}), is sufficient for an accurate calibration. We compare it with two alternatives that select the same number of correspondences among the same confidence-filtered candidates: farthest-point sampling (FPS), which enforces this coverage explicitly, and confidence-only sampling (Top-P), which keeps the most confident candidates regardless of where and when they lie. All runs use NLF on EgoHumans. We measure the coverage on the image region where the subjects are detected, that is, the convex hull of all confidence-filtered candidates, and report the results in Table~\ref{tab:sampling_egohumans}. Sec.~\ref{supp-sec:sampling_categories} of the supplementary materials details both alternatives and gives these results per category (Table~\ref{supp-tab:sampling_egohumans_categories}) and on Human3.6M (Table~\ref{supp-tab:sampling_h36m}).
\begin{table}[!t]
\centering
\caption{Sampling strategies on EgoHumans~\cite{egohumans}: farthest-point sampling (FPS), confidence-only sampling (Top-P) and our random subsampling, with the same budget ($P=3000$, $M=500$) and the same confidence-filtered candidates ($\tau=\kappa=0.6$). Coverage is the area of the convex hull of the selected correspondences over that of the candidates. Medians over the sequences reconstructed by all runs; the best is in bold and the second best underlined. TE, s-TE, W-MPJPE and PA-MPJPE are in meters, and AE in degrees. Runtime covers the two sampling stages.}
\maxsizebox{\linewidth}{!}{
    \begin{tabular}{l|c|ccc|cc|c}
        \toprule
        \multirow{2}{*}{\textbf{Sampling}} & \multirow{2}{*}{\textbf{Coverage (\%)}$\uparrow$} & \multicolumn{3}{c|}{\textbf{Camera Metrics}} & \multicolumn{2}{c|}{\textbf{Human Metrics}} & \multicolumn{1}{c}{\textbf{Runtime (s)}} \\
        & & TE$\downarrow$ & s-TE$\downarrow$ & AE$\downarrow$ & W-MPJPE$\downarrow$ & PA-MPJPE$\downarrow$ & Sampling$\downarrow$ \\
        \midrule
        FPS & \textbf{99} & \textbf{0.220} & \textbf{0.141} & \textbf{0.46} & \textbf{0.099} & 0.016 & 5.28 \\
        Top-P & 46 & 0.256 & 0.171 & 0.68 & 0.111 & 0.018 & \textbf{3.39} \\
        Random (ours) & \underline{94} & \underline{0.229} & \underline{0.148} & \underline{0.47} & \underline{0.101} & 0.016 & \underline{3.42} \\
        \bottomrule
    \end{tabular}
}
\label{tab:sampling_egohumans}
\end{table}%

Random sampling covers a median of 94\% of this region, close to the 99\% of FPS, whereas Top-P covers only 46\%. The calibration errors follow the same order: the median errors of random sampling are within 1\,cm and 0.01\textdegree{} of FPS, with the same median PA-MPJPE, whereas Top-P raises the median angular error from 0.47 to 0.68\textdegree{}. Tennis is the only category with a large difference, where FPS reduces the scale-aligned translation error from 0.134 to 0.040\,m and Top-P is far less accurate. On Human3.6M with RTMPose, Top-P is again the least accurate, and FPS gains only on the metric translation error.

The coverage obtained implicitly by our random sampling strategy is thus sufficient in all categories but tennis, and enforcing it explicitly makes sampling 1.5 times slower for a marginal gain. We therefore use random sampling in all experiments. However, none of these strategies can select correspondences outside the region where the subjects move, so the image borders, where lens distortion is strongest, remain weakly constrained (Sec.~\ref{sec:limitations}).

\subsubsection{Impact of Keypoint Confidence}
\label{sec:ablation_study_confidence}
The detection confidence is used in our pipeline in two ways. As filters, it discards correspondences with a pair score below $\tau$, and keeps for bundle adjustment only the keypoints scoring above $\kappa$ in at least two views. As weights, it reduces the influence of unreliable keypoints in bundle adjustment, triangulation and scale estimation. We measure the effect of each use separately: we disable both filters while keeping the weights, then disable the weights while keeping both filters. The experiments use RTMPose on Human3.6M, both on the original images and with synthetic occlusions~\cite{bragagnoloMultiviewPoseFusion2024} from PASCAL VOC objects~\cite{everinghamPascalVisualObject2010}, a harder setting with more unreliable keypoints. Table~\ref{supp-tab:confidence_weights_full} of the supplementary materials gives the effect of the weights with other detectors and on EgoHumans.

\paragraph{Confidence filters} Table~\ref{tab:sampling_confidence_filter_h36m} compares the pipeline with and without the filters; without them, correspondences and bundle adjustment keypoints are drawn uniformly at random among all candidates. The filters give RANSAC a cleaner input, especially under occlusion, where they discard 28.0\% of the candidate correspondences: the share of sampled correspondences that agree with the ground-truth epipolar geometry rises from 67.3 to 79.3\%, and the inlier ratio from 61.2 to 71.9\%. The adaptive stopping criterion of RANSAC therefore ends the search after fewer iterations in every sequence (a median of 25 instead of 56 under occlusion, and 13 instead of 18 on the original images), with similar final accuracy.
\begin{table}[!hb]
\centering
\caption{Effect of the confidence filters ($\tau=\kappa=0.5$, $P=4000$, $M=400$) on Human3.6M~\cite{human36m} with RTMPose, on the original images and with synthetic occlusions~\cite{bragagnoloMultiviewPoseFusion2024}, with the weights kept. Removed is the mean percentage of candidates discarded, and GT-consistent the percentage of sampled correspondences within the RANSAC threshold of the ground-truth epipolar geometry. RANSAC values are medians over camera pairs, and metrics medians over the sequences reconstructed by all Human3.6M runs with RTMPose.}
\newcolumntype{H}{>{\setbox0\hbox\bgroup}c<{\egroup}@{}}
\setlength{\tabcolsep}{3pt}
\maxsizebox{\linewidth}{!}{
    \begin{tabular}{ll|cc|ccc|HHcHcHcH|HHHHcH}
        \toprule
        \multirow{2}{*}{\textbf{Input}} & \multirow{2}{*}{\textbf{Filter}} & \multicolumn{2}{c|}{\textbf{Removed}} & \multicolumn{3}{c|}{\textbf{RANSAC}} & \multicolumn{8}{c|}{\makecell[b]{\textbf{Camera}\\\textbf{Metrics}}} & \multicolumn{6}{c}{\makecell[b]{\textbf{Human}\\\textbf{Metrics}}} \\
        & & \makecell[t]{Corr.\\(\%)} & \makecell[t]{BA\\(\%)} & \makecell[t]{Inliers\\(\%)$\uparrow$} & \makecell[t]{GT-consistent\\(\%)$\uparrow$} & Iter.$\downarrow$ & \makecell[t]{TE\\(m)$\downarrow$} & p90 & \makecell[t]{s-TE\\(m)$\downarrow$} & p90 & \makecell[t]{AE\\(\textdegree)$\downarrow$} & p90 & \makecell[t]{FoV\\(\textdegree)$\downarrow$} & p90 & \makecell[t]{W-MPJPE\\(m)$\downarrow$} & p90 & \makecell[t]{GA-MPJPE\\(m)$\downarrow$} & p90 & \makecell[t]{PA-MPJPE\\(m)$\downarrow$} & p90 \\
        \midrule
        \multirow{2}{*}{Original} & -- & -- & -- & 76.1 & 83.6 & 18 & 0.143 & 0.222 & 0.019 & 0.052 & \textbf{0.81} & 1.11 & \textbf{0.24} & \textbf{0.96} & 0.043 & 0.060 & 0.022 & 0.026 & 0.022 & 0.026 \\
         & \checkmark & 15.1 & 7.6 & \textbf{81.6} & \textbf{89.0} & \textbf{13} & \textbf{0.130} & \textbf{0.202} & \textbf{0.016} & \textbf{0.040} & 0.82 & \textbf{0.96} & 0.25 & 1.02 & \textbf{0.039} & \textbf{0.056} & 0.022 & 0.026 & 0.022 & 0.026 \\
        \midrule
        \multirow{2}{*}{Occluded} & -- & -- & -- & 61.2 & 67.3 & 56 & \textbf{0.093} & 0.228 & 0.018 & 0.057 & 0.84 & 1.10 & 0.41 & 1.20 & 0.049 & 0.066 & 0.031 & 0.037 & 0.031 & 0.037 \\
         & \checkmark & 28.0 & 13.8 & \textbf{71.9} & \textbf{79.3} & \textbf{25} & 0.113 & \textbf{0.201} & 0.018 & \textbf{0.046} & \textbf{0.83} & \textbf{1.04} & \textbf{0.34} & \textbf{0.97} & 0.049 & \textbf{0.065} & 0.031 & 0.037 & 0.031 & 0.037 \\
        \bottomrule
    \end{tabular}
}
\label{tab:sampling_confidence_filter_h36m}
\end{table}%

\paragraph{Confidence weights} Table~\ref{tab:confidence_weights_off} compares the pipeline with and without the weights; without them, all observations of the retained keypoints contribute equally, including the low-confidence ones. The weights consistently improve the reconstructed poses: they lower the PA-MPJPE in every sequence, most under occlusion, where its median drops from 0.036 to 0.031\,m (from 0.023 to 0.022\,m on the original images), with the same trend for the other detectors and on EgoHumans. Their impact on calibration is small: the medians of the scale-aligned translation error and of the angular error change by at most 1\,mm and 0.01\textdegree{}.
\begin{table}[!t]
\centering
\caption{Effect of the confidence weights on Human3.6M~\cite{human36m} with RTMPose, on the original images and with synthetic occlusions~\cite{bragagnoloMultiviewPoseFusion2024}, with the filters kept in both runs. Settings and sequences match Table~\ref{tab:sampling_confidence_filter_h36m}, and each cell gives the median over sequences.}
\newcolumntype{H}{>{\setbox0\hbox\bgroup}c<{\egroup}@{}}
\maxsizebox{\linewidth}{!}{
    \begin{tabular}{ll|HHcHcHcH|HHHHcH}
        \toprule
        \multirow{2}{*}{\textbf{Input}} & \multirow{2}{*}{\textbf{Weights}} & \multicolumn{8}{c|}{\textbf{Camera Metrics}} & \multicolumn{6}{c}{\textbf{Human Metrics}} \\
        & & TE (m)$\downarrow$ & p90 & s-TE (m)$\downarrow$ & p90 & AE (\textdegree)$\downarrow$ & p90 & FoV (\textdegree)$\downarrow$ & p90 & W-MPJPE (m)$\downarrow$ & p90 & GA-MPJPE (m)$\downarrow$ & p90 & PA-MPJPE (m)$\downarrow$ & p90 \\
        \midrule
        \multirow{2}{*}{Original} & -- & \textbf{0.126} & 0.205 & 0.016 & 0.048 & 0.82 & 0.97 & \textbf{0.21} & \textbf{0.99} & 0.040 & 0.056 & 0.023 & 0.028 & 0.023 & 0.028 \\
         & \checkmark & 0.130 & \textbf{0.202} & 0.016 & \textbf{0.040} & 0.82 & \textbf{0.96} & 0.25 & 1.02 & \textbf{0.039} & 0.056 & \textbf{0.022} & \textbf{0.026} & \textbf{0.022} & \textbf{0.026} \\
        \midrule
        \multirow{2}{*}{Occluded} & -- & \textbf{0.098} & \textbf{0.200} & 0.019 & \textbf{0.037} & 0.83 & \textbf{0.98} & \textbf{0.31} & 1.11 & 0.052 & 0.068 & 0.036 & 0.044 & 0.036 & 0.044 \\
         & \checkmark & 0.113 & 0.201 & \textbf{0.018} & 0.046 & 0.83 & 1.04 & 0.34 & \textbf{0.97} & \textbf{0.049} & \textbf{0.065} & \textbf{0.031} & \textbf{0.037} & \textbf{0.031} & \textbf{0.037} \\
        \bottomrule
    \end{tabular}
}
\label{tab:confidence_weights_off}
\end{table}%

The filters thus give RANSAC a cleaner input and the weights make the reconstructed poses more accurate. Calibration accuracy depends little on either use because calibration is strongly overdetermined and robust: RANSAC rejects the remaining outlier correspondences, and bundle adjustment fits $M$ keypoints over all views under a Huber loss. Triangulation, in contrast, solves each keypoint of each frame from only a few views, so that a single unreliable view displaces the point unless its weight is reduced. Both uses discard or down-weight unreliable keypoints rather than select the most confident ones, which would lower the coverage and degrade the calibration (Sec.~\ref{sec:ablation_study_sampling}).

\subsubsection{Impact of Global Scaling Strategy}
\label{sec:ablation_study_global_scale}

Table~\ref{tab:ablation_study_global_scaling} presents the ablation study evaluating the impact of the global scale estimation strategy used on reconstruction accuracy and camera metrics.
\begin{table}[!b]
\centering
\caption{Ablation study on the impact of different global scale estimation strategies (SMPL-based and metric depth-based using MoGe-2~\cite{wangMoGeUnlockingAccurate2025}) on reconstruction accuracy (W-MPJPE, in meters) and camera translation error (TE, in meters). All camera parameters and 2D keypoints are fixed to ground-truth values, isolating the effect of scale estimation. Results are reported for EgoHumans and Human3.6M datasets. For the oracle shape, the SMPL shape parameters are fixed to those fitted to the ground-truth bone lengths, giving an upper bound on SMPL-based scaling.}
\maxsizebox{\linewidth}{!}{
\begin{tabular}{ll|c|c}
    \toprule
    & \textbf{Scaling Method} & \textbf{W-MPJPE} $\downarrow$ & \textbf{TE} $\downarrow$ \\
    \midrule
    \multirow{3}{*}{EgoHumans}
& SMPL-based & 0.138 & 0.252 \\
& SMPL-based, oracle shape & 0.019 & 0.027 \\
& Metric Depth-based & 0.595 & 1.074 \\
\hline
\multirow{3}{*}{Human3.6M}
& SMPL-based & 0.021 & 0.103 \\
& SMPL-based, oracle shape & 0.002 & 0.008 \\
& Metric Depth-based & 0.145 & 0.721 \\
    \bottomrule
\end{tabular}
}
\label{tab:ablation_study_global_scaling}
\end{table}%
\begin{figure}[tbp]
  \centering
  \includegraphics[width=\linewidth, max width=8.5cm]{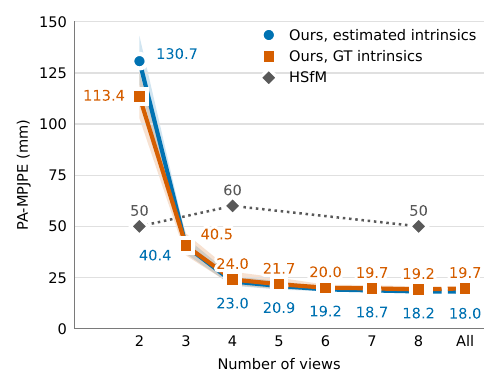}
  \caption{Mean PA-MPJPE (in millimeters) against the number of views on the EgoHumans~\cite{egohumans} sequences, with estimated and ground-truth intrinsics. Bands give the 95\% confidence interval of the mean over sequences. Camera sets are nested from two to eight views; \emph{All} uses every static camera of the sequence (8 to 19).}
  \label{fig:n_views}
\end{figure}%
In this analysis, all camera parameters and 2D keypoints are fixed to their ground-truth values; only the global scale factor is computed and applied. This design isolates the effect of scale estimation on the reconstructed scene, independent of errors from camera calibration or 2D keypoint detection. SMPL-based scaling is substantially more accurate on both datasets, with a W-MPJPE of 0.138 against 0.595\,m on EgoHumans and 0.021 against 0.145\,m on Human3.6M. Its remaining error mostly comes from the mean shape prior: fixing $\beta$ to an oracle shape fitted to the ground-truth bone lengths (Sec.~\ref{sec:smpl_scale}) reduces the W-MPJPE to 0.019\,m on EgoHumans and 0.002\,m on Human3.6M. The error of metric depth-based scaling comes from the depth estimator itself: on Human3.6M, the estimated scale is too small by 12 to 16\% in all sequences, a magnitude comparable to the mean relative depth error that MoGe-2 reports on its metric depth benchmarks (13.6\% with ground-truth intrinsics and 15.7\% without, averaged over seven datasets)~\cite{wangMoGeUnlockingAccurate2025}. Metric depth-based scaling therefore remains a fallback for scenes where no human prior is available. A detailed analysis is provided in Sec.~\ref{supp-sec:depth_scale_analysis} of the supplementary materials.

\subsubsection{Impact of the Number of Views}
\label{sec:ablation_study_n_views}

We evaluate the effect of the number of cameras on EgoHumans, whose sequences have 8 to 19 cameras. For each sequence, we run the full pipeline on the subsets of cameras of~\citet{mullerReconstructingPeoplePlaces2025}, and on nested sets in between. Results are given in Fig.~\ref{fig:n_views} and Table~\ref{supp-tab:n_views} of the supplementary materials. Over all the sequences, the mean PA-MPJPE drops from 131\,mm with two views to 40\,mm with three and 23\,mm with four, then decreases slowly to 18\,mm with eight views. Using ground-truth instead of estimated intrinsics changes the error by less than 2\,mm from three views on. With two views, Kineo is less accurate than HSfM (50\,mm): a single camera pair gives the calibration no redundancy, and the estimated direction between the two cameras is off by more than 15\textdegree{} in 25 of the two-view sequences. Kineo, however, benefits from every additional view and is more accurate than every reported HSfM result from three views on, whereas the PA-MPJPE of HSfM remains between 0.05 and 0.06\,m from two to eight views.

\subsection{Qualitative Evaluation}
\begin{figure}
  \centering
  \captionsetup[sub]{justification=raggedright,singlelinecheck=false}
  \begin{subfigure}[b]{0.49\linewidth}
    \centering
    \includegraphics[width=\textwidth]{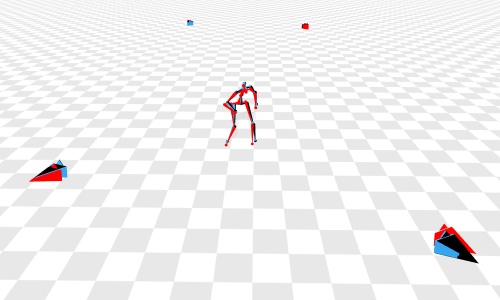}
    \caption{Frame with the minimum W-MPJPE (0.076 m) from HSfM's best sequence, \textit{S11\_Eating 1}. The overall sequence has a mean W-MPJPE of 0.165$\pm$0.276 m.}
    \label{fig:h36m_qualitative_comparison_hsfm_best}
  \end{subfigure}
  \hfill
  \begin{subfigure}[b]{0.49\linewidth}
    \centering
    \includegraphics[width=\textwidth]{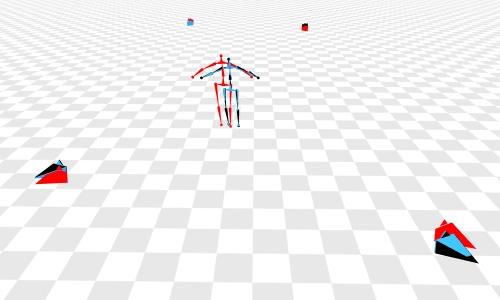}
    \caption{Frame with the minimum W-MPJPE (0.011 m) from Kineo's best sequence, \textit{S11\_WalkTogether 1}. The overall sequence has a mean W-MPJPE of 0.016$\pm$0.003 m.}
    \label{fig:h36m_qualitative_comparison_kineo_best}
  \end{subfigure}
  \caption{Qualitative comparison between HSfM (\textbf{\textcolor{red}{red}}) and Kineo (\textbf{\textcolor{cyan}{blue}}) on their respective best-performing sequences from Human3.6M. Ground-truth cameras and skeletons are shown in \textbf{\textcolor{black}{black}}. Visualization is scale-aligned.}
  \label{fig:h36m_qualitative_comparison}
\end{figure}
Figure~\ref{fig:h36m_qualitative_comparison} presents qualitative comparisons between the best sequences obtained with HSfM and Kineo. For each method, we show the sequence with the minimum mean W-MPJPE. In both cases, Kineo shows better camera pose estimation, with lower mean W-MPJPE and standard deviation. The full comparison on all Human3.6M sequences is available in Table~\ref{supp-tab:h36m_quantitative_full} of the supplementary materials.

\begin{figure*}[htbp]
    \centering
    \begin{subfigure}[c]{0.24\textwidth}
        \centering
        \includegraphics[width=\textwidth,trim={15cm 0 15cm 7cm},clip]{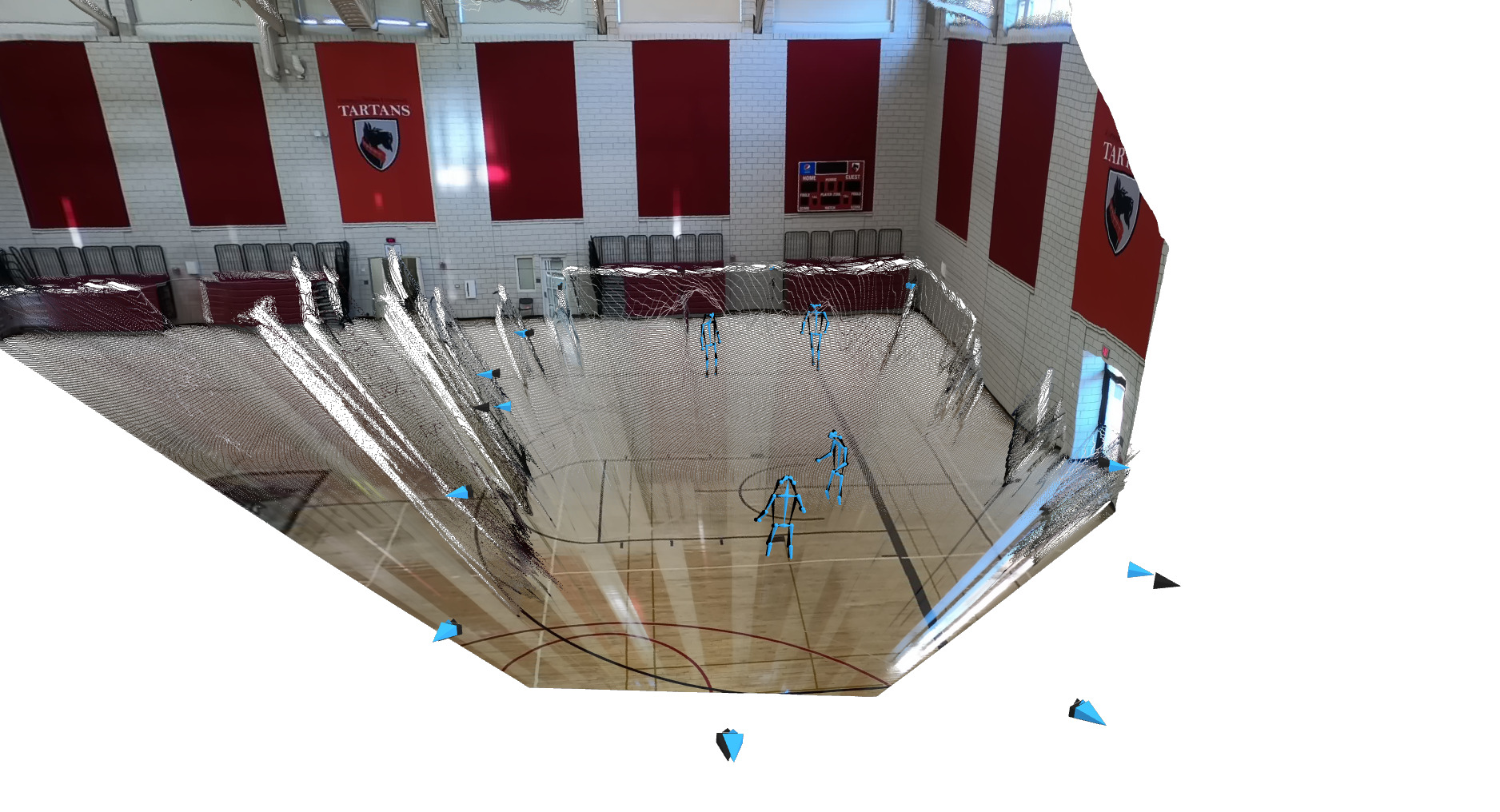}
        \caption{Badminton}
        \label{fig:a}
    \end{subfigure}
    \begin{subfigure}[c]{0.24\textwidth}
        \centering
        \includegraphics[width=\textwidth,trim={15cm 0 15cm 7cm},clip]{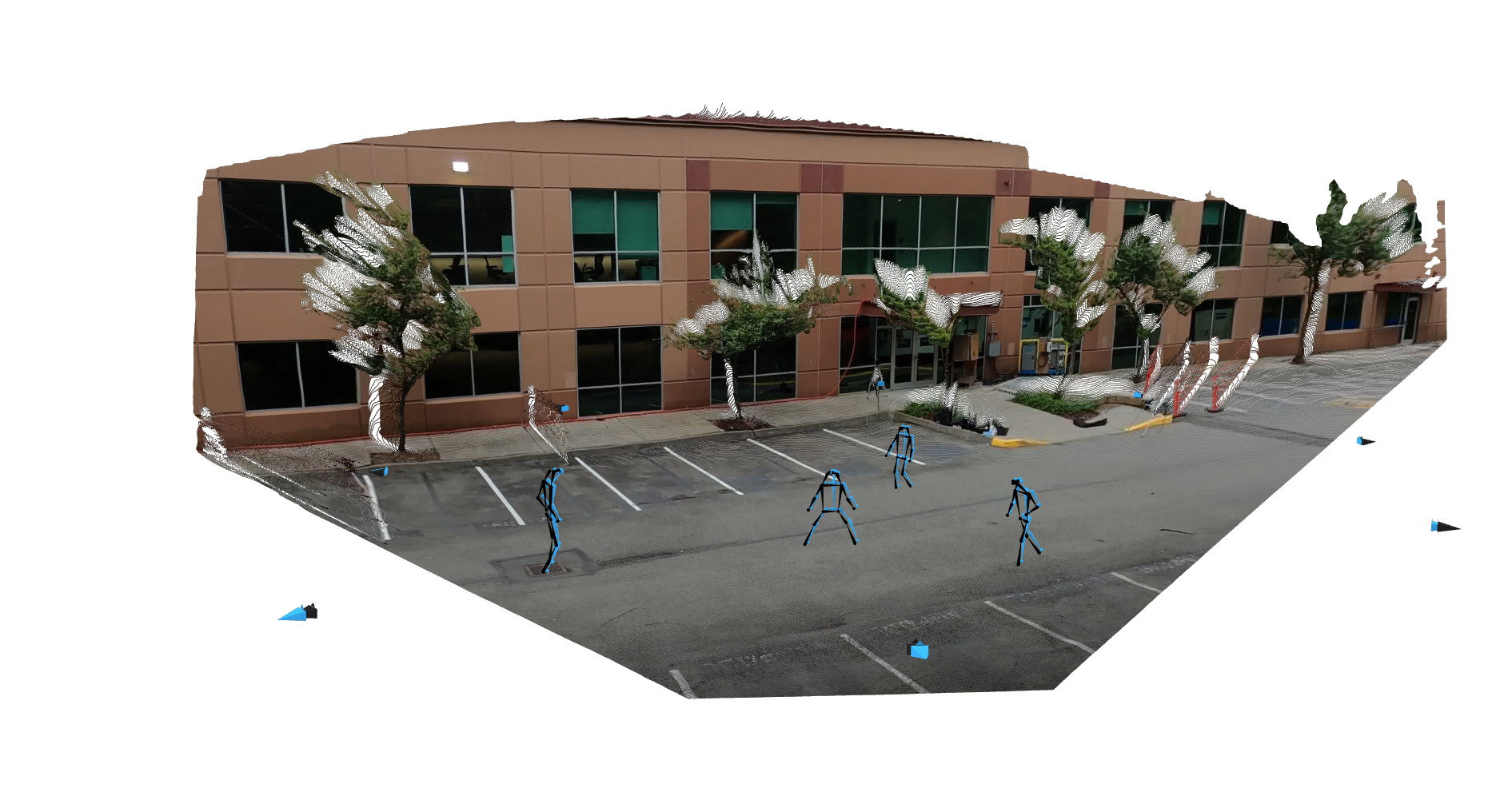}
        \caption{Basketball}
        \label{fig:b}
    \end{subfigure}
    \begin{subfigure}[c]{0.24\textwidth}
        \centering
        \includegraphics[width=\textwidth,trim={10cm 0 15cm 5cm},clip]{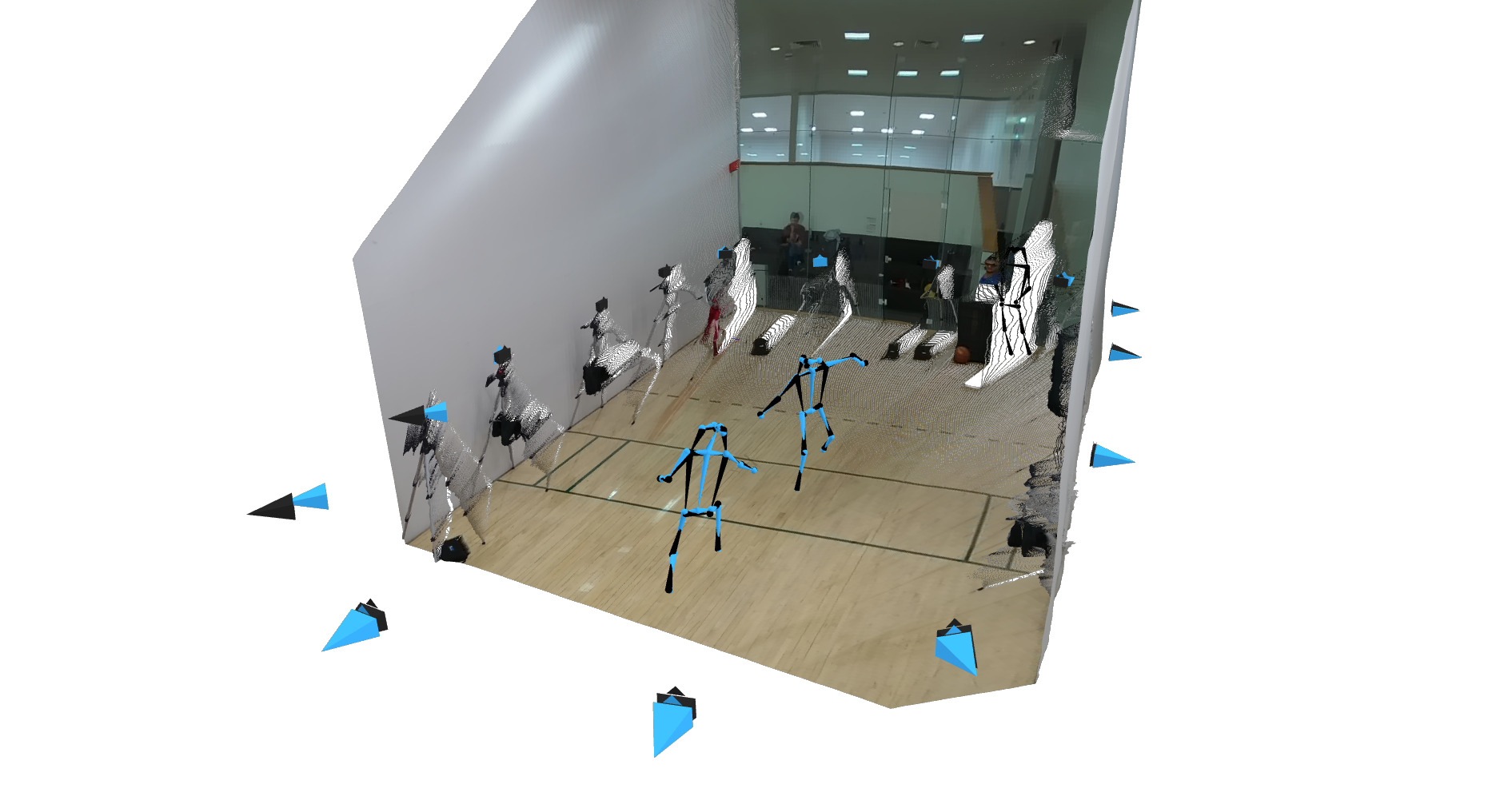}
        \caption{Fencing}
        \label{fig:c}
    \end{subfigure}
    \begin{subfigure}[c]{0.24\textwidth}
        \centering
        \includegraphics[width=\textwidth,trim={14cm 3cm 6cm 5cm},clip]{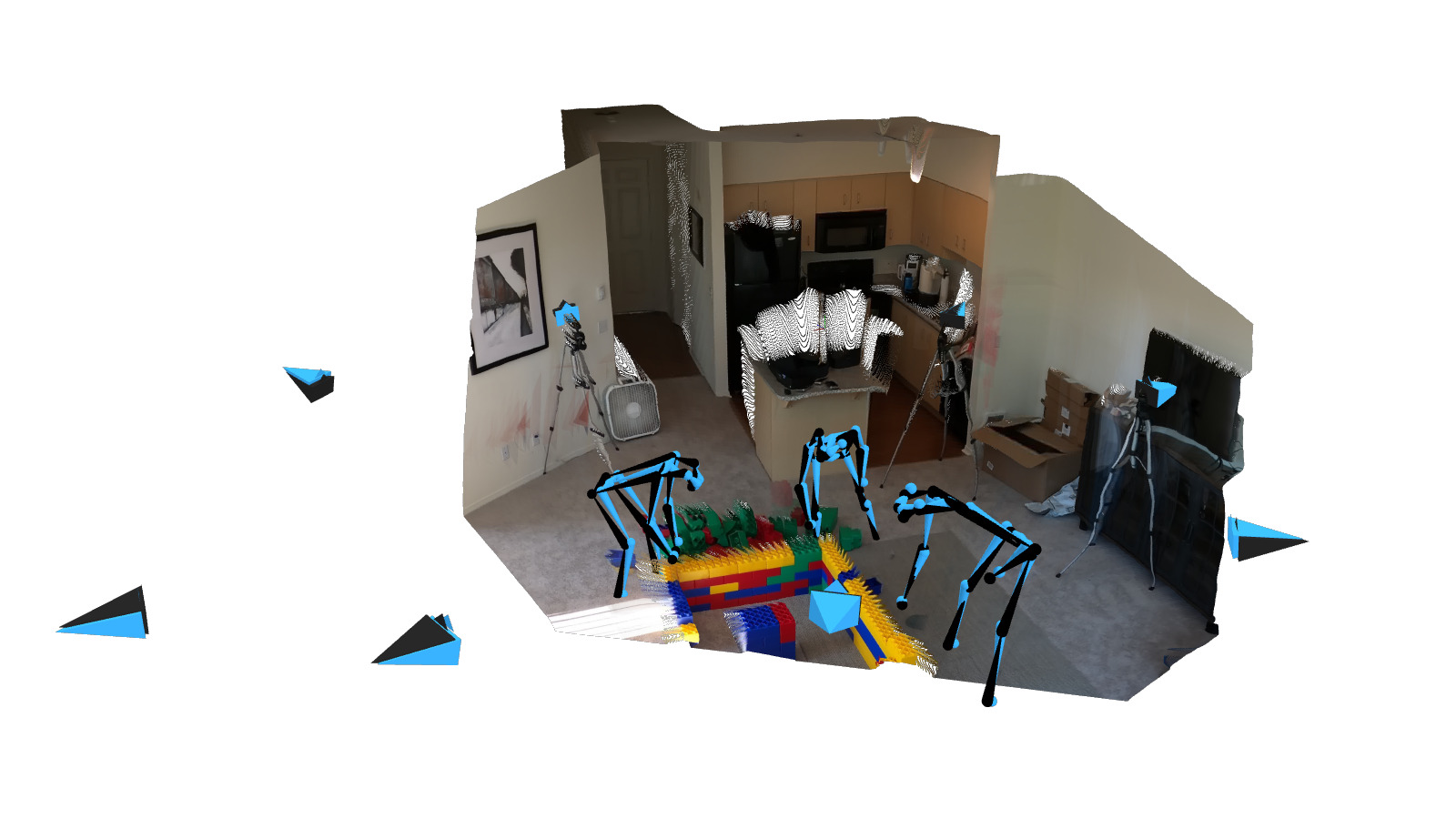}
        \caption{Lego Assembly}
        \label{fig:d}
    \end{subfigure}
    
    \vspace{1em}
    
    \begin{subfigure}[c]{0.32\textwidth}
        \centering
        \includegraphics[width=\textwidth,trim={2cm 3cm 6cm 8cm},clip]{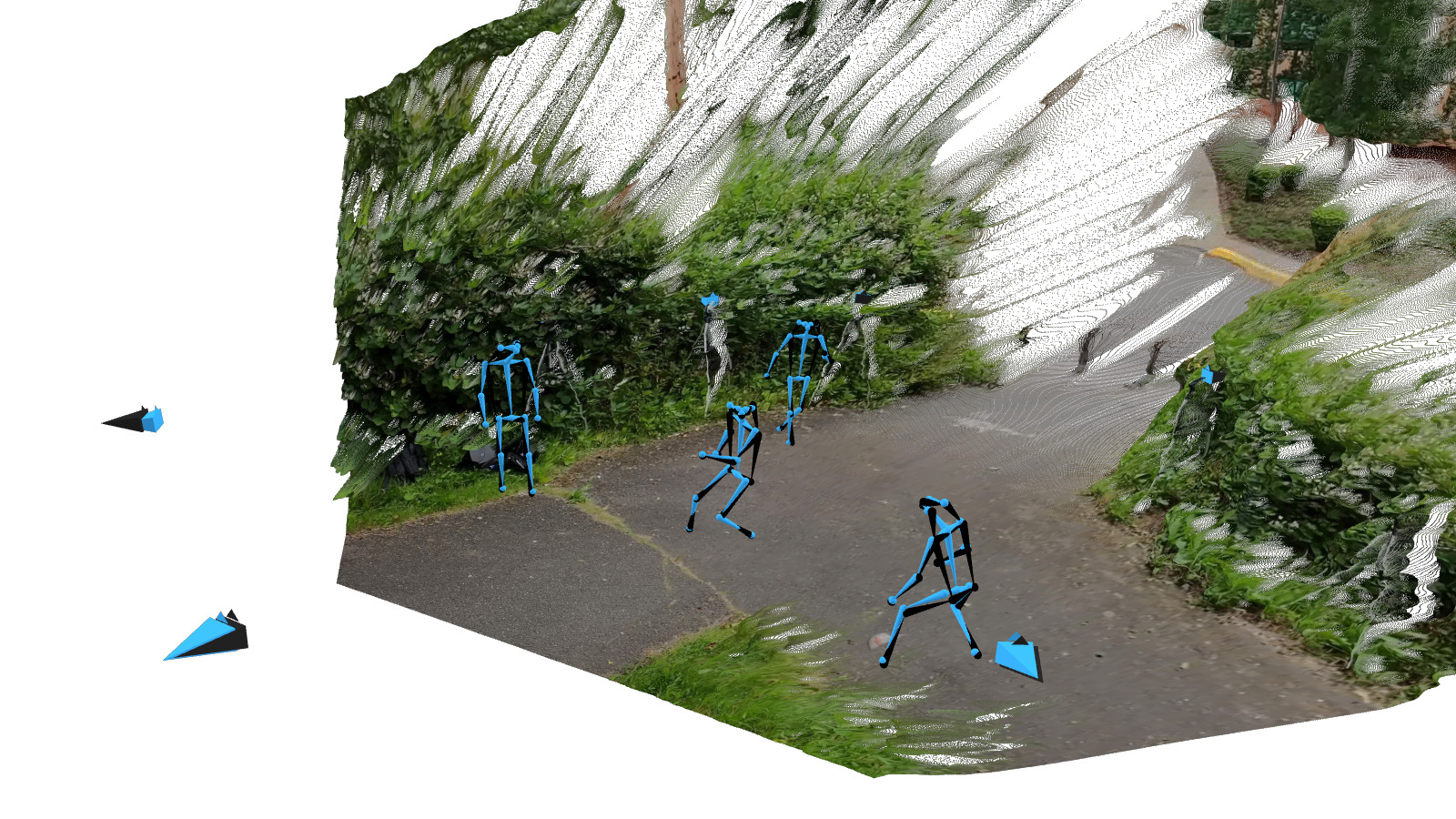}
        \caption{Tagging}
        \label{fig:e}
    \end{subfigure}
    \begin{subfigure}[c]{0.32\textwidth}
        \centering
        \includegraphics[width=\textwidth,trim={15cm 0cm 5cm 10cm},clip]{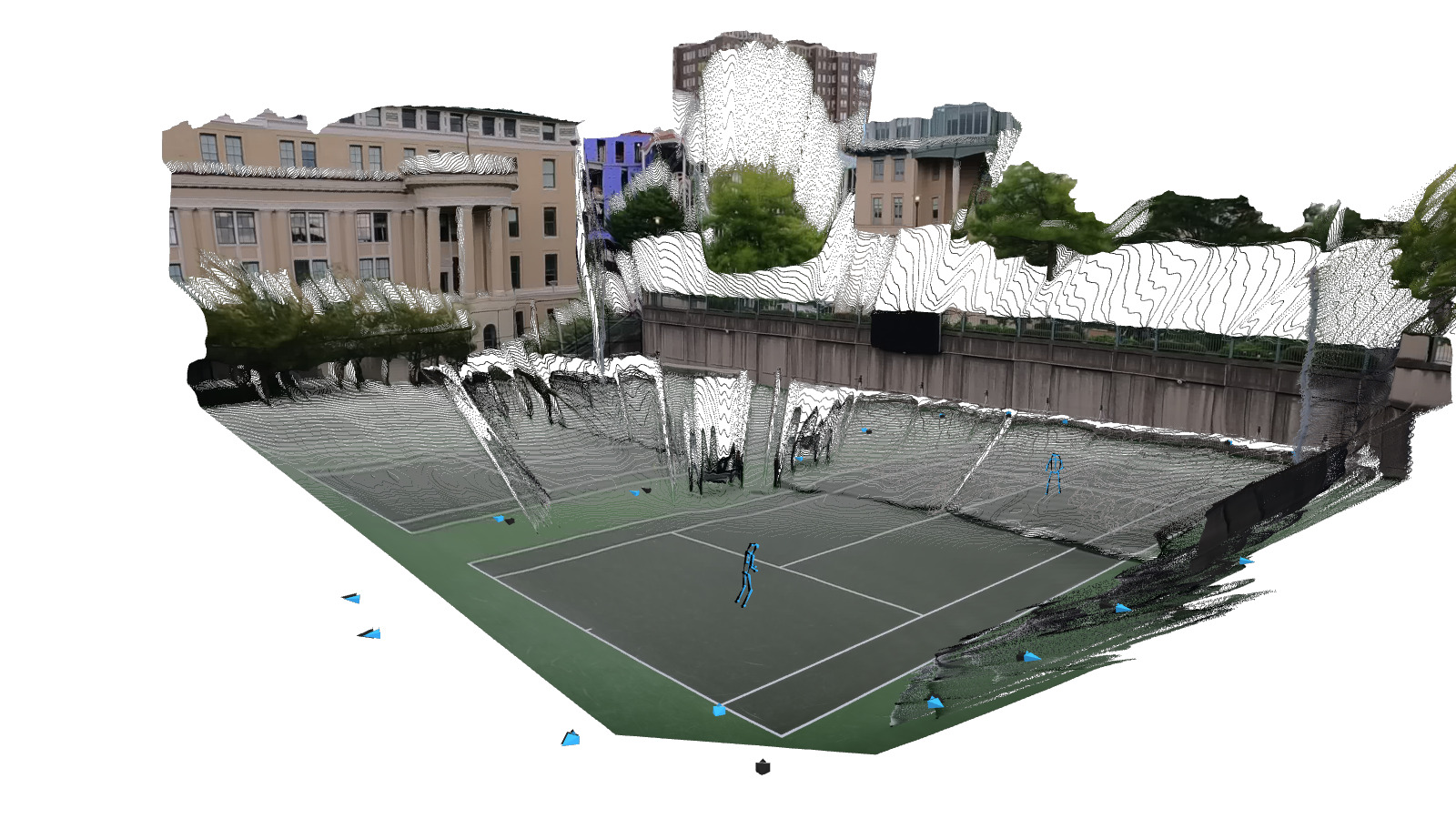}
        \caption{Tennis}
        \label{fig:f}
    \end{subfigure}
    \begin{subfigure}[c]{0.32\textwidth}
        \centering
        \includegraphics[width=\textwidth,trim={5cm 3cm 10cm 5cm},clip]{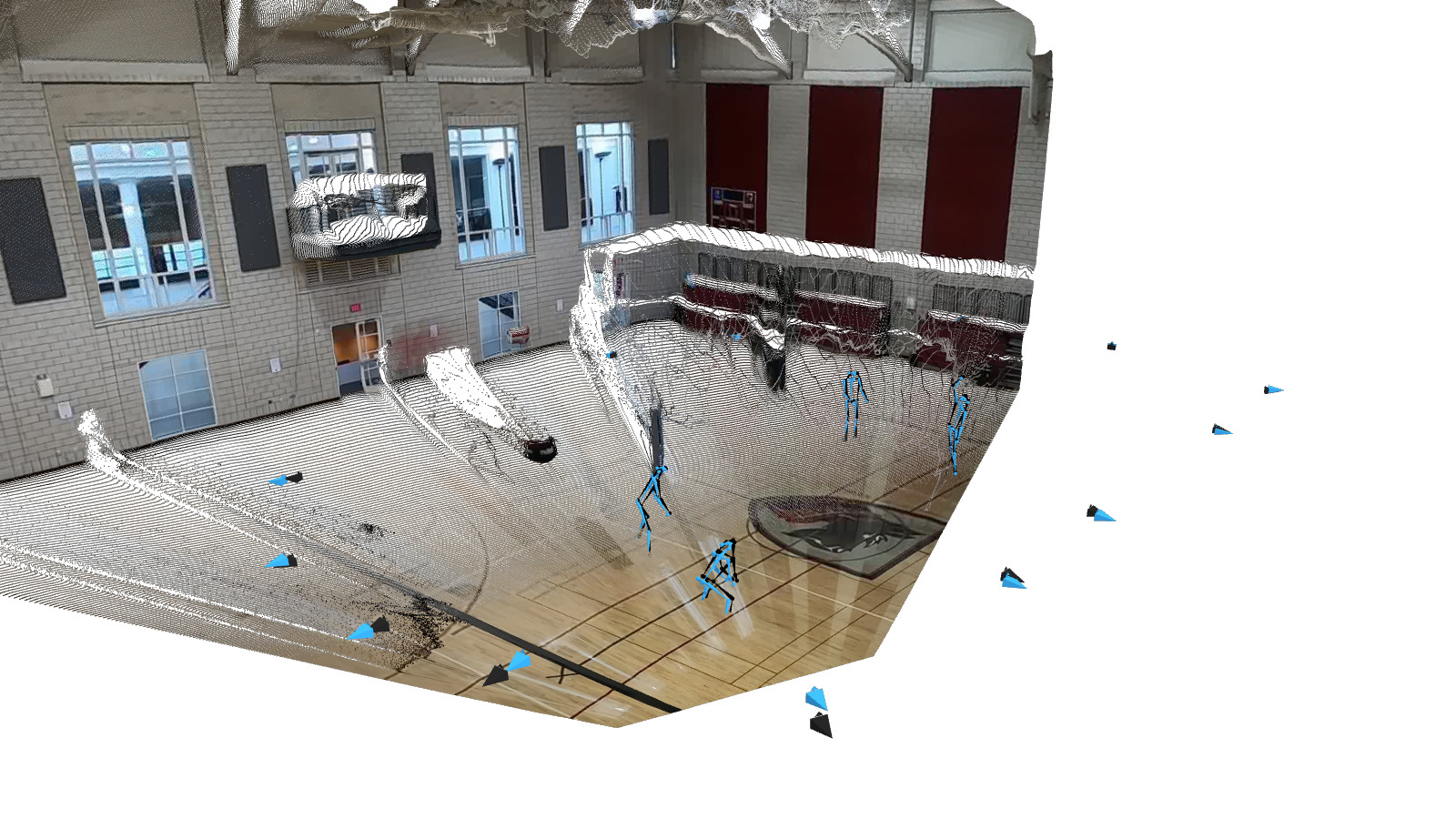}
        \caption{Volleyball}
        \label{fig:g}
    \end{subfigure}
    
    \caption{Qualitative results on sequences from the \textit{EgoHumans} dataset. Kineo estimations are visualized in \textbf{\textcolor{cyan}{blue}}, while ground-truth cameras and skeletons are shown in \textbf{\textcolor{black}{black}}. The scene pointmap is predicted as a downstream task of our method using MoGe-2~\cite{wangMoGeUnlockingAccurate2025}.}
    \label{fig:egohumans_qualitative}
\end{figure*}

Figure~\ref{fig:egohumans_qualitative} illustrates qualitative results obtained using Kineo on representative sequences from the EgoHumans dataset, including activities such as badminton, basketball, fencing, lego assembly, tagging, tennis, and volleyball. For each sequence, the scene pointmaps were reconstructed from a single camera view. The images were first undistorted using the estimated distortion coefficients, and the estimated focal length was subsequently provided to MoGe-2~\cite{wangMoGeUnlockingAccurate2025} to enhance the accuracy of the reconstruction. Qualitatively, Kineo produces estimated results (in blue) that closely align with the ground truth (in black).

A qualitative example of lens undistortion on EgoHumans is shown in Figure~\ref{fig:distortion_error_maps}. The original GoPro frames exhibit pronounced barrel distortion, particularly visible along the image borders. By estimating and correcting for this distortion, we recover images close to a pinhole projection in the regions where the subjects move; towards the image borders, where the distortion is extrapolated, residual distortion remains (Sec.~\ref{sec:ablation_study_distortion}). Additional qualitative results are provided in Fig.~\ref{supp-fig:egohumans_distortions_full} of the supplementary materials. Videos and interactive visualizations of reconstructed sequences are described in Sec.~\ref{supp-sec:additional_files} of the supplementary materials.

\section{Practical Applications}
Kineo provides two processing modes: offline and online. The offline mode is the primary mode, designed for unconstrained, real-world video capture, delivering high reconstruction accuracy and supporting long sequences without on-site calibration. The online mode enables real-time processing of live video streams, offering immediate feedback for interactive applications.

\subsection{In-the-Wild Capture (Offline Mode)}
\label{sec:use_case_offline}
Kineo enables rapid deployment using battery-powered cameras and eliminates the need for on-site calibration and thereby facilitates outdoor data capture by non-experts within minutes. Kineo was used in a real-world use case to capture the gestures of craftsmen in an experimental archaeology worksite, where a castle is currently being built using medieval techniques. The goal was to preserve this intangible cultural heritage. For this case, the emphasis was placed on reconstruction accuracy, achieved through the use of high-capacity keypoint detection models and spatio-temporal subsampling across the entire video sequence. 

\begin{figure}[htbp]
    \centering
    \begin{subfigure}[b]{0.49\linewidth}
        \centering
        \includegraphics[width=\textwidth]{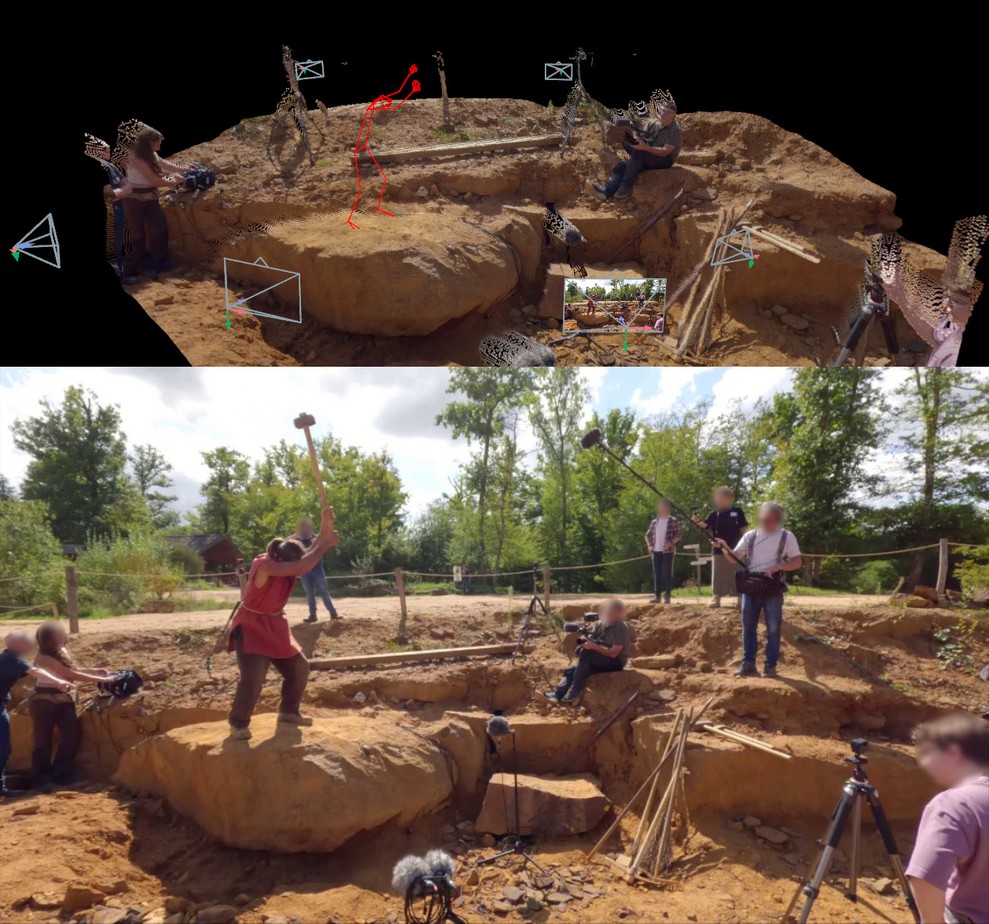}
        \label{fig:stonemason}
    \end{subfigure}
    \hfill
    \begin{subfigure}[b]{0.49\linewidth}
        \centering
        \includegraphics[width=\textwidth]{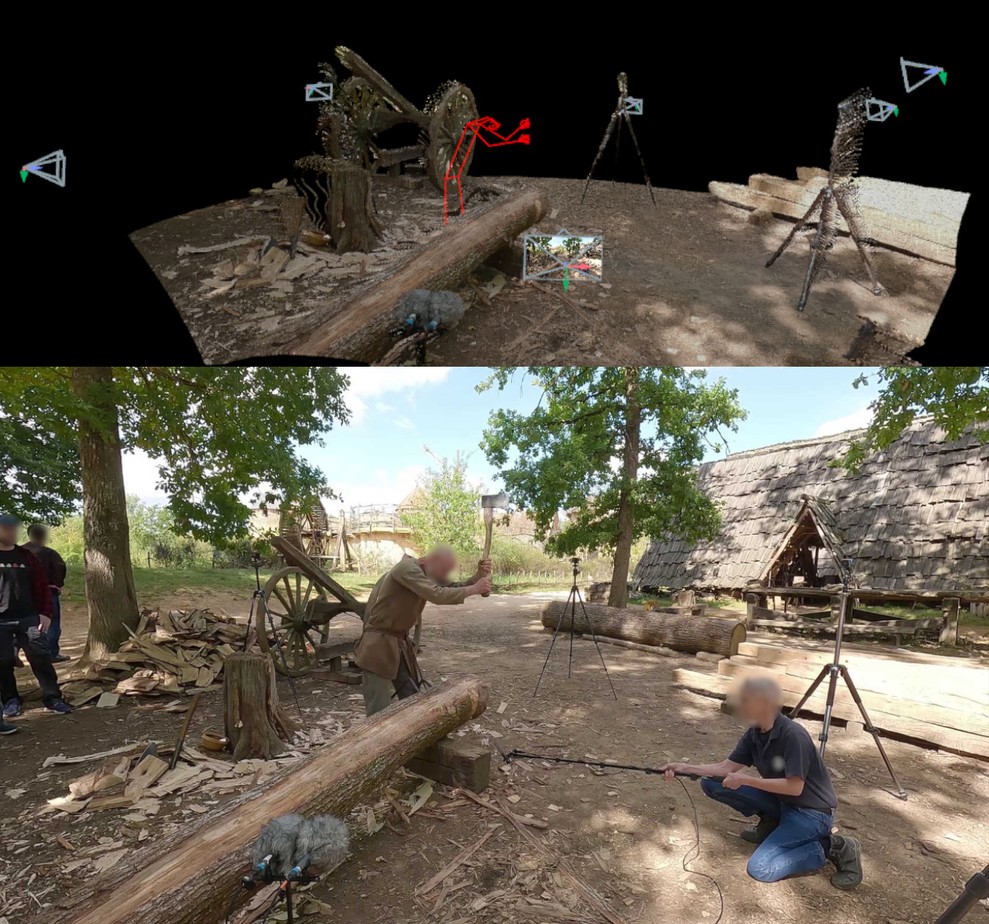}
        \label{fig:wood_hewer}
    \end{subfigure}

    \caption{Examples of craftsmen recorded at an experimental archaeology worksite. Both sequences were captured using six uncalibrated and unsynchronized GoPro Hero11 Black cameras, with no calibration procedure necessary.}
    \label{fig:use_case_guedelon}
\end{figure}

A notable advantage of Kineo compared to prior works is its capacity to process significantly longer video sequences than those typically found in public datasets, with durations ranging from five to thirty minutes. Two illustrative examples of craftsmen motion capture are presented in Figure~\ref{fig:use_case_guedelon}. These sequences were captured using six uncalibrated and unsynchronized GoPro Hero11 Black cameras, with the entire setup installation completed in approximately ten minutes. All subsequent processing was performed on a local workstation equipped with an NVIDIA RTX 3080 Ti. For instance, a representative five-minute sequence (yielding a cumulative 30 minutes of footage) required approximately 11 minutes for keypoint detection using DWPose, followed by 30 seconds for camera calibration with Kineo. The most computationally demanding stage involved segmentation and cross-view re-identification of individuals using a custom implementation of SAM2, which took approximately 2.5 hours to process all views. We used this step here to obtain the best possible association; it is not part of the runtime experiments (Fig.~\ref{fig:runtime_metrics}) or of the online mode (Sec.~\ref{sec:use_case_online}). This step was optimized to prevent out-of-memory errors, as the original SAM2 lacked effective memory-management strategies. Future work may explore more efficient tracking methods to further reduce processing time for this stage.

\subsection{Real-Time Capture (Online Mode)}
\label{sec:use_case_online}
\begin{table}[b]
\centering
\caption{Offline and online modes of Kineo on Human3.6M~\cite{human36m}. The online mode calibrates on the first 10\,s only, then reconstructs each four-view frame as it arrives, at the given latency (median\,/\,95th percentile) and per-view frame rate on an RTX 3090. Person boxes are detected with YOLOX-tiny~\cite{geYOLOXExceedingYOLO2021}, and 2D keypoints with RTMPose~\cite{rtmpose} trained on the Human3.6M training set, so that they follow its skeleton. TE, W-MPJPE and PA-MPJPE are in meters, AE in degrees.}
\maxsizebox{\linewidth}{!}{
    \begin{tabular}{l|cc|cc|cc}
        \toprule
        \textbf{Mode} & TE$\downarrow$ & AE$\downarrow$ & W-MPJPE$\downarrow$ & PA-MPJPE$\downarrow$ & Latency (ms)$\downarrow$ & fps$\uparrow$ \\
        \midrule
        Offline & \textbf{0.138} & \textbf{0.87} & \textbf{0.041} & \textbf{0.024} & \NA & \NA \\
        Online & 0.186 & 1.11 & 0.049 & \textbf{0.024} & 19.9\,/\,23.9 & 52.4 \\
        \bottomrule
    \end{tabular}
}
\label{tab:online_vs_offline}
\end{table}

The real-time (online) mode represents an extension of Kineo's original offline pipeline, specifically designed to process live video streams and provide immediate visual feedback during capture. This mode is particularly advantageous for applications such as interactive motion capture systems and real-time avatar animation.

In this mode, Kineo processes video input directly from multiple webcams. The system first executes the automatic calibration phase, during which a short sequence of about 10\,s is recorded to estimate camera parameters. Following calibration, Kineo continuously processes incoming frames to detect and reconstruct the subject in real time. Person detection is performed using the YOLOX-tiny~\cite{geYOLOXExceedingYOLO2021} model, while 2D keypoint estimation is carried out with DWPose. These models were selected for their ability to maintain a balance between accuracy and computational efficiency. An example video of live capture is available in the supplementary materials (Fig.~\ref{supp-fig:video_realtime}).

Table~\ref{tab:online_vs_offline} compares this mode to the offline mode on Human3.6M, with the same pipeline and the same frames, the only difference being the calibration window. Both modes detect keypoints with RTMPose, whose skeleton matches the Human3.6M annotations. Calibrating on the first 10\,s instead of the whole sequence costs 0.05\,m of camera translation error and 8\,mm of W-MPJPE, while PA-MPJPE is unchanged. In exchange, reconstruction runs at 52\,fps per view on four views (about 210\,fps in total), above the 50\,fps of the dataset's cameras.

\section{Limitations}
\label{sec:limitations}

Kineo replaces synchronization hardware and calibration targets with assumptions about the capture and priors on the subjects, so that its accuracy depends on how well these hold. We discuss below the scope of these assumptions, in the order of the pipeline.

Audio synchronization assumes a sound common to all cameras and a constant offset between them (Sec.~\ref{sec:temporal_sync}), and reaches frame-level accuracy most reliably when the audio contains sharp transients, such as a clap during the capture (Sec.~\ref{sec:ablation_study_audio_sync}).

Calibration relies on redundancy between views: with only two cameras, the relative pose of the single camera pair cannot be cross-checked against other pairs, and Kineo is less accurate than HSfM (Sec.~\ref{sec:ablation_study_n_views}). Moreover, since it relies on the keypoints of the subjects, the lens distortion is best constrained where they are observed, which is also where the skeletons are triangulated, whereas towards the image borders it is constrained only by the few keypoints observed there and by the regularizer, and can overfit them (Sec.~\ref{sec:ablation_study_distortion}). This mainly matters for tasks that use the whole undistorted image, such as predicting the scene pointmaps.

Each frame is triangulated independently, so that the 3D keypoints show the jitter of the 2D detections (Sec.~\ref{sec:triangulation}), and the scene pointmaps are tied to the skeletons only through the cameras and the global scale, so that feet can penetrate the floor (Sec.~\ref{sec:scene_pointmaps}). Temporal filtering and contact constraints could address both.

SMPL-based scaling is biased towards the mean SMPL shape (Sec.~\ref{sec:ablation_study_global_scale}), a bias that a known subject height can reduce when available (Sec.~\ref{sec:smpl_scale}). When no human is present, metric depth-based scaling follows the accuracy of the monocular depth estimator, and therefore directly benefits from its progress.

Finally, the consistency score reflects the agreement between views rather than the 3D error (Sec.~\ref{sec:pairwise_reprojection_confidence_score}). Propagating the 2D detection uncertainty through triangulation~\cite{hartleyMultipleViewGeometry2003} could complement it with a 3D uncertainty, for instance to weight temporal filtering.

\section{Conclusion}
We presented Kineo, a fully automatic, calibration-free pipeline for metric multi-view markerless motion capture that operates on unsynchronized, uncalibrated, consumer RGB cameras. By leveraging a confidence-aware, keypoint-first approach, Kineo discards unreliable correspondences and subsamples the remaining ones across frames and views, enabling robust calibration and reconstruction without the computational cost of per-frame parametric fitting. Its design couples classical SfM geometry with modern keypoint detection, resulting in a modular, scalable, and future-proof system: improvements in 2D detectors directly translate into higher 3D accuracy without architectural changes.

Extensive evaluations conducted on the EgoHumans and Human3.6M datasets demonstrate that Kineo significantly reduces the performance gap between calibration-free and calibrated methods, surpassing previous state-of-the-art calibration-free approaches. Moreover, its computational efficiency enables both offline processing of extended video sequences and real-time operation. Notably, Kineo scales effectively to real-world, in-the-wild capture scenarios, achieving processing speeds that, in some configurations, allow it to process videos faster than their actual duration.

Kineo has already helped us record the gestures of craftsmen at a medieval castle construction site, to preserve this intangible cultural heritage. We hope it will help others capture human motion in the many places where dedicated systems cannot go.

\begin{acks}
This work was supported by the Auvergne-Rhône-Alpes region as part of the PROMESS project. This work was granted access to the HPC resources of IDRIS under the allocation 2025-AD010614830 made by GENCI. We also express our gratitude to the Guédelon Castle for kindly welcoming us and permitting the captures that were essential to this study.
\end{acks}

\bibliographystyle{ACM-Reference-Format}
\bibliography{kineo}

  \clearpage
  \onecolumn
  \setcounter{section}{0}
  \setcounter{table}{0}
  \setcounter{figure}{0}
  \renewcommand{\thesection}{S\arabic{section}}
  \renewcommand{\thetable}{S\arabic{table}}
  \renewcommand{\thefigure}{S\arabic{figure}}
  \renewcommand{\theHsection}{S\arabic{section}}
  \renewcommand{\theHtable}{S\arabic{table}}
  \renewcommand{\theHfigure}{S\arabic{figure}}
  \pdfbookmark[0]{Supplementary Materials}{supplementary}
  {\centering\LARGE\sffamily\bfseries Supplementary Materials\par}
  \bigskip
  The supplementary material provides additional results and resources that complement the main paper. It includes extended quantitative and qualitative evaluations on the Human3.6M and EgoHumans datasets, highlighting both 3D human reconstruction and camera calibration performance. In addition, we provide video and interactive files, enabling exploration and qualitative assessment of our method. Full instructions for accessing and visualizing these resources are included in Section~\ref{supp-sec:additional_files}. The sections follow the order of the main paper. Sec.~\ref{supp-sec:notation} lists the notation of the main paper and the implementation details and hyperparameters needed to reproduce our results. Sec.~\ref{supp-sec:sync} details the evaluation of temporal synchronization on AIST++, and Sec.~\ref{supp-sec:egohumans} the evaluation of the estimated lens distortion on EgoHumans. Sec.~\ref{supp-sec:ablations} extends the ablation study of the main paper, and Sec.~\ref{supp-sec:h36m} gives per-sequence results on Human3.6M.

\section{Notation and Implementation Details}
\label{supp-sec:notation}

\subsection{Notation}
\label{supp-sec:notation_table}

Table~\ref{supp-tab:notation} lists the symbols used in the method of the main paper (Sections~\ref{sec:temporal_sync} to~\ref{sec:global_scale}), grouped by pipeline stage and by the section that introduces them. The values used in our experiments are given in Tables~\ref{supp-tab:implementation_details} and~\ref{supp-tab:sampling_settings}.

{\small
\begin{longtable}{@{}p{0.20\linewidth}p{0.76\linewidth}@{}}
\caption{Notation used in the main paper.}
\label{supp-tab:notation}\\
\toprule
Symbol & Description \\
\midrule
\endfirsthead
\toprule
Symbol & Description \\
\midrule
\endhead
\bottomrule
\endfoot
\multicolumn{2}{@{}l}{\textit{Temporal synchronization (Sec.~\ref{sec:temporal_sync})}} \\*
$\gamma_k$ & Temporal lag of camera $k$, relative to a reference camera \\
$\bar{\gamma}_k$ & Ground-truth temporal lag of camera $k$ (AIST++) \\
$H$ & Hop length used during the Short-Time Fourier Transform \\
$f_s$, $f_r$ & Audio sample rate and video frame rate \\
$d_1$, $d_2$ & Distance from each camera to the sound source \\
$c$ & Speed of sound in air \\
$\Delta t_1$, $\Delta t_2$ & Synchronization errors due to the hop length and to the difference in camera-to-source distances \\
\midrule
\multicolumn{2}{@{}l}{\textit{Cameras (Sec.~\ref{sec:automatic_camera_calibration})}} \\*
$C_i$ & Camera $i$, a node of the camera graph \\
$N$ & Number of views \\
$K_i$, $f_i$ & Intrinsic matrix and focal length of camera $C_i$ \\
$\mathbf{d}_i$ & Brown-Conrady distortion coefficients of camera $C_i$ \\
$[R\mid\mathbf{t}]_i$ & Extrinsics of camera $C_i$ \\
$\pi(\cdot)$ & Projection function \\
$E_{ij}$ & Essential matrix between cameras $C_i$ and $C_j$ \\
$R_{ij}$, $\hat{\mathbf{t}}_{ij}$ & Relative rotation and translation direction from camera $C_i$ to $C_j$ \\
\midrule
\multicolumn{2}{@{}l}{\textit{Pairwise calibration (Secs.~\ref{sec:graph_relative_extrinsics} and~\ref{sec:graph_edge_reliability})}} \\*
$\mathbf{x}_{i,k}$ & 2D keypoint $k$ in view $i$ \\
$w_{i,k}$ & 2D keypoint detector confidence score, in $[0,1]$ \\
$w_{ij,k}$ & Score of a point correspondence between cameras $C_i$ and $C_j$, the geometric mean of the detector confidences \\
$\tau$ & Threshold on $w_{ij,k}$ for retaining a correspondence \\
$P$ & Number of sampled correspondences \\
$\bar{\mathbf{x}}_{i,k}$ & Image coordinates of keypoint $k$ in view $i$ divided by the image width and height, in $[0,1]^2$ \\
$\bar{t}_k$ & Time of the frame of keypoint $k$ divided by the sequence duration, in $[0,1]$ \\
$\hat{\mathbf{x}}_{i,p}$ & Sampled correspondence $p$ in view $i$, normalized by the intrinsic matrix \\
$\mathcal{S}(\cdot)$ & Normalized Sampson distance \\
$e_{ij}$ & Calibration score of the camera pair, the average Sampson distance over the $P$ sampled correspondences, used as the weight of the edge connecting $C_i$ and $C_j$ \\
\midrule
\multicolumn{2}{@{}l}{\textit{Relative scales and absolute extrinsics (Secs.~\ref{sec:graph_relative_scales} and~\ref{sec:graph_absolute_extrinsics})}} \\*
$\lambda_{ij}$ & Relative scale factor of the camera pair $(C_i, C_j)$ \\
$\boldsymbol{\lambda}$ & Unknown relative scale factors, one per camera pair, $\boldsymbol{\lambda}\in\mathbb{R}^{Q}$ \\
$Q$, $L$ & Number of camera pairs and number of loop constraints \\
$A$ & Coefficients from the loop constraints, $A\in\mathbb{R}^{3L\times Q}$ \\
$c_{ijk}$, $\omega_{ijk}$ & Cost and weight of the loop through cameras $C_i$, $C_j$, $C_k$ \\
$W$ & Diagonal matrix of loop weights, $W\in\mathbb{R}^{3L\times 3L}$ \\
$\mu$ & Regularizer weight \\
$\mathcal{P}^*_k$ & Path from $C_1$ to $C_k$ in the minimum spanning tree \\
\midrule
\multicolumn{2}{@{}l}{\textit{Bundle adjustment (Sec.~\ref{sec:graph_bundle_adjustment})}} \\*
$\kappa$ & Minimum keypoint confidence threshold \\
$\mathbf{X}_m$, $M$ & Triangulated 3D point $m$ and number of 3D points \\
$\mathbf{x}_{i,m}$, $w_{i,m}$ & Observed 2D detection of point $m$ in view $i$ and its detector confidence \\
$\epsilon_{i,m}$ & Reprojection residual, expressed in pixels \\
$\mathcal{H}_\delta$ & Huber loss with threshold $\delta$, on the reprojection residuals \\
$\mathcal{L}_\text{reproj}$ & Weighted average reprojection error \\
$\mu_\mathbf{d}$ & Weight of the L2 regularizer on the distortion coefficients (third pass) \\
\midrule
\multicolumn{2}{@{}l}{\textit{Pairwise reprojection consistency score (Sec.~\ref{sec:pairwise_reprojection_confidence_score})}} \\*
$\tilde{\epsilon}_{i,m}$ & Reprojection residual normalized by the focal length \\
$\eta$ & Rate of the exponential decay \\
$s_{i,m}$ & Reprojection score of point $m$ in view $i$, in $[0,1]$ \\
$s_{ij,m}$ & Agreement between views $i$ and $j$ for point $m$ \\
$s_m$ & Consistency score of the 3D point $m$, in $[0,1]$ \\
\midrule
\multicolumn{2}{@{}l}{\textit{SMPL-based scale estimation (Sec.~\ref{sec:smpl_scale})}} \\*
$\alpha$; $\alpha_\text{init}$, $\alpha^*$ & Global scale factor; its initial and optimized values \\
$\beta$; $\beta_\text{init}$ & SMPL shape parameters; their initial value, the mean shape \\
$M(\beta)$ & Unposed SMPL model, giving the body vertices \\
$\mathcal{J}$, $K$ & Joint regressor and number of keypoints \\
$\hat{\mathbf{j}}$ & Skeleton keypoints of the SMPL model \\
$\mathbf{j}^t$, $T$ & 3D skeleton keypoints in arbitrary units at frame $t$, and number of frames \\
$\mathcal{B}$ & Set of bones defined by skeleton connectivity \\
$\ell^t_{kl}$, $\ell_{kl}$ & Length of bone $(k,l)$ in arbitrary units at frame $t$, and aggregated over the frames \\
$s^t_{kl}$, $s_{kl}$ & Confidence of bone $(k,l)$ at frame $t$, and aggregated over the frames \\
$\operatorname{med}_\text{w}$ & Weighted median \\
$\hat{\ell}_{kl}$ & Length of bone $(k,l)$ in the SMPL skeleton \\
$(k^*,l^*)$ & Most confident bone \\
$\mathcal{L}_\text{bones}$, $\mathcal{L}_\beta$ & Bone length consistency and body shape losses \\
$\mathcal{H}_\delta$ & Huber loss with threshold $\delta$, on the bone length differences \\
$\mathcal{L}_\text{tot}$; $\lambda_\text{bones}$, $\lambda_\beta$ & Total loss; weights of its terms \\
\midrule
\multicolumn{2}{@{}l}{\textit{Metric depth-based scale estimation (Sec.~\ref{sec:depth_scale})}} \\*
$D$ & Predicted metric depth map \\
$(x_c, y_c, z_c)$, $(u, v)$ & Keypoint in camera space and its projected image coordinates \\
$d_a$, $d_m$ & Keypoint depth in arbitrary and metric units \\
$h_i$ & Reliability score of a frame for view $i$ \\
$\zeta$ & Keypoint confidence threshold for depth sampling \\
\end{longtable}
}

\subsection{Implementation Details}
\label{supp-sec:implementation_details}

Table~\ref{supp-tab:implementation_details} lists, stage by stage, the implementation choices and hyperparameters shared by all experiments unless stated otherwise, including those that the main paper does not state. The values that depend on the detector or on the experiment are given in Sec.~\ref{supp-sec:experimental_settings}. Unless stated otherwise, the optimizers are the L-BFGS implementation of PyTorch with a learning rate of 1 and a strong Wolfe line search, run for a number of steps of at most 20 iterations each, and stopped early when the gradient falls below $10^{-5}$ or the loss changes by less than $10^{-9}$ for five consecutive steps.

{\small
\begin{longtable}{@{}p{0.30\linewidth}p{0.66\linewidth}@{}}
\caption{Implementation details and hyperparameters of each stage of the pipeline, shared by all experiments unless stated otherwise.}
\label{supp-tab:implementation_details}\\
\toprule
Parameter & Value \\
\midrule
\endfirsthead
\toprule
Parameter & Value \\
\midrule
\endhead
\bottomrule
\endfoot
\multicolumn{2}{@{}l}{\textit{Temporal synchronization (Sec.~\ref{sec:temporal_sync})}} \\*
Reference camera & First camera; the other audio tracks are resampled to its sample rate \\
Audio features & 13 MFCCs over 128 HTK mel bands (log-mel in dB), 2048-point FFT, 40\,ms Hann window, 5\,ms hop ($H=240$ at 48\,kHz) \\
Feature normalization & Each coefficient is standardized to zero mean and unit variance \\
Lag estimation & Maximum of the full cross-correlation, computed by FFT, without interpolation \\
\midrule
\multicolumn{2}{@{}l}{\textit{2D keypoints}} \\*
Detectors & NLF-L (multi-dataset, v0.3.2); RTMPose-l ($384\times288$) trained on Human3.6M; DWPose-l ($384\times288$, COCO-WholeBody and UBody); batches of 32 frames at full precision, except in the runtime experiment of Fig.~\ref{fig:runtime_metrics} (settings given there) and in the online experiment of Table~\ref{tab:online_vs_offline}, where RTMPose runs at half precision, through TensorRT for the streamed frames \\
Person boxes & Ground-truth 2D bounding boxes on all datasets, except in the runtime experiment of Fig.~\ref{fig:runtime_metrics} and the online experiment of Table~\ref{tab:online_vs_offline}, both on single-person Human3.6M sequences, where they are detected by RTMDet-tiny and YOLOX-tiny, respectively, with the most confident box kept per view; their detection time is included in the reported runtimes and latencies \\
Output skeleton & Human3.6M (17 joints) on Human3.6M; on EgoHumans and AIST++, COCO-19 with NLF, and the 17 COCO body joints with DWPose and ground-truth keypoints \\
Temporal resampling & Keypoints are resampled onto the synchronized timeline at 10\,fps (Human3.6M), 20\,fps (EgoHumans, AIST++) or 50\,fps (runtime and online experiments); a keypoint is not interpolated when either neighboring detection is more than 0.3\,s away \\
\midrule
\multicolumn{2}{@{}l}{\textit{Intrinsics initialization (Sec.~\ref{sec:intrinsics_init})}} \\*
Model & MoGe-2 ViT-L (normal variant), on the first frame of each view \\
Initial intrinsics & Focal length and principal point predicted by MoGe-2; zero distortion \\
\midrule
\multicolumn{2}{@{}l}{\textit{Pairwise calibration (Sec.~\ref{sec:graph_relative_extrinsics})}} \\*
Correspondence selection & Uniform random selection of at most $P$ correspondences with $w_{ij,k}>\tau$ (Table~\ref{supp-tab:sampling_settings}) \\
Essential matrix & Five-point algorithm in RANSAC (OpenCV), on keypoints undistorted with the current camera estimate; inlier threshold 3\,px, at most 1000 iterations, confidence in Sec.~\ref{supp-sec:experimental_settings} \\
Edge weight $e_{ij}$ & Sampson distance averaged over all $P$ sampled correspondences, on intrinsics-normalized coordinates; edges with fewer than five correspondences receive an infinite weight \\
\midrule
\multicolumn{2}{@{}l}{\textit{Relative scales (Sec.~\ref{sec:graph_relative_scales})}} \\*
Optimization & L-BFGS on $\log\boldsymbol{\lambda}$, initialized at $\boldsymbol{\lambda}=\mathbf{1}$, 10 steps of at most 20 iterations; $\mu=10^{-3}$ \\
\midrule
\multicolumn{2}{@{}l}{\textit{Absolute extrinsics (Sec.~\ref{sec:graph_absolute_extrinsics})}} \\*
Spanning tree & Kruskal's algorithm (NetworkX), rooted at the first camera \\
\midrule
\multicolumn{2}{@{}l}{\textit{Bundle adjustment (Sec.~\ref{sec:graph_bundle_adjustment})}} \\*
Keypoint selection & Uniform random selection of at most $M$ keypoints detected with $w_{i,k}>\kappa$ in at least two views (Table~\ref{supp-tab:sampling_settings}), initialized by confidence-weighted DLT \\
Parameterization & Rotations in the 6D representation; first camera fixed; principal point fixed; single focal length $f_x=f_y$; intrinsics shared across views, initialized by averaging the per-view estimates normalized by the image resolution \\
Distortion model & Brown-Conrady, five coefficients ($k_1$, $k_2$, $p_1$, $p_2$, $k_3$), regularization weight $\mu_\mathbf{d}=1$ \\
Loss & Confidence-weighted Huber loss on the reprojection error, $\delta=1$\,px \\
Optimization & L-BFGS on CPU, at most 300 steps per pass \\
\midrule
\multicolumn{2}{@{}l}{\textit{Triangulation and consistency score (Secs.~\ref{sec:triangulation} and~\ref{sec:pairwise_reprojection_confidence_score})}} \\*
Triangulation & DLT with the rows of each view pair weighted by $\sqrt{w_{i,m}\,w_{j,m}}$, solved by SVD, without confidence threshold \\
Consistency score & Decay rate $\eta=1$ \\
\midrule
\multicolumn{2}{@{}l}{\textit{SMPL-based scale estimation (Sec.~\ref{sec:smpl_scale})}} \\*
Body model & Neutral SMPL, 10 shape parameters \\
Bones & 16 bones of the Human3.6M skeleton, 18 of the COCO-19 skeleton or 19 of the COCO-17 skeleton, from the corresponding SMPL joint regressor \\
Optimization & L-BFGS on $\log\alpha$ and $\beta$, at most 100 steps; $\delta=0.5$\,m, $\lambda_\text{bones}=\lambda_\beta=1$; one $\alpha$ shared by all subjects and one $\beta$ per subject, initialized at the mean shape ($\beta_\text{init}=\mathbf{0}$) \\
\midrule
\multicolumn{2}{@{}l}{\textit{Metric depth-based scale estimation (Sec.~\ref{sec:depth_scale})}} \\*
Model & MoGe-2 ViT-L (normal variant), same as for intrinsics initialization \\
Threshold and decay rate & $\zeta=0.3$, $\eta=1$ \\
\midrule
\multicolumn{2}{@{}l}{\textit{Other settings}} \\*
Random seeds & 19 for the pipeline, except 42 in the runtime experiment with NLF; 42 for the synthetic occlusions \\
Synthetic occlusions & PASCAL VOC 2012 objects pasted on three views, following the protocol cited in the main paper \\
\end{longtable}
}

\subsection{Experimental Settings}
\label{supp-sec:experimental_settings}

The confidence thresholds $\tau$ and $\kappa$ and the sampling budgets $P$ and $M$ are set empirically per detector, as the confidence scores of different detectors are not on the same scale. Table~\ref{supp-tab:sampling_settings} lists their values in each experiment of the main paper and of these supplementary materials. The RANSAC confidence of the pairwise calibration is 0.999 for the NLF runs on EgoHumans and AIST++, and 0.99 in the other experiments.

\begin{table}[htbp]
\centering
\caption{Confidence thresholds and sampling budgets used in the experiments of the main paper and of these supplementary materials: threshold $\tau$ and budget $P$ of the pairwise calibration (Sec.~\ref{sec:graph_relative_extrinsics}), threshold $\kappa$ and budget $M$ of bundle adjustment (Sec.~\ref{sec:graph_bundle_adjustment}). Experiments that take the cameras from the ground truth do not run calibration and are omitted. Ground-truth keypoints all have a confidence of 1, so that their thresholds only discard missing or non-interpolated keypoints.}
\label{supp-tab:sampling_settings}
\begin{tabular}{@{}llcccc@{}}
\toprule
Experiment & 2D keypoints & $\tau$ & $P$ & $\kappa$ & $M$ \\
\midrule
Tables~\ref{tab:human_metrics}, \ref{tab:camera_metrics}, \ref{tab:aistpp_sync}, \ref{tab:ablation_study_detectors} and~\ref{tab:sampling_egohumans}, Sec.~\ref{sec:ablation_study_distortion}, Figs.~\ref{fig:runtime_metrics} and~\ref{fig:n_views}; Tables~\ref{supp-tab:sampling_egohumans_categories}, \ref{supp-tab:confidence_weights_full} and~\ref{supp-tab:n_views}, Fig.~\ref{supp-fig:egohumans_distortions_full} & NLF & 0.6 & 3000 & 0.6 & 500 \\
Tables~\ref{tab:online_vs_offline}, \ref{tab:ablation_study_detectors}, \ref{tab:sampling_confidence_filter_h36m} and~\ref{tab:confidence_weights_off}, Fig.~\ref{fig:runtime_metrics}; Table~\ref{supp-tab:sampling_h36m} & RTMPose & 0.5 & 4000 & 0.5 & 400 \\
Table~\ref{tab:ablation_study_detectors}; Table~\ref{supp-tab:confidence_weights_full} & DWPose & 0.5 & 4000 & 0.5 & 400 \\
Tables~\ref{tab:ablation_study_camera_parameters} and~\ref{tab:ablation_study_detectors}, Human3.6M & Ground truth & 0.4 & 4000 & 0.4 & 200 \\
Tables~\ref{tab:ablation_study_camera_parameters} and~\ref{tab:ablation_study_detectors}, EgoHumans & Ground truth & 0.4 & 4000 & 0.5 & 200 \\
\bottomrule
\end{tabular}
\end{table}

\section{Temporal Synchronization}
\label{supp-sec:sync}
This section details the evaluation of temporal synchronization of Sec.~\ref{sec:quantitative_sync} of the main paper: how we recover the ground-truth time offsets of AIST++ (Sec.~\ref{supp-sec:aistpp_gt_offsets}), and the robustness of audio-based synchronization to degraded audio (Sec.~\ref{supp-sec:audio_sync_robustness}). AIST++ is the only one of our datasets that supports this evaluation: Human3.6M does not include audio, and only a single audio file per multi-view sequence of EgoHumans is currently available for download.

\subsection{Ground-Truth Offsets on AIST++}
\label{supp-sec:aistpp_gt_offsets}
We use the 451 sequences of the AIST++ pose estimation test split, excluding those that AIST++ flags as poorly annotated. AIST++ provides camera parameters and 3D keypoint annotations for the nine camera views of each sequence, whose videos come from the AIST Dance Video Database (\url{https://aistdancedb.ongaaccel.jp/}) and are encoded at 59.94\,fps. The database provides both the raw, unsynchronized videos, each carrying the audio recorded by the camera microphone, and refined versions trimmed to the dance performance, which AIST++ uses. As described on the database website, the refined versions were produced independently for each video: the start of the music is located in the raw audio by template matching on the count-in, a sequence of eight beat clicks preceding the music, with a target precision of 10\,ms (with a manual fallback when an automatic consistency check fails); the video is then trimmed at that instant and its audio replaced by the original music file. However, the ground-truth offset values are not released, nor are the click templates or the detection parameters, so this procedure cannot be re-run to recover the exact values. We instead recover them by locating each refined video in its raw recording: the frame $n_k$ at which view $k$ was trimmed is the integer shift that minimizes the mean absolute per-pixel difference between the frames of the two videos over the whole refined video, so that the ground-truth offsets are quantized to one frame. The ground-truth offset of view $k$ to the reference view is then $\bar{\gamma}_k = (n_k - n_1)/f_r$. In seven of the 451 sequences, the refined video of one camera is trimmed about one second later than the other eight, whereas the audio of the raw recordings places all cameras within a frame of each other, so this delay comes from the trimming; we exclude them and evaluate on the remaining 444 sequences. The excluded sequences, with the camera trimmed late and its delay relative to the other cameras, are:
\begin{itemize}
  \item \texttt{gBR\_sBM\_cAll\_d06\_mBR4\_ch06}, camera c03, 1.00\,s;
  \item \texttt{gBR\_sBM\_cAll\_d06\_mBR4\_ch08}, camera c03, 1.00\,s;
  \item \texttt{gBR\_sBM\_cAll\_d06\_mBR4\_ch10}, camera c03, 1.01\,s;
  \item \texttt{gJS\_sBM\_cAll\_d03\_mJS4\_ch04}, camera c04, 0.99\,s;
  \item \texttt{gJS\_sBM\_cAll\_d03\_mJS4\_ch05}, camera c04, 1.01\,s;
  \item \texttt{gJS\_sBM\_cAll\_d03\_mJS4\_ch07}, camera c04, 1.01\,s;
  \item \texttt{gJS\_sBM\_cAll\_d03\_mJS4\_ch08}, camera c04, 1.01\,s.
\end{itemize}

\subsection{Robustness of Audio-Based Synchronization}
\label{supp-sec:audio_sync_robustness}
Table~\ref{supp-tab:audio_sync_robustness_full} completes Table~\ref{tab:audio_sync_robustness} of the main paper with the mean and the 95th percentile (P95) of the absolute offset error, on the same sequences. The degraded conditions follow this protocol. To remove the count-in, each view is cut at the first frame of its refined video, which marks the start of the music, plus a random delay of up to 1\,s drawn once per camera, so that the cut edges carry no timing information; this leaves a median of 11\,s of the 22\,s recordings. The periodic beep, a 50\,ms, 1\,kHz tone with a raised-cosine envelope, whose amplitude before the envelope equals the RMS level of the recording, is added every 0.5\,s on the ground-truth timeline, so that the beeps are simultaneous in all views. The background noise comes from the park (NPARK) and cafeteria (PCAFETER) recordings of DEMAND: each view receives a different channel of the five-minute, 16-channel recording, with starting points spaced 33\,s apart, one ninth of its length, so that the nine views span the whole recording, and the signal-to-noise ratio compares the recording power to the noise power.

\begin{table}[ht]
\centering
\caption{Robustness of audio-based synchronization on AIST++ (444 sequences, 59.94\,fps), full results of Table~\ref{tab:audio_sync_robustness} of the main paper. For each of the eight non-reference views of a sequence, the error is the absolute difference between its estimated offset to the reference view and the ground-truth offset. \emph{Count-in removed} keeps only the music and is the baseline for the conditions below it.}
\maxsizebox{\linewidth}{!}{
\begin{tabular}{ll|ccc|ccc}
    \toprule
    & & \multicolumn{3}{c|}{\textbf{Offset error (ms)}} & \multicolumn{3}{c}{\textbf{Views within tolerance (\%)}} \\
    Audio condition & SNR & Median$\downarrow$ & Mean$\downarrow$ & P95$\downarrow$ & $\leq$17\,ms (1 fr.)$\uparrow$ & $\leq$33\,ms (2 fr.)$\uparrow$ & $\leq$67\,ms (4 fr.)$\uparrow$ \\
    \midrule
  Raw audio (with count-in) & -- & 6.9 & 8.9 & 21.9 & 89 & 99 & 100 \\
  Count-in removed & -- & 9.0 & 14.5 & 47.8 & 72 & 87 & 100 \\
  \quad + Periodic beep & -- & 7.8 & 9.4 & 28.6 & 87 & 96 & 100 \\
  \midrule
  \multirow{3}{*}{\quad + Park ambient} & 20\,dB & 9.0 & 16.8 & 52.8 & 68 & 80 & 100 \\
  & 10\,dB & 12.1 & 19.7 & 53.6 & 62 & 73 & 99 \\
  & 0\,dB & 14.0 & 34.3 & 57.8 & 52 & 63 & 99 \\
  \midrule
  \multirow{3}{*}{\quad + Cafeteria babble} & 20\,dB & 11.3 & 17.5 & 52.8 & 66 & 78 & 100 \\
  & 10\,dB & 12.9 & 21.4 & 57.2 & 58 & 68 & 99 \\
  & 0\,dB & 23.3 & 47.5 & 58.5 & 44 & 54 & 99 \\
  \bottomrule
\end{tabular}
}
\label{supp-tab:audio_sync_robustness_full}
\end{table}

\FloatBarrier
\section{Evaluation of the Estimated Distortion}
\label{supp-sec:egohumans}

\begin{figure}[p]
  \centering
  \setkeys{Gin}{width=0.265\linewidth}
  \newcommand{\drow}[1]{\rotatebox{90}{\makebox[0.1491\linewidth][c]{\scriptsize #1}}}
  \newcommand{\dsub}[1]{\parbox[t]{0.265\linewidth}{\centering\scriptsize #1}}
  \setlength{\tabcolsep}{1pt}
  \renewcommand{\arraystretch}{0.6}
  \begin{tabular}{@{}cccc@{}}
    \drow{Tagging} & \includegraphics{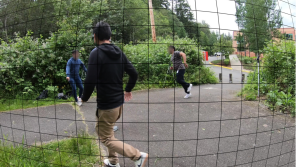} & \includegraphics{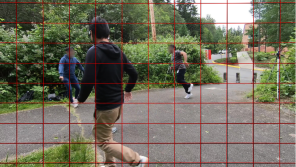} & \includegraphics{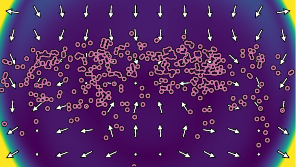} \\
    \drow{Lego Assembly} & \includegraphics{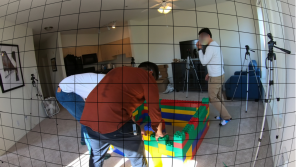} & \includegraphics{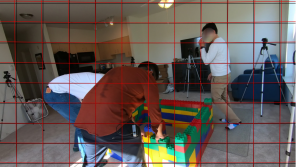} & \includegraphics{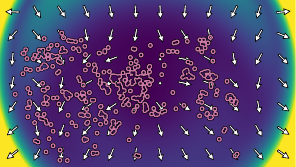} \\
    \drow{Fencing} & \includegraphics{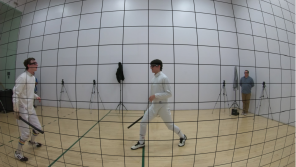} & \includegraphics{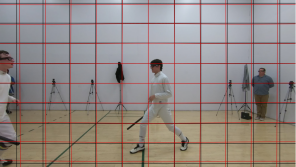} & \includegraphics{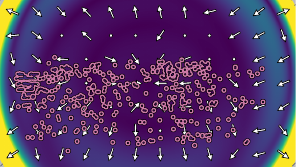} \\
    \drow{Basketball} & \includegraphics{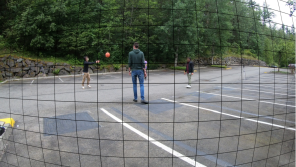} & \includegraphics{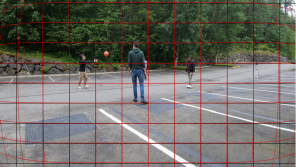} & \includegraphics{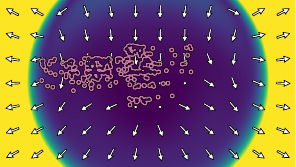} \\
    \drow{Volleyball} & \includegraphics{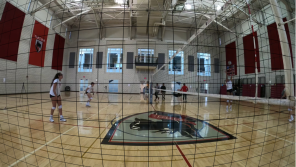} & \includegraphics{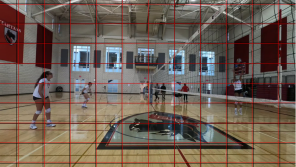} & \includegraphics{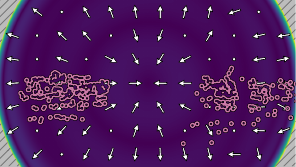} \\
    \drow{Badminton} & \includegraphics{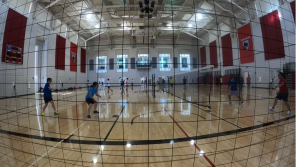} & \includegraphics{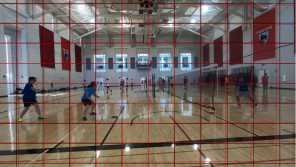} & \includegraphics{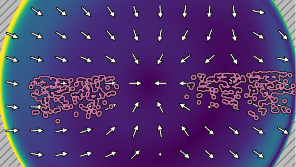} \\
    \drow{Tennis} & \includegraphics{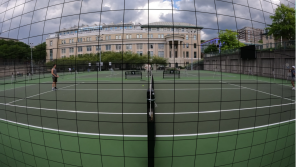} & \includegraphics{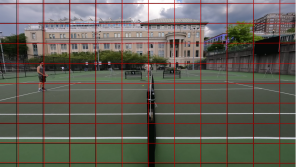} & \includegraphics{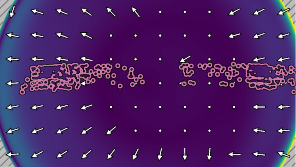} \\
    & \dsub{(a) Original distorted image with the ground-truth grid} & \dsub{(b) Undistorted image (estimated) with both grids} & \dsub{(c) Error map, focal length aligned to the ground truth} \\
  \end{tabular}

  \vspace{4pt}
  \includegraphics[width=0.85\linewidth]{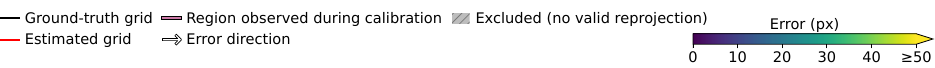}
  \caption{Estimated distortion on EgoHumans, one view per scene, with the same panels as Fig.~\ref{fig:distortion_error_maps} of the main paper. The error in (c) is shown on a linear color scale saturated at 50\,px, with its direction as unit arrows (dots below 1\,px); hatched samples are excluded (see Protocol). Each row shows a representative sequence of its scene from the camera whose keypoints cover the largest share of the image, among the cameras of the rig's majority focal length.}
  \label{supp-fig:egohumans_distortions_full}
\end{figure}

This section details the evaluation of Sec.~\ref{sec:ablation_study_distortion} of the main paper, on the distortion coefficients estimated by Kineo on the EgoHumans dataset, which features challenging real-world captures with wide-angle GoPro cameras. Figure~\ref{supp-fig:egohumans_distortions_full} extends Fig.~\ref{fig:distortion_error_maps} of the main paper to one view per scene.

\paragraph{Protocol} The ground-truth cameras of EgoHumans follow the OpenCV fisheye model and ours the Brown-Conrady model, and in both the focal length and the distortion are coupled, so we compare the mappings they induce rather than the coefficients themselves. For each view, a uniform grid of $540\times960$ pixels is unprojected with the ground-truth camera, the fisheye angle being found by Newton iterations, the resulting rays are reprojected with the estimated camera, and the error is the displacement between the two, in pixels of the full-resolution image. Pixels whose ray falls behind the estimated camera, or beyond the radius where its radial distortion polynomial folds back, are excluded (1.7\% of the samples for the median view). Each view is split into the region observed during calibration, within 25 pixels of a keypoint used in bundle adjustment, and the unobserved remainder of the image.

\paragraph{Focal length rescaling} Without rescaling, the error is 7.1\,px in the observed region against 14.0\,px in the unobserved region. This error mixes two effects: a wrong focal length alone already shifts every pixel away from or towards the image center, in proportion to its distance from it, and the focal length and the distortion are estimated together. To look at the distortion alone, we rescale the estimated focal length of each view, about its principal point, by the least-squares factor that best matches the ground truth in the observed region, which gives the error of 2.0\,px in the observed region reported in the main paper. The median focal length error, given by this factor, is 1.3\%. After rescaling, the median error in the unobserved region is 3.9\,px, but its 95th percentile reaches 19.5\,px, as the error grows towards the image borders. In two fencing and three tennis sequences, the median error over the image exceeds 50\,px; after rescaling the focal length, it falls below 50\,px in all of them, so that their error comes mainly from the focal length.

\paragraph{Shared intrinsics} As the intrinsics are shared across views (Sec.~\ref{sec:graph_bundle_adjustment} of the main paper), the keypoints of all views constrain a single estimate: a view observing subjects only near its center benefits from the keypoints that other views observe near their borders, whereas a view whose camera differs from the shared estimate is biased. On fencing, three of the cameras have a 9\% longer ground-truth focal length than the others, and the shared focal length is off by 3.8\% for the median view, which raises the error without rescaling in the observed region to 21.2\,px.

\paragraph{Per category} Table~\ref{supp-tab:distortion_error} gives the errors per category. After rescaling, a larger observed area does not give a lower error. Tennis has the smallest observed area (3.0\%) and the lowest error in the observed region. Lego Assembly has the largest (6.7\%) and the highest error in the unobserved region. The error instead grows with the distance from the image center. Over all views, unobserved pixels closer to the center than the farthest point of the observed region have a median error of 3.0\,px, and pixels farther out 12.5\,px. The error is higher farther out in every category. The 95th percentile of the unobserved error ranges from 7.7\,px (volleyball) to about 50\,px (fencing and Lego Assembly).

\begin{table}[htbp]
\centering
\caption{Distortion error (px) per EgoHumans category, in the observed and unobserved regions, without and with focal length rescaling. Each cell is the median over the views of a category of the per-view median error, or of its 95th percentile (p95). Obs.\ area is the share of the image covered by the observed region, and Focal the relative focal length error given by the rescaling factor.}
\maxsizebox{\linewidth}{!}{
    \begin{tabular}{l|cc|ccc|ccc}
        \toprule
        \multirow{2}{*}{\textbf{Category}} & \textbf{Obs.\ area} & \textbf{Focal} & \multicolumn{3}{c|}{\textbf{Without rescaling}} & \multicolumn{3}{c}{\textbf{With rescaling}} \\
        & (\%) & (\%)$\downarrow$ & Observed$\downarrow$ & Unobserved$\downarrow$ & Unobs.\ p95$\downarrow$ & Observed$\downarrow$ & Unobserved$\downarrow$ & Unobs.\ p95$\downarrow$ \\
        \midrule
        Badminton & 5.0 & 1.5 & 9.6 & 15.4 & 45.2 & 2.6 & 4.1 & 20.1 \\
        Basketball & 4.9 & 0.9 & 5.5 & 10.5 & 34.7 & 1.1 & 4.1 & 20.8 \\
        Fencing & 5.3 & 3.8 & 21.2 & 44.4 & 116.1 & 2.9 & 5.4 & 49.2 \\
        Lego Assembly & 6.7 & 0.6 & 4.5 & 9.9 & 69.0 & 2.2 & 6.9 & 52.5 \\
        Tagging & 6.3 & 1.1 & 7.2 & 11.5 & 33.0 & 1.9 & 3.3 & 18.9 \\
        Tennis & 3.0 & 0.4 & 1.9 & 3.4 & 13.3 & 0.9 & 2.4 & 11.6 \\
        Volleyball & 5.1 & 1.2 & 6.4 & 14.3 & 27.0 & 1.5 & 2.1 & 7.7 \\
        \midrule
        All & 4.9 & 1.3 & 7.1 & 14.0 & 41.9 & 2.0 & 3.9 & 19.5 \\
        \bottomrule
    \end{tabular}
}
\label{supp-tab:distortion_error}
\end{table}

\section{Ablation Study}
\label{supp-sec:ablations}
This section extends the ablation study of Sec.~\ref{sec:ablation_study} of the main paper. It details the comparison of sampling strategies per EgoHumans category and on Human3.6M (Sec.~\ref{supp-sec:sampling_categories}), evaluates the effect of the detection confidence with more detectors and on EgoHumans (Sec.~\ref{supp-sec:confidence_ablation}), analyzes the error of metric depth-based scale estimation (Sec.~\ref{supp-sec:depth_scale_analysis}), and details the evaluation of the number of views (Sec.~\ref{supp-sec:n_views}).

\subsection{Sampling Strategy}
\label{supp-sec:sampling_categories}

Sec.~\ref{sec:ablation_study_sampling} of the main paper compares our random subsampling with two alternatives that select as many correspondences among the same confidence-filtered candidates, namely $P$ per camera pair for pairwise calibration and $M$ keypoints for bundle adjustment. Farthest-point sampling (FPS) enforces spatio-temporal coverage explicitly: it draws a first candidate at random, then repeatedly adds the candidate farthest from those already selected, in Euclidean distance between feature vectors that encode where and when each candidate lies. For pairwise calibration, the feature vector of a correspondence is its position in both images and its time in the sequence:
\begin{equation*}
\left(\tfrac{1}{2}\,\bar{\mathbf{x}}_{i,k},\ \tfrac{1}{2}\,\bar{\mathbf{x}}_{j,k},\ \bar{t}_k\right)\in\mathbb{R}^5,
\end{equation*}
where $\bar{\mathbf{x}}_{i,k}\in[0,1]^2$ is the position of the keypoint in view $i$ divided by the image width and height, and $\bar{t}_k\in[0,1]$ is the time of its frame divided by the sequence duration. The factor $\tfrac{1}{2}$ keeps the image part of the distance between two correspondences at most 1, like the time part, so that image positions do not dominate. For bundle adjustment, each view runs FPS over the normalized image coordinates, without the factor $\tfrac{1}{2}$, and time of the keypoints it detects confidently, and the per-view selections are interleaved, without duplicates, until $M$ keypoints are selected.

Confidence-only sampling (Top-P) ignores coverage: it ranks the candidate correspondences by their pair score $w_{ij,k}$ and keeps the $P$ highest. For bundle adjustment, it ranks the candidate keypoints by their mean score over the views where it exceeds $\kappa$, and keeps the $M$ highest.

\begin{figure}[tbp]
    \centering
    \includegraphics[width=\linewidth]{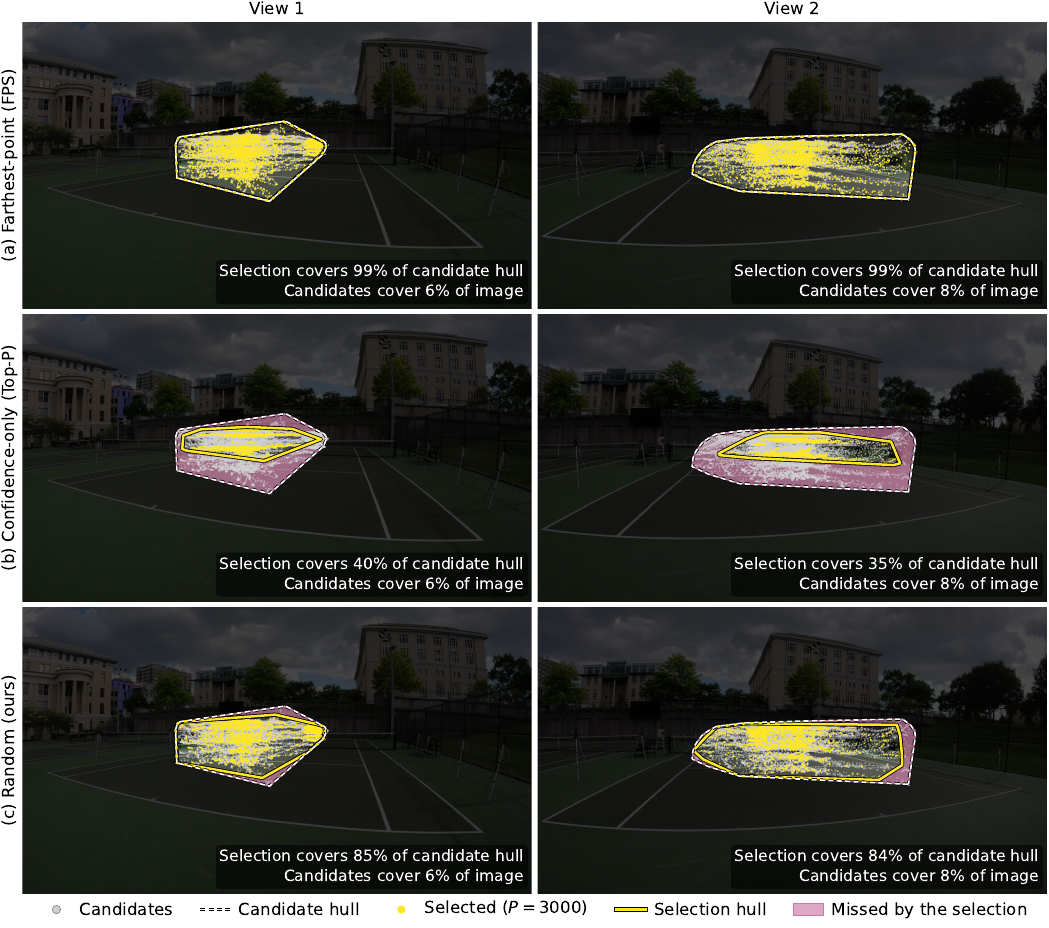}
    \caption{Correspondences selected by (a) farthest-point sampling, (b) confidence-only sampling (Top-P) and (c) our random spatio-temporal subsampling ($P=3000$) from the same confidence-filtered candidates, for the camera pair of an EgoHumans tennis sequence where the coverage of random sampling falls furthest below FPS. Whereas the convex hull of the farthest-point selection (solid) nearly matches that of the candidates (dashed), random sampling leaves the edges of the candidate region uncovered (shaded), and confidence-only sampling covers only its center.}
    \label{supp-fig:sampling_comparison}
\end{figure}

\paragraph{EgoHumans} Table~\ref{supp-tab:sampling_egohumans_categories} gives per category the comparison of Table~\ref{tab:sampling_egohumans} of the main paper, with the coverage of each strategy. In tennis, FPS also lowers the angular error from 0.52 to 0.22\textdegree{}, whereas the metric translation error barely changes, from 0.660 to 0.600\,m, as the error on the global scale factor dominates it. Top-P has a larger angular error than random sampling in every category, reaching a median angular error of 7.80\textdegree{} on tennis. The coverage of random sampling is at least 86\% in every sequence. Fig.~\ref{supp-fig:sampling_comparison} shows the tennis camera pair where it falls furthest below FPS, with 84 against 99\% and 37\% for Top-P on average over the two images; over the camera pairs of this sequence, the candidates' hull covers a median of only 11\% of the image.

Calibration accuracy also varies between categories, which differ in activity but also in scene size and camera layout. With random sampling, the scale-aligned translation error is lowest for the close-range activities (0.027\,m for Lego Assembly and 0.039\,m for Tagging) and highest for Fencing, Tennis, Badminton and Volleyball (0.130, 0.134, 0.152 and 0.199\,m), which also have the largest angular errors (0.49\textdegree{} to 0.63\textdegree{}). The large metric translation errors of Basketball, Tennis and Volleyball (0.58 to 0.66\,m) mainly come from the global scale factor, since their scale-aligned errors are 3 to 7 times lower.

\begin{table}[htbp]
\centering
\caption{Farthest-point sampling (FPS) and confidence-only sampling (Top-P) against our spatio-temporal subsampling (Random) per EgoHumans category, over the same sequences and with the same metrics as Table~\ref{tab:sampling_egohumans} of the main paper, in addition to the FoV error (in degrees) and the W-MPJPE after global alignment (GA-MPJPE, in meters). The last rows (All) are computed over all sequences. Coverage is the area of the convex hull of the selected correspondences over that of the candidates, averaged over the two images of every camera pair of a sequence. Runtime is the median time per sequence of the two sampling stages, measured separately for all strategies on the same inputs, in one process. Each cell gives the median over sequences; the best of each column is in bold and the second best is underlined.}
\maxsizebox{\linewidth}{!}{
    \begin{tabular}{ll|cccc|ccc|c|c}
        \toprule
        \multirow{2}{*}{\textbf{Category}} & \multirow{2}{*}{\textbf{Sampling}} & \multicolumn{4}{c|}{\textbf{Camera Metrics}} & \multicolumn{3}{c|}{\textbf{Human Metrics}} & \textbf{Coverage} & \multicolumn{1}{c}{\textbf{Runtime (s)}} \\
        & & TE$\downarrow$ & s-TE$\downarrow$ & AE$\downarrow$ & FoV$\downarrow$ & W-MPJPE$\downarrow$ & GA-MPJPE$\downarrow$ & PA-MPJPE$\downarrow$ & (\%)$\uparrow$ & Sampling$\downarrow$ \\
        \midrule
        \multirow{3}{*}{Badminton} & FPS & \textbf{0.216} & \textbf{0.147} & 0.49 & \underline{1.01} & \textbf{0.095} & 0.022 & 0.015 & \textbf{99.1} & 6.89  \\
         & Top-P & 0.248 & 0.177 & 0.74 & 1.05 & 0.105 & 0.030 & 0.016 & 43.7 & \textbf{4.07}  \\
         & Random (ours) & \underline{0.227} & \underline{0.152} & 0.49 & \textbf{0.99} & \underline{0.098} & 0.022 & 0.015 & \underline{94.1} & \underline{4.36}  \\
        \midrule
        \multirow{3}{*}{Basketball} & FPS & 0.655 & \underline{0.100} & \textbf{0.33} & 0.62 & 0.275 & 0.026 & 0.021 & \textbf{99.2} & 2.88  \\
         & Top-P & \textbf{0.639} & 0.103 & 0.37 & \textbf{0.61} & 0.275 & 0.027 & 0.022 & 58.6 & \textbf{1.76}  \\
         & Random (ours) & \underline{0.653} & \textbf{0.099} & \underline{0.34} & 0.62 & 0.276 & 0.026 & 0.021 & \underline{95.8} & \underline{1.83}  \\
        \midrule
        \multirow{3}{*}{Fencing} & FPS & \textbf{0.235} & \underline{0.122} & \textbf{0.59} & \underline{1.64} & \underline{0.158} & \underline{0.030} & \textbf{0.024} & \textbf{99.3} & 4.14  \\
         & Top-P & 0.270 & \textbf{0.113} & 1.13 & \textbf{1.42} & 0.205 & 0.044 & 0.025 & 42.4 & \textbf{2.60}  \\
         & Random (ours) & \underline{0.264} & 0.130 & \underline{0.63} & 1.78 & \textbf{0.153} & \textbf{0.029} & 0.025 & \underline{93.4} & \underline{2.62}  \\
        \midrule
        \multirow{3}{*}{Lego Assembly} & FPS & 0.135 & 0.027 & \textbf{0.37} & \textbf{0.46} & 0.080 & 0.027 & 0.025 & \textbf{99.3} & 2.87  \\
         & Top-P & \textbf{0.121} & 0.029 & 0.61 & 0.56 & 0.083 & 0.030 & 0.027 & 63.5 & \textbf{1.77}  \\
         & Random (ours) & \underline{0.128} & 0.027 & \underline{0.38} & \underline{0.49} & 0.080 & 0.027 & 0.025 & \underline{95.0} & \underline{1.89}  \\
        \midrule
        \multirow{3}{*}{Tagging} & FPS & \underline{0.041} & 0.039 & \textbf{0.24} & 0.53 & \underline{0.040} & 0.028 & \textbf{0.022} & \textbf{99.4} & 2.39  \\
         & Top-P & 0.045 & 0.042 & 0.33 & 0.53 & 0.046 & 0.030 & 0.025 & 59.0 & \underline{1.49}  \\
         & Random (ours) & \textbf{0.040} & 0.039 & \underline{0.25} & 0.53 & \textbf{0.039} & 0.028 & \underline{0.023} & \underline{95.1} & \textbf{1.47}  \\
        \midrule
        \multirow{3}{*}{Tennis} & FPS & \textbf{0.600} & \textbf{0.040} & \textbf{0.22} & \underline{0.20} & \textbf{0.351} & 0.015 & 0.014 & \textbf{98.6} & 12.64  \\
         & Top-P & 3.250 & 1.707 & 7.80 & 6.71 & 1.776 & 0.076 & 0.021 & 38.4 & \textbf{8.63}  \\
         & Random (ours) & \underline{0.660} & \underline{0.134} & \underline{0.52} & \textbf{0.14} & \underline{0.367} & 0.015 & 0.014 & \underline{94.2} & \underline{9.72}  \\
        \midrule
        \multirow{3}{*}{Volleyball} & FPS & \underline{0.585} & \textbf{0.185} & 0.50 & \textbf{1.10} & \textbf{0.272} & 0.026 & 0.019 & \textbf{99.1} & 7.86  \\
         & Top-P & 0.591 & 0.212 & 0.69 & 1.11 & 0.288 & 0.028 & 0.019 & 51.5 & \textbf{5.01}  \\
         & Random (ours) & \textbf{0.578} & \underline{0.199} & 0.50 & 1.11 & \underline{0.284} & 0.026 & 0.019 & \underline{92.7} & \underline{5.43}  \\
        \midrule
        \multirow{3}{*}{All} & FPS & \textbf{0.220} & \textbf{0.141} & \textbf{0.46} & \underline{0.99} & \textbf{0.099} & \textbf{0.023} & 0.016 & \textbf{99.1} & 5.28  \\
         & Top-P & 0.256 & 0.171 & 0.68 & 1.02 & 0.111 & 0.030 & 0.018 & 46.1 & \textbf{3.39}  \\
         & Random (ours) & \underline{0.229} & \underline{0.148} & \underline{0.47} & \textbf{0.98} & \underline{0.101} & \underline{0.024} & 0.016 & \underline{94.3} & \underline{3.42}  \\
        \bottomrule
    \end{tabular}
}
\label{supp-tab:sampling_egohumans_categories}
\end{table}

\paragraph{Human3.6M} Table~\ref{supp-tab:sampling_h36m} repeats the comparison on Human3.6M with RTMPose, with the settings of Table~\ref{supp-tab:sampling_settings}, on the original images and with synthetic occlusions. Top-P again has the largest camera errors, with a median scale-aligned translation error of 0.024 instead of 0.016\,m on the original images and 0.022 instead of 0.018\,m under occlusion. FPS reduces the metric translation error, from 0.130 to 0.123\,m and from 0.113 to 0.103\,m, while its scale-aligned translation error stays within 1\,mm of random sampling and its PA-MPJPE is the same, so that its gain lies in the global scale rather than in the camera geometry. Its sampling also takes 4.5 times longer on the original images, 1.99 against 0.44\,s per sequence.

\begin{table}[!hb]
\centering
\caption{Sampling strategies on Human3.6M with RTMPose, on the original images and with synthetic occlusions, with the same budget and the same confidence-filtered candidates for all strategies. Medians over the sequences of Table~\ref{tab:sampling_confidence_filter_h36m} of the main paper; the best is in bold and the second best underlined. TE, s-TE, W-MPJPE, GA-MPJPE and PA-MPJPE are in meters, and AE and FoV in degrees. Runtime covers the two sampling stages.}
\newcolumntype{H}{>{\setbox0\hbox\bgroup}c<{\egroup}@{}}
\maxsizebox{\linewidth}{!}{
    \begin{tabular}{ll|cccH|cHc|c}
        \toprule
        \multirow{2}{*}{\textbf{Input}} & \multirow{2}{*}{\textbf{Sampling}} & \multicolumn{4}{c|}{\textbf{Camera Metrics}} & \multicolumn{3}{c|}{\textbf{Human Metrics}} & \multicolumn{1}{c}{\textbf{Runtime (s)}} \\
        & & TE$\downarrow$ & s-TE$\downarrow$ & AE$\downarrow$ & FoV$\downarrow$ & W-MPJPE$\downarrow$ & GA-MPJPE$\downarrow$ & PA-MPJPE$\downarrow$ & Sampling$\downarrow$ \\
        \midrule
        \multirow{3}{*}{Original} & FPS & \textbf{0.123} & \underline{0.017} & \textbf{0.81} & \textbf{0.22} & \textbf{0.035} & 0.022 & 0.022 & 1.99 \\
         & Top-P & 0.165 & 0.024 & 0.83 & 0.32 & 0.046 & 0.022 & 0.022 & \underline{0.46} \\
         & Random (ours) & \underline{0.130} & \textbf{0.016} & \underline{0.82} & \underline{0.25} & \underline{0.039} & 0.022 & 0.022 & \textbf{0.44} \\
        \midrule
        \multirow{3}{*}{Occluded} & FPS & \textbf{0.103} & 0.018 & 0.83 & 0.34 & \textbf{0.048} & 0.031 & 0.031 & 1.99 \\
         & Top-P & 0.136 & 0.022 & 0.84 & 0.46 & 0.054 & 0.031 & 0.031 & 0.46 \\
         & Random (ours) & \underline{0.113} & 0.018 & 0.83 & 0.34 & \underline{0.049} & 0.031 & 0.031 & 0.46 \\
        \bottomrule
    \end{tabular}
}
\label{supp-tab:sampling_h36m}
\end{table}%

\subsection{Impact of Keypoint Confidence}
\label{supp-sec:confidence_ablation}

Table~\ref{supp-tab:confidence_weights_full} extends Table~\ref{tab:confidence_weights_off} of the main paper to NLF on Human3.6M and to NLF and DWPose on EgoHumans. Without the weights, every detected keypoint has a weight of one (rows \emph{Filters}); without the filters as well, every score is set to one, so that correspondences are sampled uniformly among all candidates (rows \emph{None}). The weights lower the PA-MPJPE in every sequence of every setting, by a median of 0.4\,mm with NLF on the original Human3.6M images and 3.3\,mm under occlusion, and by 0.6 and 3.1\,mm on EgoHumans with NLF and DWPose, while the medians of the camera errors change by at most 1\,mm and 0.02\textdegree{}. The 90th percentile of the angular error decreases on EgoHumans, from 0.68 to 0.57\textdegree{} with NLF and from 0.81 to 0.68\textdegree{} with DWPose, and changes by at most 0.06\textdegree{} on Human3.6M. Removing the filters as well does not change any metric consistently across sequences with NLF, on Human3.6M as on EgoHumans; with DWPose, it raises the PA-MPJPE in most sequences, by a median of 0.2\,mm.

\begin{table}[htbp]
\centering
\caption{Effect of the detection confidence with NLF and DWPose, completing Table~\ref{tab:confidence_weights_off} of the main paper. \emph{Filters + weights} is our pipeline, \emph{Filters} gives every detected keypoint a weight of one, and \emph{None} also sets every score to one. Medians over the sequences reconstructed by all runs of a detector and input, and 90th percentile (p90) of the angular error; values are not comparable across datasets, whose configurations differ. The best of each block is in bold and the second best underlined; a tie for the best marks neither.}
\maxsizebox{\linewidth}{!}{
    \begin{tabular}{llll|cccc|c}
        \toprule
        \multirow{2}{*}{\textbf{Dataset}} & \multirow{2}{*}{\textbf{Detector}} & \multirow{2}{*}{\textbf{Input}} & \multirow{2}{*}{\textbf{Confidence}} & \multicolumn{4}{c|}{\textbf{Camera Metrics}} & \textbf{Human Metrics} \\
        & & & & s-TE (m)$\downarrow$ & AE (\textdegree)$\downarrow$ & AE p90$\downarrow$ & FoV (\textdegree)$\downarrow$ & PA-MPJPE (m)$\downarrow$ \\
        \midrule
        \multirow{3}{*}{Human3.6M} & \multirow{3}{*}{NLF} & \multirow{3}{*}{Original} & Filters + weights & 0.014 & \textbf{0.82} & 1.06 & 0.28 & \textbf{0.018} \\
          &  &  & Filters & 0.014 & \underline{0.83} & \underline{1.04} & 0.27 & \underline{0.019} \\
          &  &  & None & 0.015 & \underline{0.83} & \textbf{1.01} & 0.27 & \underline{0.019} \\
        \midrule
        \multirow{3}{*}{Human3.6M} & \multirow{3}{*}{NLF} & \multirow{3}{*}{Occluded} & Filters + weights & \textbf{0.013} & \textbf{0.83} & 1.09 & 0.32 & \textbf{0.021} \\
          &  &  & Filters & \underline{0.014} & 0.85 & 1.15 & 0.32 & \underline{0.025} \\
          &  &  & None & 0.018 & \underline{0.84} & 1.09 & 0.37 & \underline{0.025} \\
        \midrule
        \multirow{3}{*}{EgoHumans} & \multirow{3}{*}{NLF} & \multirow{3}{*}{Original} & Filters + weights & \textbf{0.147} & \textbf{0.46} & \textbf{0.57} & 0.98 & \textbf{0.016} \\
          &  &  & Filters & \underline{0.148} & \underline{0.47} & \underline{0.68} & 0.99 & \underline{0.017} \\
          &  &  & None & \underline{0.148} & \underline{0.47} & 0.77 & 0.98 & \underline{0.017} \\
        \midrule
        \multirow{3}{*}{EgoHumans} & \multirow{3}{*}{DWPose} & \multirow{3}{*}{Original} & Filters + weights & 0.149 & 0.57 & \textbf{0.68} & \textbf{1.02} & \textbf{0.018} \\
          &  &  & Filters & 0.148 & 0.57 & 0.81 & \underline{1.04} & \underline{0.021} \\
          &  &  & None & 0.148 & 0.57 & \underline{0.80} & \underline{1.04} & \underline{0.021} \\
        \bottomrule
    \end{tabular}
}
\label{supp-tab:confidence_weights_full}
\end{table}

\subsection{Metric Depth-Based Scale Estimation}
\label{supp-sec:depth_scale_analysis}

With ground-truth cameras and keypoints, the scale factor of Sec.~\ref{sec:depth_scale} of the main paper is the mean ratio between the MoGe-2 depths and the ground-truth joint depths, whose correct value is 1. On Human3.6M, it is too small in all sequences, by 12 to 16\% (median 0.857). On EgoHumans, the median scale is 1.02, but it varies from 0.86 to 1.69 between sequences. A scale error displaces each joint in proportion to its distance from the center of the camera setup, which is why the W-MPJPE is larger in the wider EgoHumans scenes.

\subsection{Impact of the Number of Views}
\label{supp-sec:n_views}

Table~\ref{supp-tab:n_views} details the evaluation of Sec.~\ref{sec:ablation_study_n_views} and Fig.~\ref{fig:n_views} of the main paper. For each EgoHumans sequence, the two- and eight-view camera sets are taken from HSfM, with the exception of unavailable or moving cameras replaced by the nearest available static camera. The sets in between are nested: each adds the camera of the eight-view set farthest from those already selected, by the ground-truth camera centers. With estimated intrinsics, the pipeline matches the main experiments, with intrinsics shared by all cameras. With ground-truth intrinsics, the focal length, principal point and distortion coefficients of each camera are taken from the ground truth and kept fixed, NLF detects the keypoints with them, and a single bundle adjustment pass refines the extrinsics.

With two views, ground-truth intrinsics lower the mean PA-MPJPE from 130.7 to 113.4\,mm and the median angular error from 1.74 to 0.31\textdegree{}. From three views on, they still lower the angular error, but the mean PA-MPJPE is up to 1.7\,mm higher than with estimated intrinsics, as a few sequences are reconstructed less accurately.

\begin{table}[htbp]
\centering
\caption{PA-MPJPE (mean $\pm$ standard deviation and median, in millimeters) and median angular error (AE, in degrees) against the number of views, on the EgoHumans sequences of Fig.~\ref{fig:n_views} of the main paper, with estimated and ground-truth intrinsics. HSfM gives the mean PA-MPJPE reported in its own evaluation, to the centimeter.}
\maxsizebox{\linewidth}{!}{
\begin{tabular}{l|ccc|ccc|c}
    \toprule
    & \multicolumn{3}{c|}{\textbf{Estimated intrinsics}} & \multicolumn{3}{c|}{\textbf{Ground-truth intrinsics}} & \textbf{HSfM} \\
    \textbf{Views} & PA-MPJPE$\downarrow$ & Median$\downarrow$ & AE$\downarrow$ & PA-MPJPE$\downarrow$ & Median$\downarrow$ & AE$\downarrow$ & PA-MPJPE$\downarrow$ \\
    \midrule
    2 & 130.7 $\pm$ 65.0 & 131.2 & 1.74 & 113.4 $\pm$ 60.3 & 115.6 & 0.31 & 50 \\
    3 & 40.4 $\pm$ 21.4 & 36.6 & 0.61 & 40.5 $\pm$ 29.0 & 35.9 & 0.26 & -- \\
    4 & 23.0 $\pm$ 12.8 & 18.6 & 0.59 & 24.0 $\pm$ 18.1 & 18.7 & 0.26 & 60 \\
    5 & 20.9 $\pm$ 11.2 & 17.0 & 0.47 & 21.7 $\pm$ 15.1 & 17.1 & 0.23 & -- \\
    6 & 19.2 $\pm$ 5.2 & 16.6 & 0.40 & 20.0 $\pm$ 7.6 & 16.8 & 0.23 & -- \\
    7 & 18.7 $\pm$ 5.0 & 16.0 & 0.36 & 19.7 $\pm$ 7.8 & 16.4 & 0.21 & -- \\
    8 & 18.2 $\pm$ 4.6 & 15.8 & 0.35 & 19.2 $\pm$ 7.3 & 16.0 & 0.21 & 50 \\
    \midrule
    All & 18.0 $\pm$ 5.0 & 15.9 & 0.46 & 19.7 $\pm$ 10.2 & 15.4 & 0.23 & -- \\
    \bottomrule
\end{tabular}
}
\label{supp-tab:n_views}
\end{table}

\section{Extended Results on Human3.6M}
\label{supp-sec:h36m}

This section presents extended quantitative and qualitative results on the Human3.6M dataset. Table~\ref{supp-tab:h36m_quantitative_full} provides detailed quantitative and qualitative comparisons of Kineo against the state-of-the-art calibration-free method HSfM, highlighting improvements in both 3D human reconstruction and camera parameter estimation. It extends the Human3.6M rows of Table~\ref{tab:human_metrics} and Fig.~\ref{fig:h36m_qualitative_comparison} of the main paper to the 47 sequences, out of 54, for which HSfM returns a reconstruction, leaving out four frames of three sequences that it does not reconstruct. Kineo estimates all camera parameters, including distortion, with NLF keypoints and SMPL-based scaling. Its mean W-MPJPE per sequence ranges from 0.016 to 0.085\,m, without a clear trend by action: the ranges of the different actions largely overlap.

\begin{longtblr}[
  label = {supp-tab:h36m_quantitative_full},
  caption = {Comparison of the W-MPJPE metric (lower is better) across Human3.6M sequences for Kineo (\textbf{\textcolor{cyan}{blue}}), HSfM (\textbf{\textcolor{red}{red}}), and ground truth (\textbf{\textcolor{black}{black}}). For each sequence, the table reports the minimum, median, maximum, mean, and standard deviation of W-MPJPE over all frames, along with visualizations of the corresponding frames. All estimates are scale-aligned for visualization. Overall, Kineo achieves lower W-MPJPE and demonstrates greater robustness with smaller standard deviation compared to HSfM.},
]{
width=\textwidth,
cells={valign=m,halign=c},
colspec={Q[c,1.8cm] *{3}{X[c]} Q[c,1.8cm] *{3}{X[c]} Q[c,1.8cm]},
hline{1,Z} = {1pt},
hline{3} = {0.4pt},
vline{2-9} = {0.4pt}, %
rowhead=2,
column{2-4,6-8} = {colsep=3pt}
}
\SetCell[r=2]{c} \textbf{Sequence} &
\SetCell[c=4]{c} \textbf{HSfM W-MPJPE} & & & &
\SetCell[c=4]{c} \textbf{Kineo W-MPJPE} & & & \\

& Min\,$\downarrow$ & Median\,$\downarrow$ & Max\,$\downarrow$ & Mean$\pm$Std\,$\downarrow$ &
  Min\,$\downarrow$ & Median\,$\downarrow$ & Max\,$\downarrow$ & Mean$\pm$Std\,$\downarrow$ \\

\small{S9\_\allowbreak Directions 1} & 
{0.144\\\includegraphics[width=\linewidth]{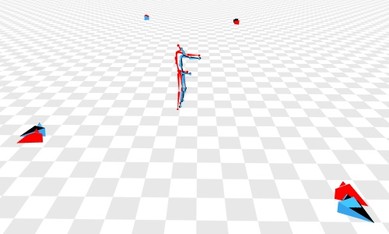}} & 
{0.283\\\includegraphics[width=\linewidth]{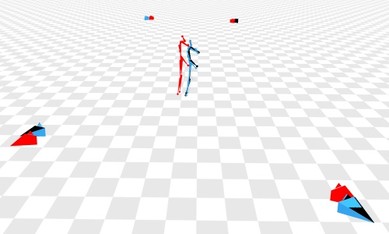}} & 
{13.112\\\includegraphics[width=\linewidth]{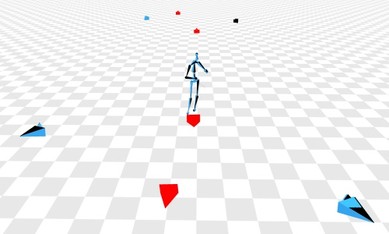}} & 
0.693$\pm$2.007 & 
{0.026\\\includegraphics[width=\linewidth]{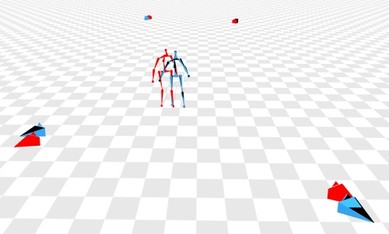}} & 
{0.036\\\includegraphics[width=\linewidth]{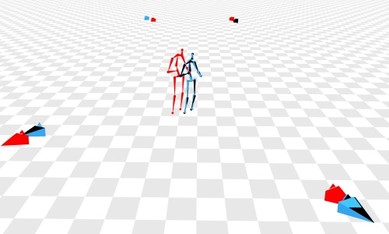}} & 
{0.066\\\includegraphics[width=\linewidth]{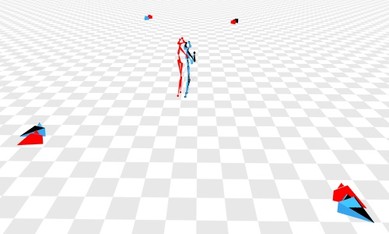}} & 
0.036$\pm$0.006 \\
\hline
\small{S11\_\allowbreak Directions 1} & 
{0.122\\\includegraphics[width=\linewidth]{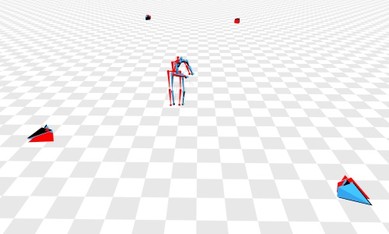}} & 
{0.167\\\includegraphics[width=\linewidth]{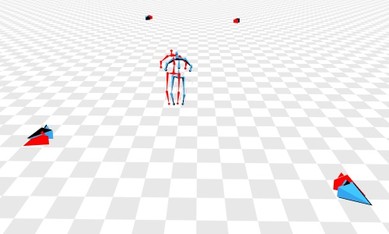}} & 
{11.841\\\includegraphics[width=\linewidth]{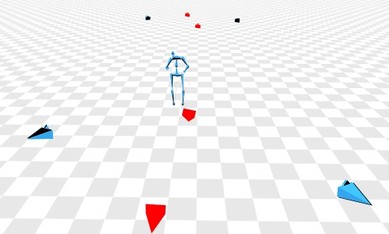}} & 
0.450$\pm$1.375 & 
{0.046\\\includegraphics[width=\linewidth]{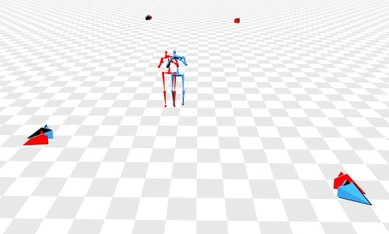}} & 
{0.050\\\includegraphics[width=\linewidth]{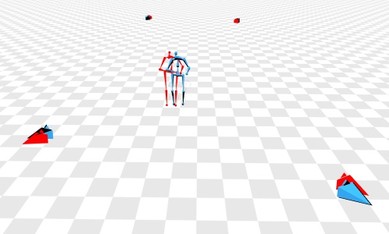}} & 
{0.053\\\includegraphics[width=\linewidth]{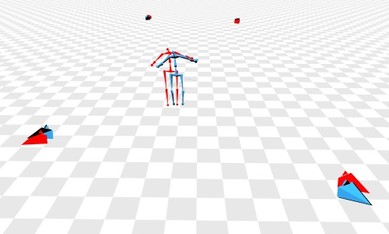}} & 
0.050$\pm$0.001 \\
\hline
\small{S9\_\allowbreak Directions} & 
{0.138\\\includegraphics[width=\linewidth]{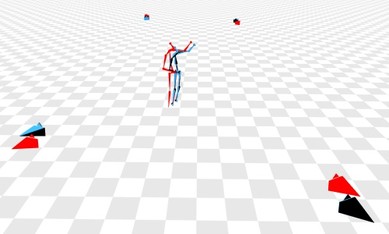}} & 
{0.270\\\includegraphics[width=\linewidth]{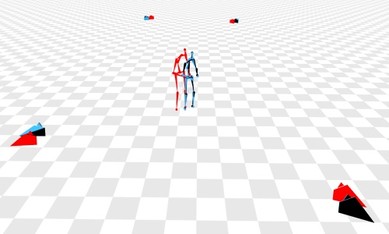}} & 
{12.893\\\includegraphics[width=\linewidth]{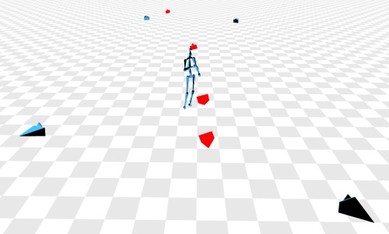}} & 
1.679$\pm$3.633 & 
{0.027\\\includegraphics[width=\linewidth]{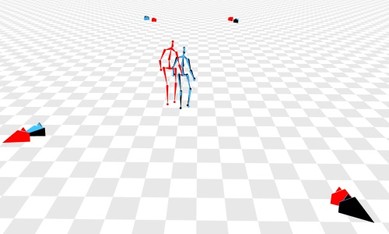}} & 
{0.036\\\includegraphics[width=\linewidth]{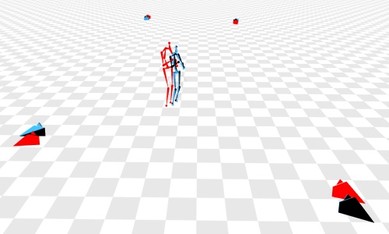}} & 
{0.058\\\includegraphics[width=\linewidth]{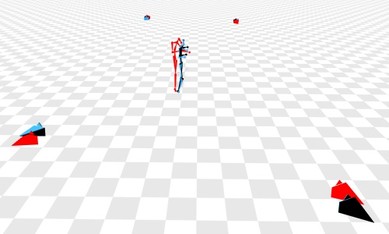}} & 
0.037$\pm$0.005 \\
\hline
\small{S11\_\allowbreak Directions} & 
{0.143\\\includegraphics[width=\linewidth]{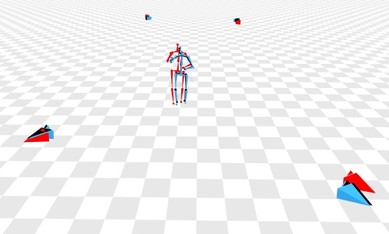}} & 
{0.208\\\includegraphics[width=\linewidth]{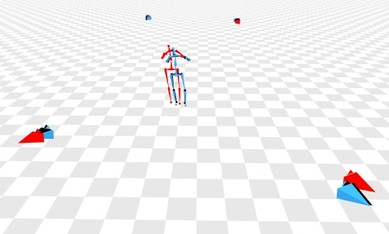}} & 
{3.127\\\includegraphics[width=\linewidth]{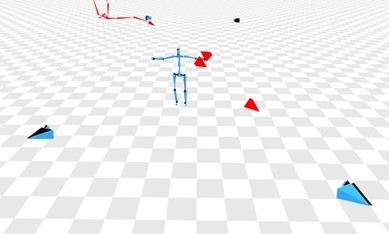}} & 
0.225$\pm$0.207 & 
{0.036\\\includegraphics[width=\linewidth]{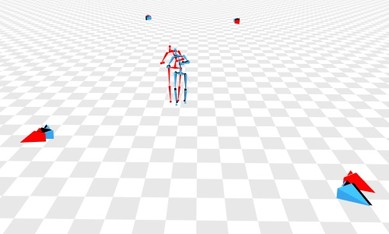}} & 
{0.040\\\includegraphics[width=\linewidth]{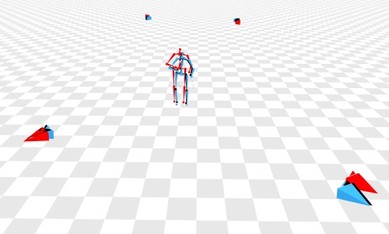}} & 
{0.048\\\includegraphics[width=\linewidth]{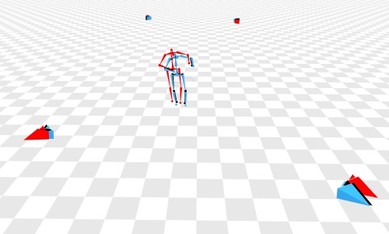}} & 
0.040$\pm$0.002 \\
\hline
\small{S9\_\allowbreak Discussion 2} & 
{0.075\\\includegraphics[width=\linewidth]{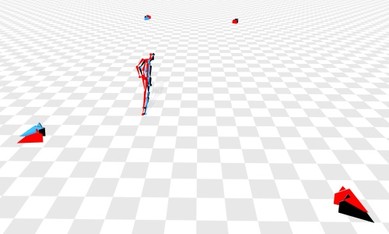}} & 
{0.252\\\includegraphics[width=\linewidth]{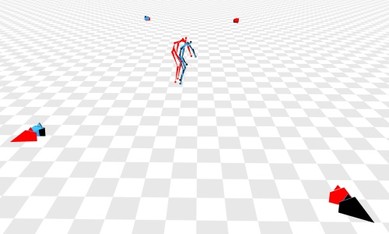}} & 
{12.641\\\includegraphics[width=\linewidth]{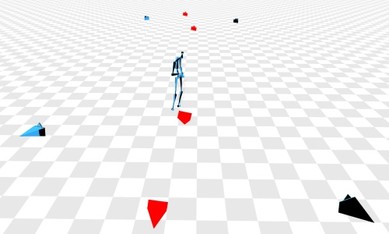}} & 
0.694$\pm$1.829 & 
{0.028\\\includegraphics[width=\linewidth]{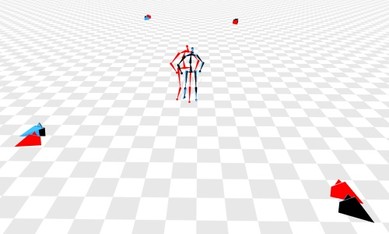}} & 
{0.042\\\includegraphics[width=\linewidth]{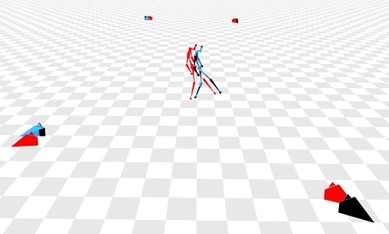}} & 
{0.077\\\includegraphics[width=\linewidth]{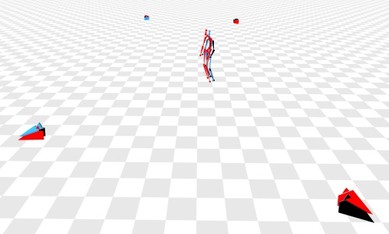}} & 
0.044$\pm$0.010 \\
\hline
\small{S11\_\allowbreak Discussion 2} & 
{0.103\\\includegraphics[width=\linewidth]{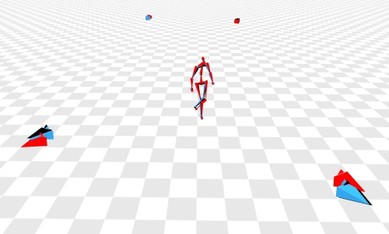}} & 
{0.172\\\includegraphics[width=\linewidth]{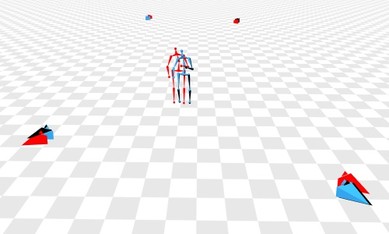}} & 
{12.000\\\includegraphics[width=\linewidth]{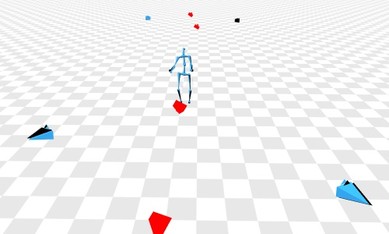}} & 
0.670$\pm$1.832 & 
{0.021\\\includegraphics[width=\linewidth]{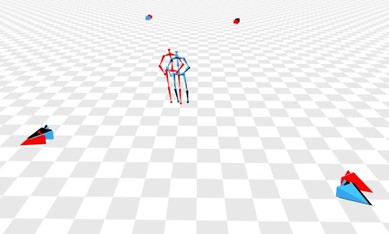}} & 
{0.027\\\includegraphics[width=\linewidth]{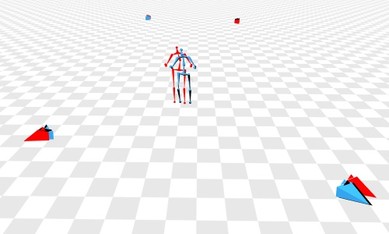}} & 
{0.042\\\includegraphics[width=\linewidth]{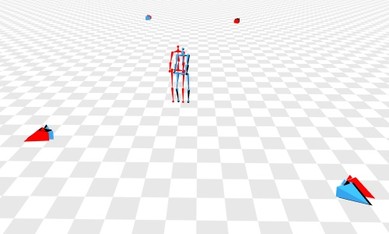}} & 
0.028$\pm$0.004 \\
\hline
\small{S9\_\allowbreak Eating 1} & 
{0.083\\\includegraphics[width=\linewidth]{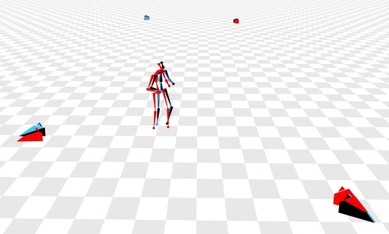}} & 
{0.246\\\includegraphics[width=\linewidth]{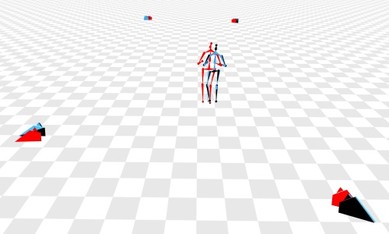}} & 
{4.316\\\includegraphics[width=\linewidth]{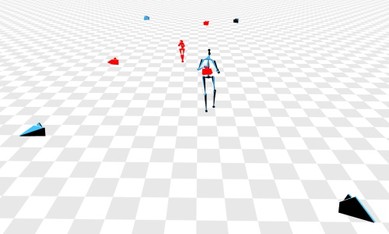}} & 
0.261$\pm$0.247 & 
{0.036\\\includegraphics[width=\linewidth]{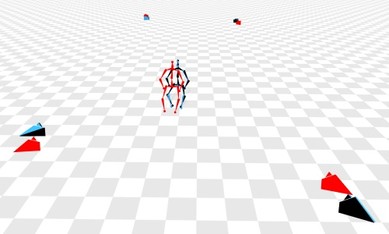}} & 
{0.045\\\includegraphics[width=\linewidth]{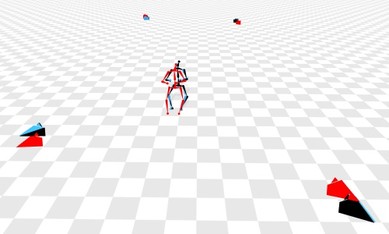}} & 
{0.091\\\includegraphics[width=\linewidth]{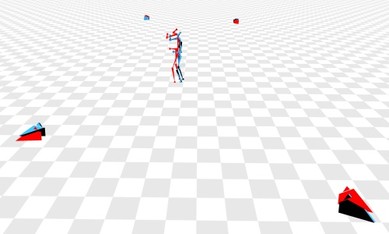}} & 
0.051$\pm$0.014 \\
\hline
\small{S11\_\allowbreak Eating 1} & 
{0.076\\\includegraphics[width=\linewidth]{figs/min_max_median/S11_Eating_1_hsfm_min.jpg}} & 
{0.136\\\includegraphics[width=\linewidth]{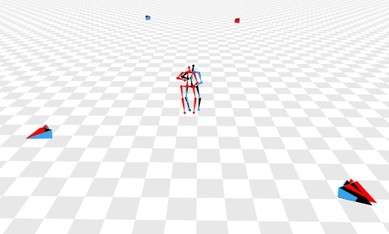}} & 
{3.761\\\includegraphics[width=\linewidth]{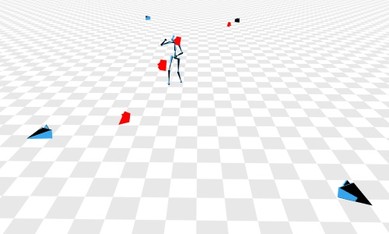}} & 
0.165$\pm$0.276 & 
{0.017\\\includegraphics[width=\linewidth]{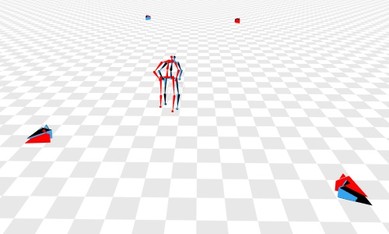}} & 
{0.024\\\includegraphics[width=\linewidth]{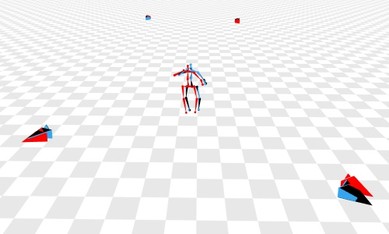}} & 
{0.037\\\includegraphics[width=\linewidth]{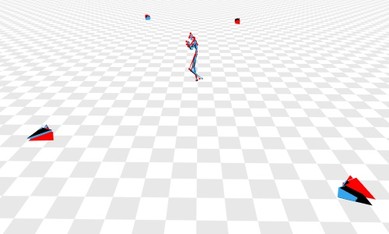}} & 
0.024$\pm$0.003 \\
\hline
\small{S9\_\allowbreak Eating} & 
{0.083\\\includegraphics[width=\linewidth]{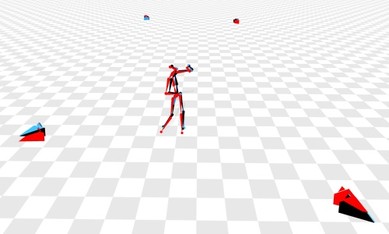}} & 
{0.231\\\includegraphics[width=\linewidth]{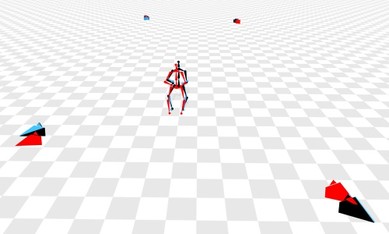}} & 
{4.559\\\includegraphics[width=\linewidth]{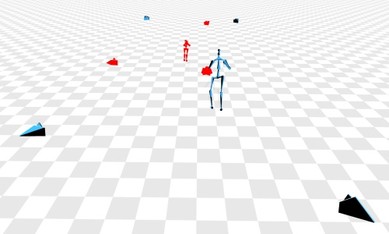}} & 
0.248$\pm$0.262 & 
{0.035\\\includegraphics[width=\linewidth]{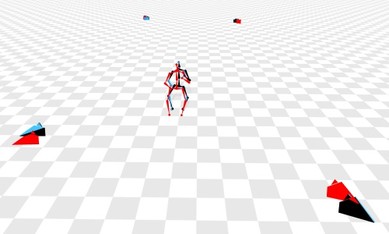}} & 
{0.044\\\includegraphics[width=\linewidth]{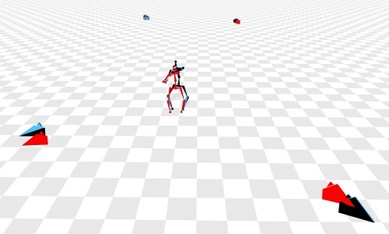}} & 
{0.086\\\includegraphics[width=\linewidth]{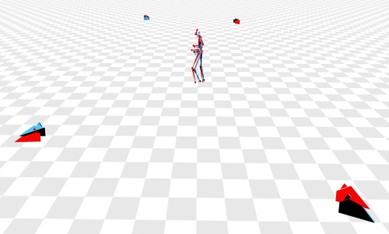}} & 
0.048$\pm$0.012 \\
\hline
\small{S11\_\allowbreak Eating} & 
{0.061\\\includegraphics[width=\linewidth]{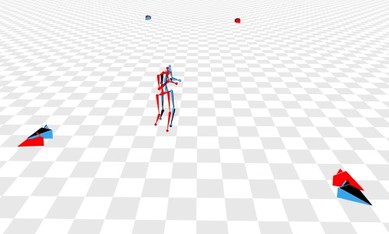}} & 
{0.137\\\includegraphics[width=\linewidth]{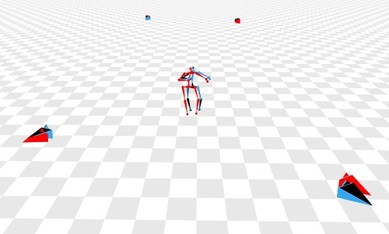}} & 
{9.391\\\includegraphics[width=\linewidth]{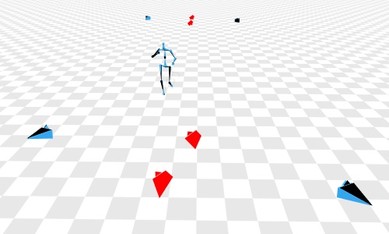}} & 
0.198$\pm$0.546 & 
{0.018\\\includegraphics[width=\linewidth]{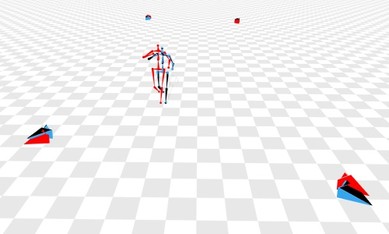}} & 
{0.027\\\includegraphics[width=\linewidth]{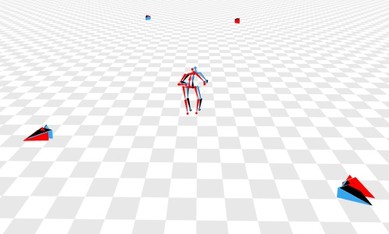}} & 
{0.041\\\includegraphics[width=\linewidth]{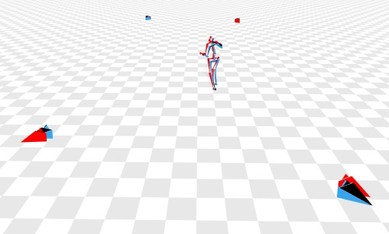}} & 
0.027$\pm$0.004 \\
\hline
\small{S11\_\allowbreak Greeting 2} & 
{0.080\\\includegraphics[width=\linewidth]{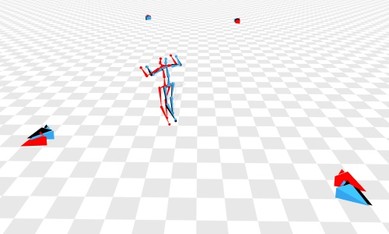}} & 
{0.171\\\includegraphics[width=\linewidth]{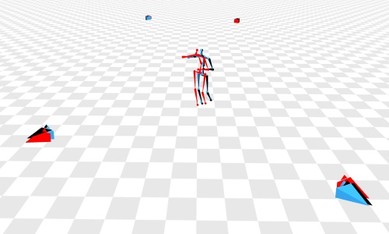}} & 
{12.049\\\includegraphics[width=\linewidth]{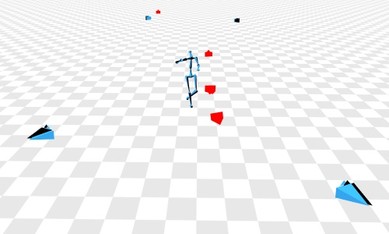}} & 
0.857$\pm$2.343 & 
{0.018\\\includegraphics[width=\linewidth]{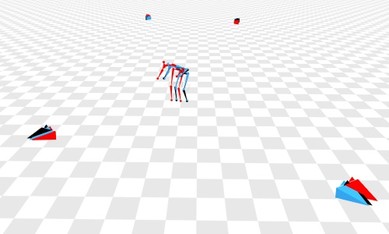}} & 
{0.025\\\includegraphics[width=\linewidth]{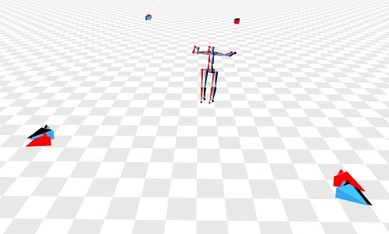}} & 
{0.039\\\includegraphics[width=\linewidth]{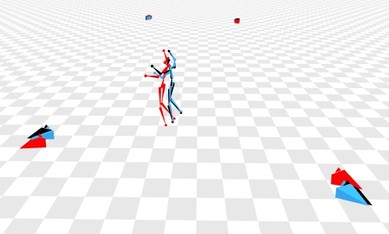}} & 
0.026$\pm$0.004 \\
\hline
\small{S11\_\allowbreak Greeting} & 
{0.083\\\includegraphics[width=\linewidth]{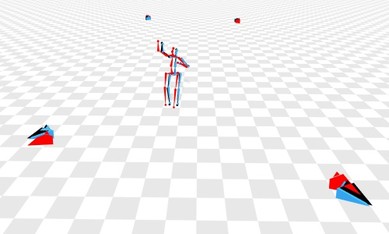}} & 
{0.172\\\includegraphics[width=\linewidth]{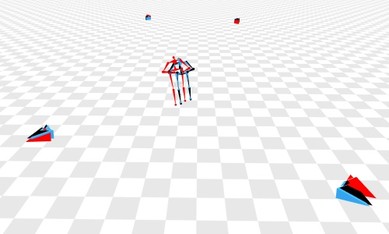}} & 
{10.481\\\includegraphics[width=\linewidth]{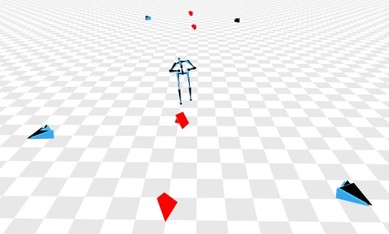}} & 
0.462$\pm$1.236 & 
{0.014\\\includegraphics[width=\linewidth]{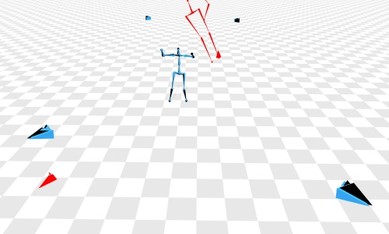}} & 
{0.020\\\includegraphics[width=\linewidth]{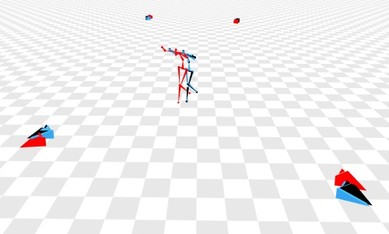}} & 
{0.047\\\includegraphics[width=\linewidth]{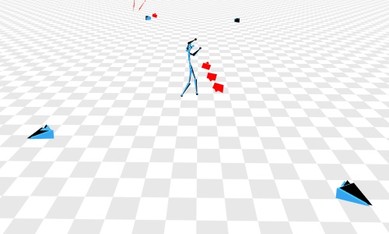}} & 
0.021$\pm$0.004 \\
\hline
\small{S9\_\allowbreak Phoning 1} & 
{0.091\\\includegraphics[width=\linewidth]{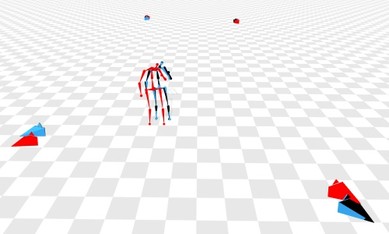}} & 
{0.200\\\includegraphics[width=\linewidth]{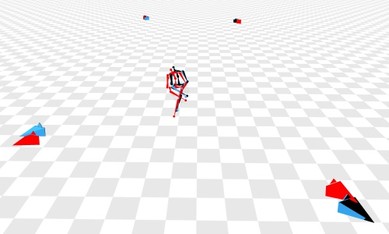}} & 
{4.507\\\includegraphics[width=\linewidth]{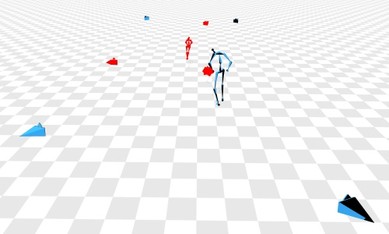}} & 
0.223$\pm$0.342 & 
{0.030\\\includegraphics[width=\linewidth]{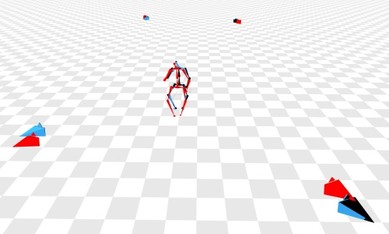}} & 
{0.043\\\includegraphics[width=\linewidth]{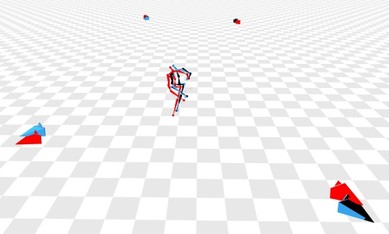}} & 
{0.064\\\includegraphics[width=\linewidth]{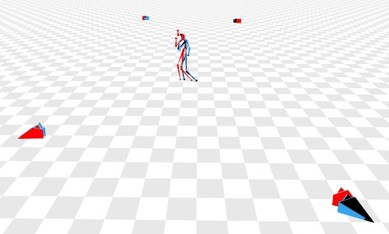}} & 
0.044$\pm$0.007 \\
\hline
\small{S11\_\allowbreak Phoning 3} & 
{0.067\\\includegraphics[width=\linewidth]{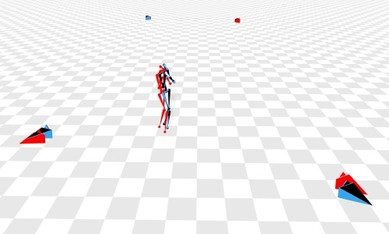}} & 
{0.163\\\includegraphics[width=\linewidth]{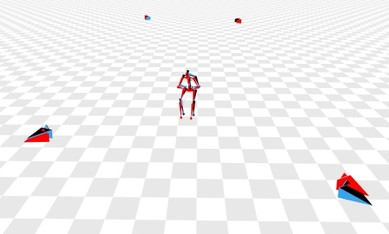}} & 
{7.464\\\includegraphics[width=\linewidth]{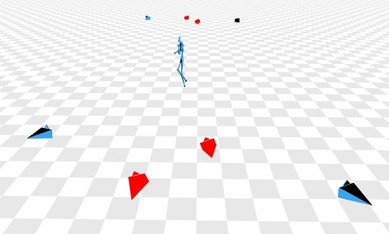}} & 
0.252$\pm$0.478 & 
{0.014\\\includegraphics[width=\linewidth]{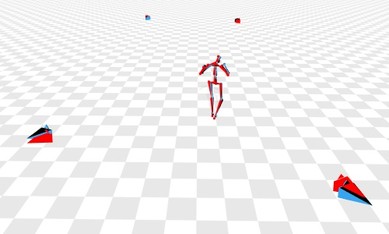}} & 
{0.020\\\includegraphics[width=\linewidth]{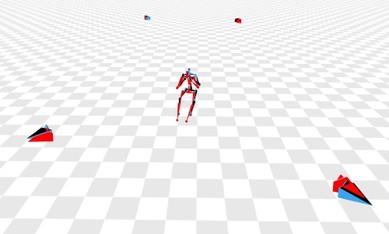}} & 
{0.041\\\includegraphics[width=\linewidth]{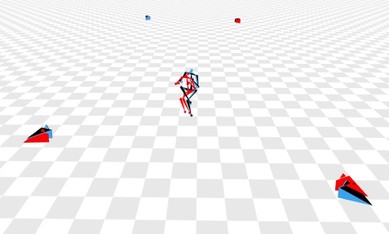}} & 
0.021$\pm$0.004 \\
\hline
\small{S9\_\allowbreak Phoning} & 
{0.090\\\includegraphics[width=\linewidth]{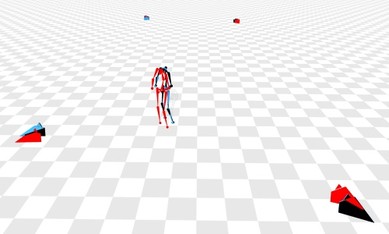}} & 
{0.192\\\includegraphics[width=\linewidth]{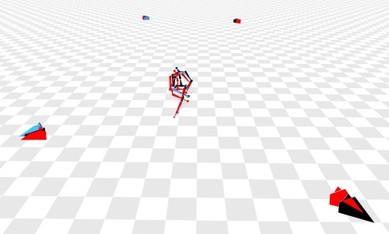}} & 
{4.523\\\includegraphics[width=\linewidth]{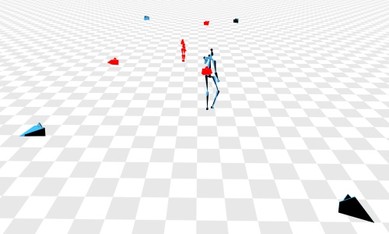}} & 
0.253$\pm$0.461 & 
{0.029\\\includegraphics[width=\linewidth]{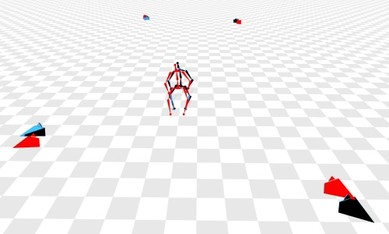}} & 
{0.040\\\includegraphics[width=\linewidth]{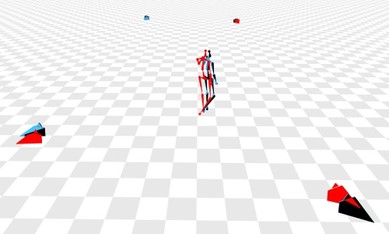}} & 
{0.063\\\includegraphics[width=\linewidth]{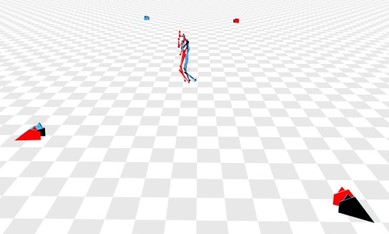}} & 
0.042$\pm$0.006 \\
\hline
\small{S11\_\allowbreak Phoning 2} & 
{0.047\\\includegraphics[width=\linewidth]{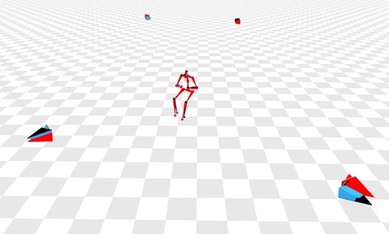}} & 
{0.142\\\includegraphics[width=\linewidth]{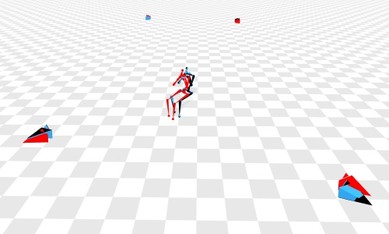}} & 
{4.623\\\includegraphics[width=\linewidth]{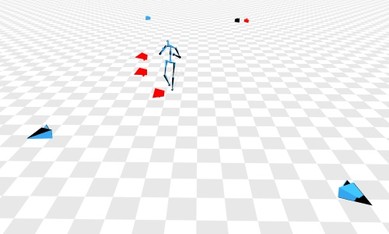}} & 
0.188$\pm$0.255 & 
{0.014\\\includegraphics[width=\linewidth]{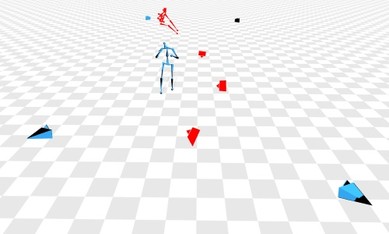}} & 
{0.019\\\includegraphics[width=\linewidth]{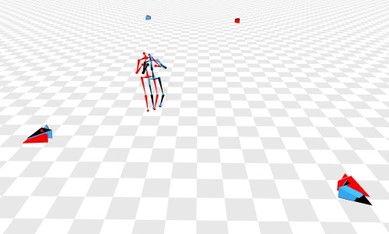}} & 
{0.029\\\includegraphics[width=\linewidth]{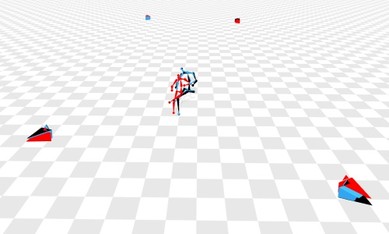}} & 
0.019$\pm$0.003 \\
\hline
\small{S9\_\allowbreak Posing 1} & 
{0.145\\\includegraphics[width=\linewidth]{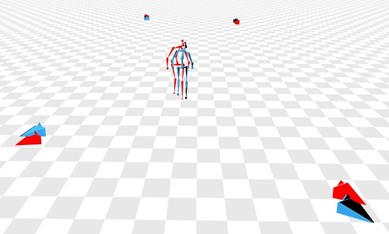}} & 
{0.282\\\includegraphics[width=\linewidth]{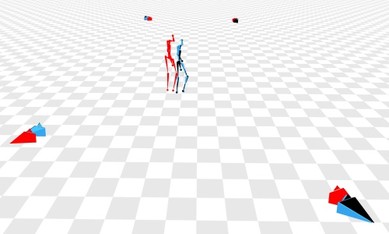}} & 
{12.729\\\includegraphics[width=\linewidth]{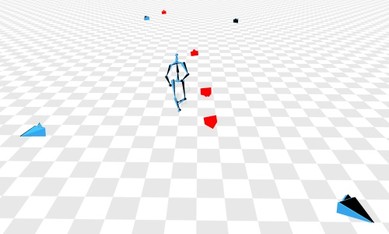}} & 
0.858$\pm$2.445 & 
{0.029\\\includegraphics[width=\linewidth]{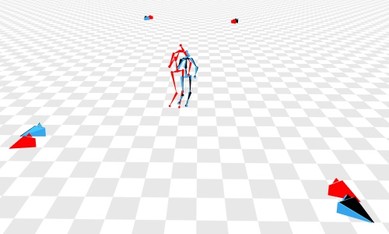}} & 
{0.036\\\includegraphics[width=\linewidth]{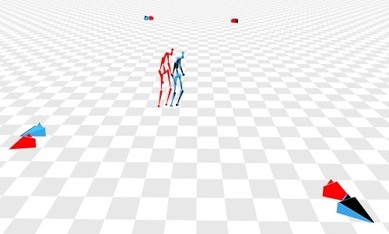}} & 
{0.047\\\includegraphics[width=\linewidth]{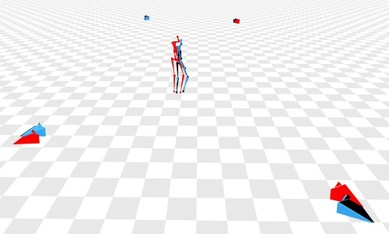}} & 
0.036$\pm$0.004 \\
\hline
\small{S11\_\allowbreak Posing 1} & 
{0.080\\\includegraphics[width=\linewidth]{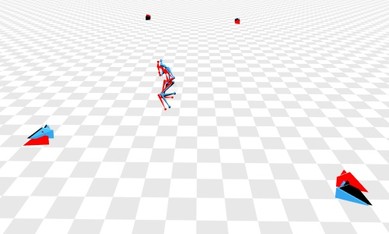}} & 
{0.154\\\includegraphics[width=\linewidth]{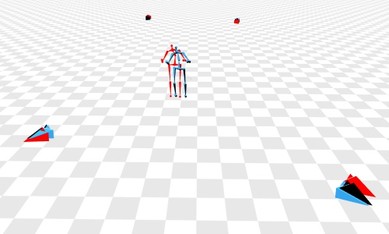}} & 
{4.729\\\includegraphics[width=\linewidth]{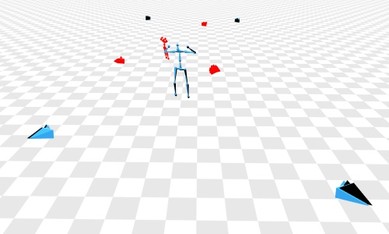}} & 
0.449$\pm$0.874 & 
{0.021\\\includegraphics[width=\linewidth]{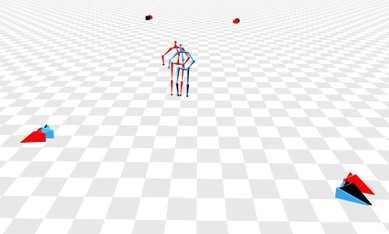}} & 
{0.027\\\includegraphics[width=\linewidth]{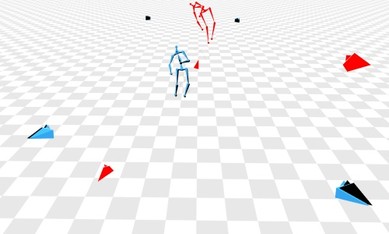}} & 
{0.037\\\includegraphics[width=\linewidth]{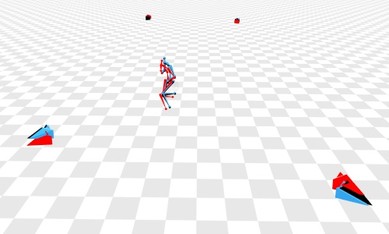}} & 
0.028$\pm$0.003 \\
\hline
\small{S9\_\allowbreak Posing} & 
{0.130\\\includegraphics[width=\linewidth]{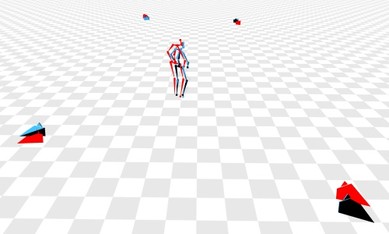}} & 
{0.291\\\includegraphics[width=\linewidth]{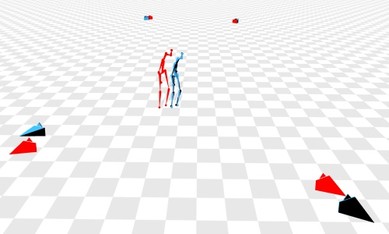}} & 
{11.036\\\includegraphics[width=\linewidth]{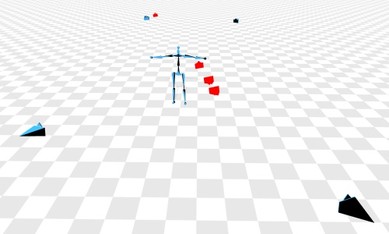}} & 
0.563$\pm$1.171 & 
{0.031\\\includegraphics[width=\linewidth]{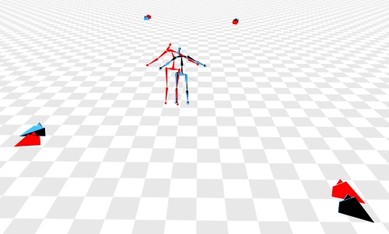}} & 
{0.037\\\includegraphics[width=\linewidth]{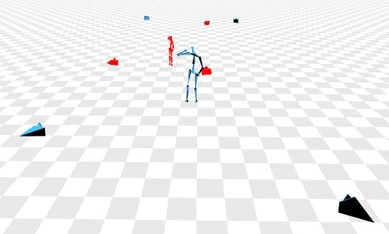}} & 
{0.059\\\includegraphics[width=\linewidth]{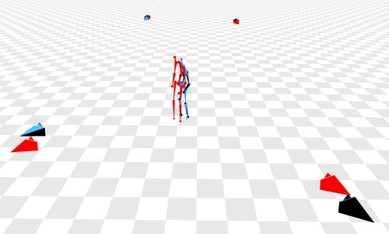}} & 
0.039$\pm$0.006 \\
\hline
\small{S11\_\allowbreak Posing} & 
{0.106\\\includegraphics[width=\linewidth]{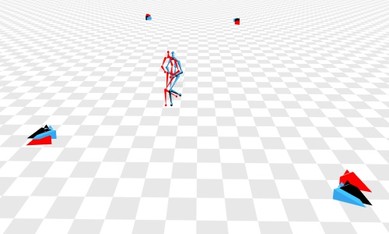}} & 
{0.170\\\includegraphics[width=\linewidth]{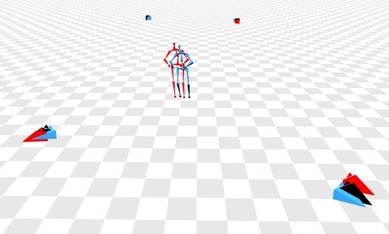}} & 
{5.765\\\includegraphics[width=\linewidth]{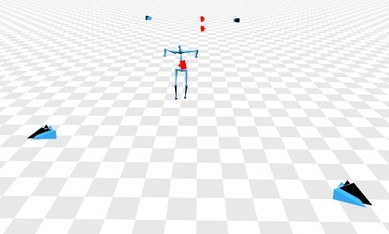}} & 
0.684$\pm$1.242 & 
{0.025\\\includegraphics[width=\linewidth]{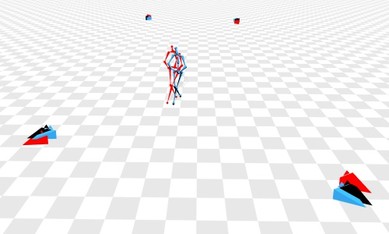}} & 
{0.030\\\includegraphics[width=\linewidth]{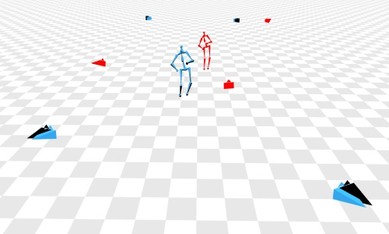}} & 
{0.039\\\includegraphics[width=\linewidth]{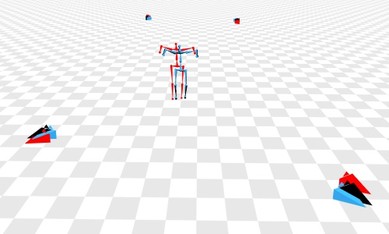}} & 
0.031$\pm$0.003 \\
\hline
\small{S9\_\allowbreak Purchases 1} & 
{0.131\\\includegraphics[width=\linewidth]{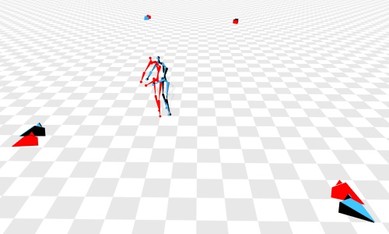}} & 
{0.232\\\includegraphics[width=\linewidth]{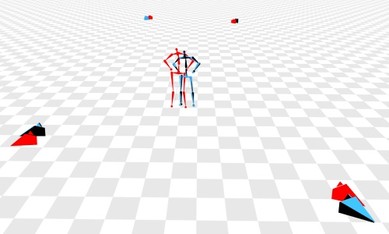}} & 
{4.635\\\includegraphics[width=\linewidth]{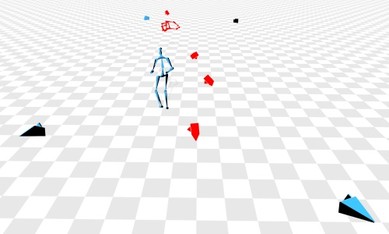}} & 
0.285$\pm$0.318 & 
{0.032\\\includegraphics[width=\linewidth]{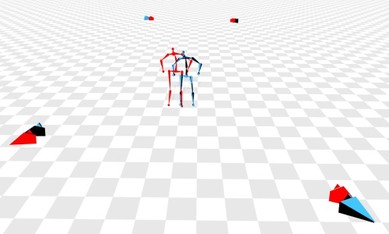}} & 
{0.038\\\includegraphics[width=\linewidth]{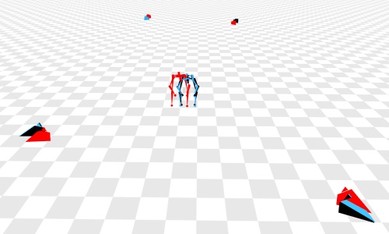}} & 
{0.086\\\includegraphics[width=\linewidth]{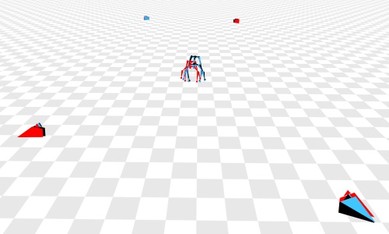}} & 
0.042$\pm$0.011 \\
\hline
\small{S11\_\allowbreak Purchases 1} & 
{0.112\\\includegraphics[width=\linewidth]{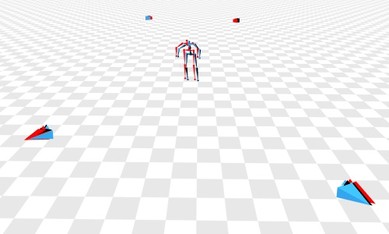}} & 
{0.182\\\includegraphics[width=\linewidth]{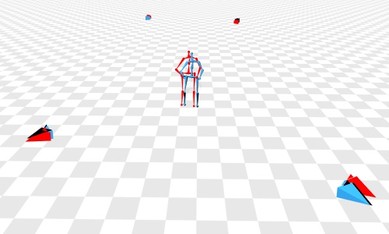}} & 
{7.179\\\includegraphics[width=\linewidth]{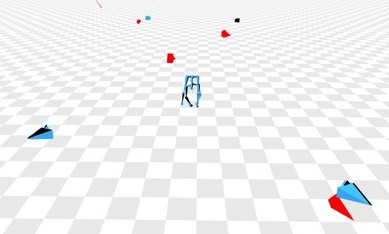}} & 
0.358$\pm$0.667 & 
{0.019\\\includegraphics[width=\linewidth]{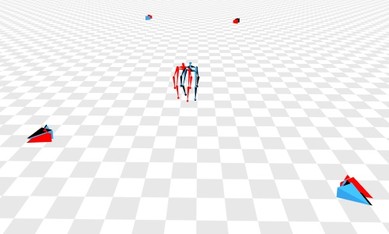}} & 
{0.025\\\includegraphics[width=\linewidth]{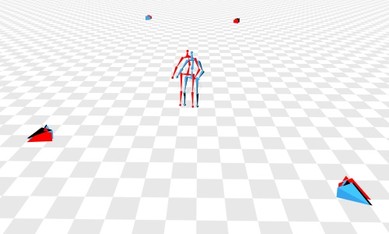}} & 
{0.034\\\includegraphics[width=\linewidth]{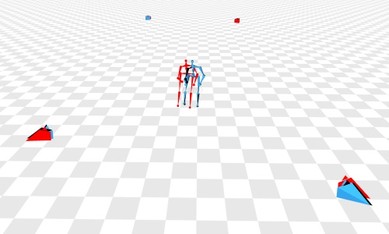}} & 
0.025$\pm$0.003 \\
\hline
\small{S9\_\allowbreak Purchases} & 
{0.135\\\includegraphics[width=\linewidth]{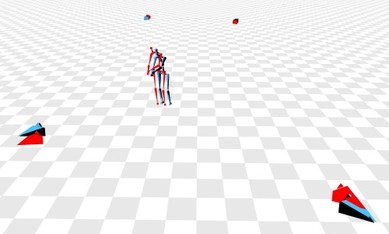}} & 
{0.305\\\includegraphics[width=\linewidth]{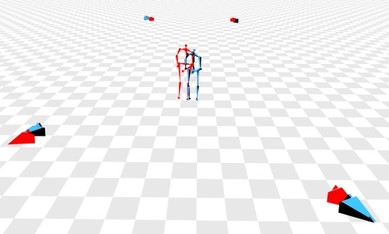}} & 
{12.024\\\includegraphics[width=\linewidth]{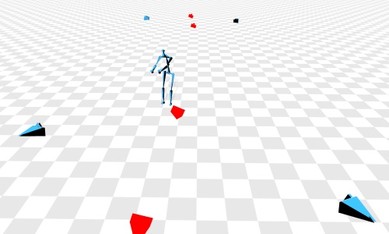}} & 
0.377$\pm$0.729 & 
{0.034\\\includegraphics[width=\linewidth]{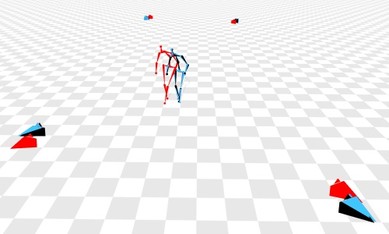}} & 
{0.040\\\includegraphics[width=\linewidth]{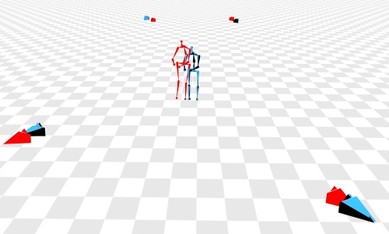}} & 
{0.093\\\includegraphics[width=\linewidth]{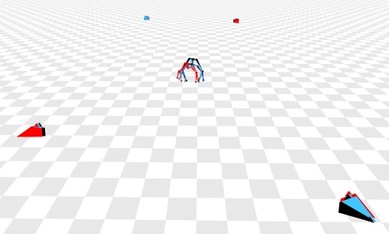}} & 
0.046$\pm$0.014 \\
\hline
\small{S9\_\allowbreak Sitting 1} & 
{0.119\\\includegraphics[width=\linewidth]{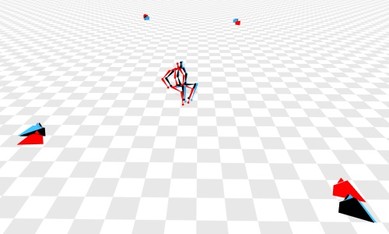}} & 
{0.234\\\includegraphics[width=\linewidth]{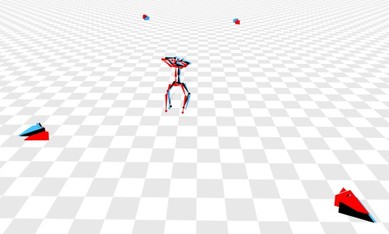}} & 
{0.564\\\includegraphics[width=\linewidth]{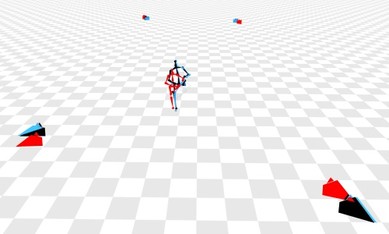}} & 
0.245$\pm$0.067 & 
{0.071\\\includegraphics[width=\linewidth]{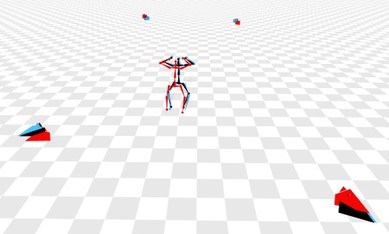}} & 
{0.084\\\includegraphics[width=\linewidth]{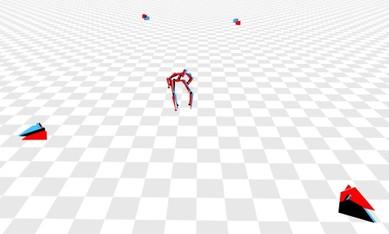}} & 
{0.102\\\includegraphics[width=\linewidth]{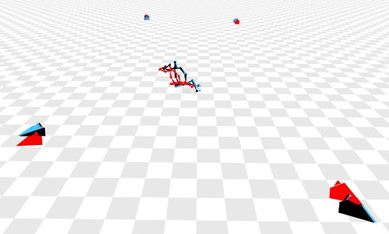}} & 
0.085$\pm$0.005 \\
\hline
\small{S11\_\allowbreak Sitting 1} & 
{0.071\\\includegraphics[width=\linewidth]{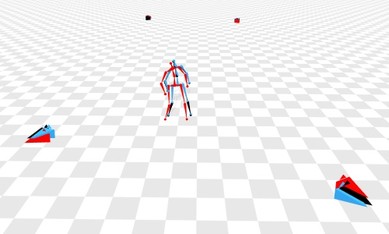}} & 
{0.198\\\includegraphics[width=\linewidth]{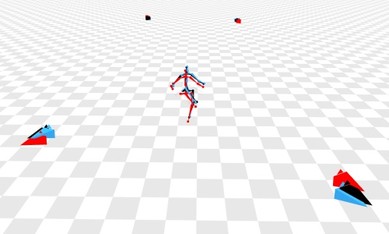}} & 
{3.758\\\includegraphics[width=\linewidth]{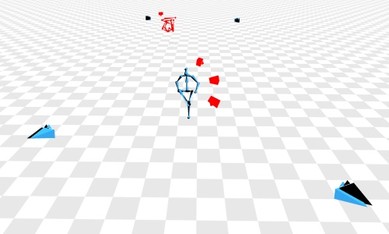}} & 
0.256$\pm$0.256 & 
{0.031\\\includegraphics[width=\linewidth]{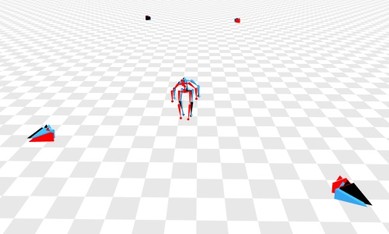}} & 
{0.037\\\includegraphics[width=\linewidth]{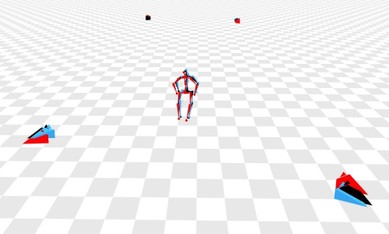}} & 
{0.051\\\includegraphics[width=\linewidth]{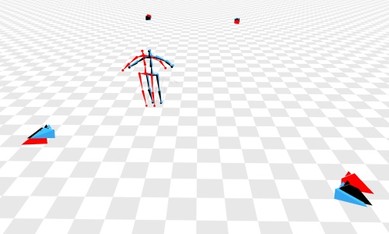}} & 
0.038$\pm$0.003 \\
\hline
\small{S9\_\allowbreak Sitting} & 
{0.128\\\includegraphics[width=\linewidth]{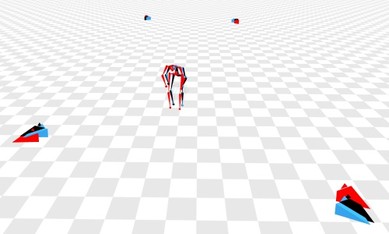}} & 
{0.241\\\includegraphics[width=\linewidth]{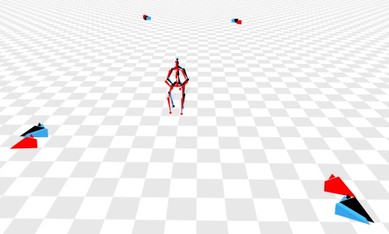}} & 
{0.496\\\includegraphics[width=\linewidth]{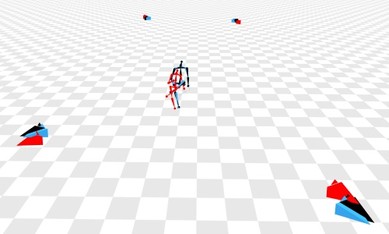}} & 
0.248$\pm$0.064 & 
{0.048\\\includegraphics[width=\linewidth]{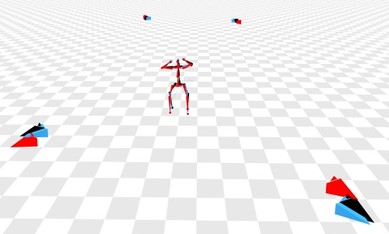}} & 
{0.057\\\includegraphics[width=\linewidth]{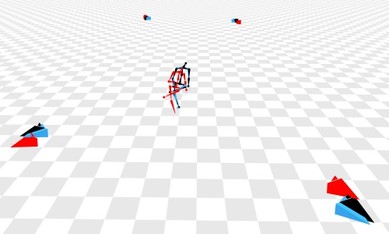}} & 
{0.074\\\includegraphics[width=\linewidth]{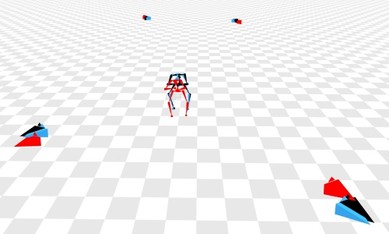}} & 
0.058$\pm$0.004 \\
\hline
\small{S11\_\allowbreak Sitting} & 
{0.068\\\includegraphics[width=\linewidth]{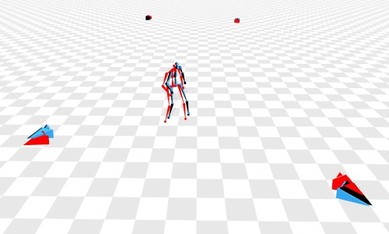}} & 
{0.243\\\includegraphics[width=\linewidth]{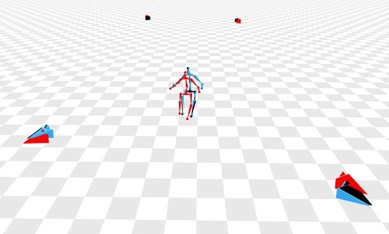}} & 
{0.626\\\includegraphics[width=\linewidth]{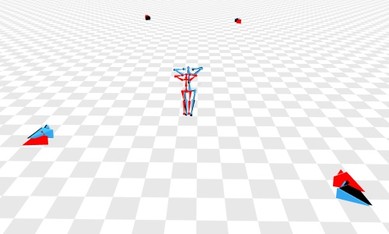}} & 
0.262$\pm$0.134 & 
{0.018\\\includegraphics[width=\linewidth]{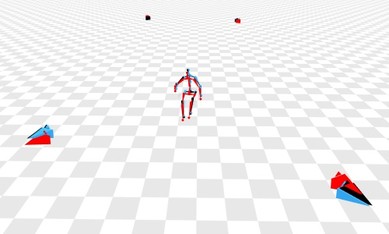}} & 
{0.025\\\includegraphics[width=\linewidth]{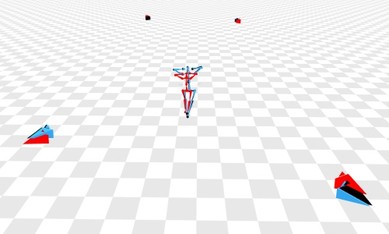}} & 
{0.045\\\includegraphics[width=\linewidth]{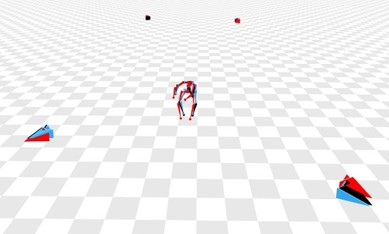}} & 
0.025$\pm$0.004 \\
\hline
\small{S11\_\allowbreak SittingDown} & 
{0.087\\\includegraphics[width=\linewidth]{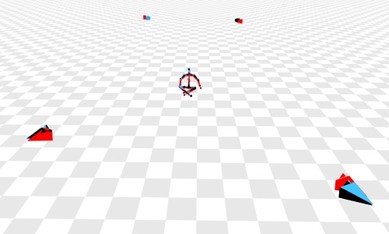}} & 
{0.831\\\includegraphics[width=\linewidth]{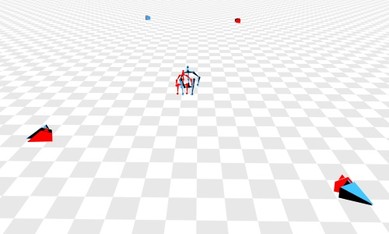}} & 
{11.474\\\includegraphics[width=\linewidth]{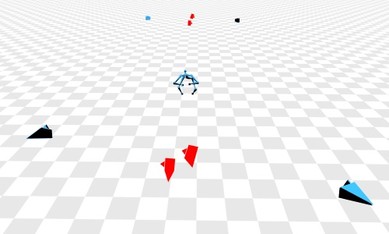}} & 
1.477$\pm$2.312 & 
{0.025\\\includegraphics[width=\linewidth]{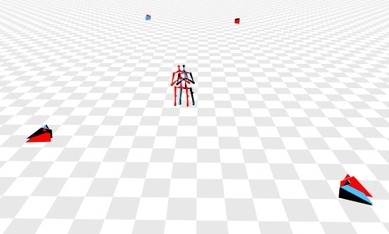}} & 
{0.032\\\includegraphics[width=\linewidth]{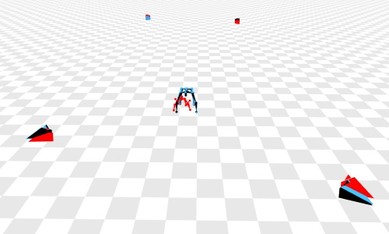}} & 
{0.044\\\includegraphics[width=\linewidth]{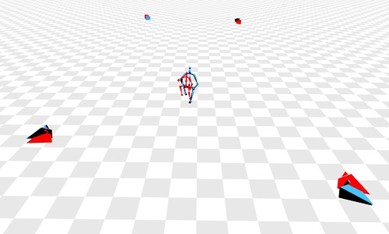}} & 
0.032$\pm$0.004 \\
\hline
\small{S11\_\allowbreak SittingDown 1} & 
{0.098\\\includegraphics[width=\linewidth]{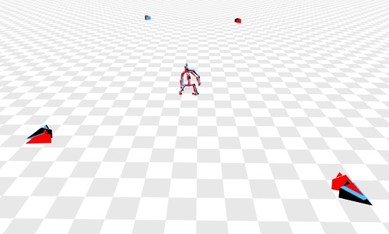}} & 
{0.820\\\includegraphics[width=\linewidth]{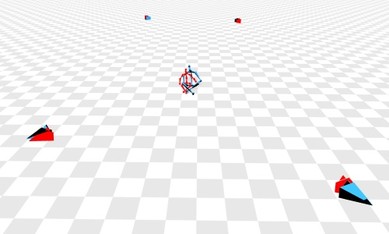}} & 
{9.471\\\includegraphics[width=\linewidth]{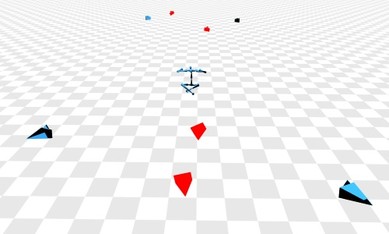}} & 
0.949$\pm$0.942 & 
{0.031\\\includegraphics[width=\linewidth]{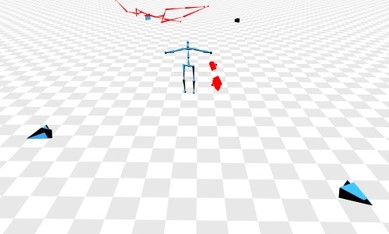}} & 
{0.041\\\includegraphics[width=\linewidth]{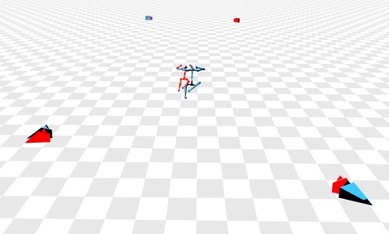}} & 
{0.053\\\includegraphics[width=\linewidth]{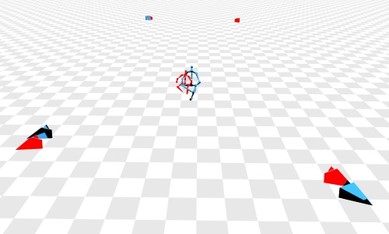}} & 
0.041$\pm$0.004 \\
\hline
\small{S9\_\allowbreak Smoking 1} & 
{0.072\\\includegraphics[width=\linewidth]{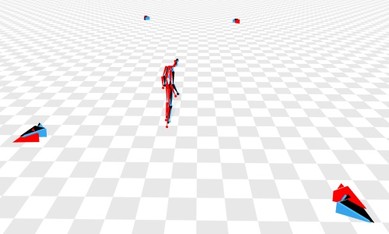}} & 
{0.189\\\includegraphics[width=\linewidth]{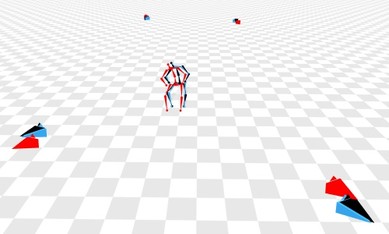}} & 
{12.239\\\includegraphics[width=\linewidth]{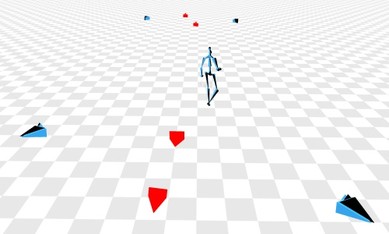}} & 
0.211$\pm$0.421 & 
{0.028\\\includegraphics[width=\linewidth]{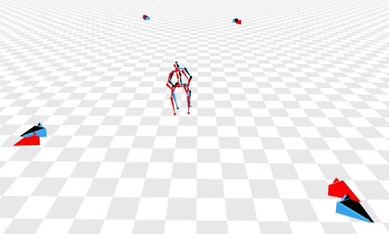}} & 
{0.041\\\includegraphics[width=\linewidth]{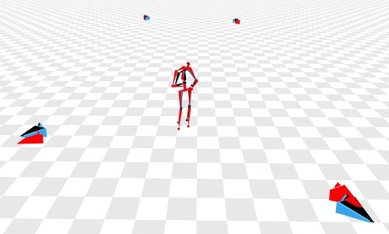}} & 
{0.064\\\includegraphics[width=\linewidth]{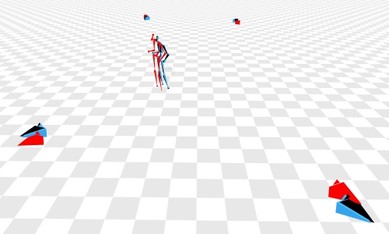}} & 
0.042$\pm$0.008 \\
\hline
\small{S11\_\allowbreak Smoking 2} & 
{0.068\\\includegraphics[width=\linewidth]{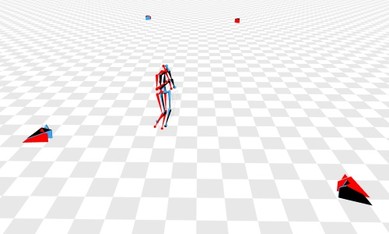}} & 
{0.169\\\includegraphics[width=\linewidth]{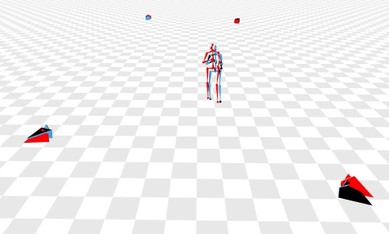}} & 
{4.724\\\includegraphics[width=\linewidth]{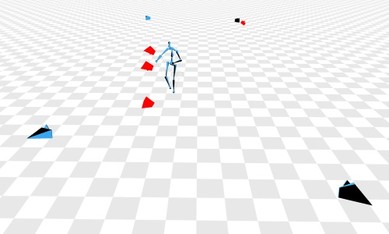}} & 
0.262$\pm$0.425 & 
{0.012\\\includegraphics[width=\linewidth]{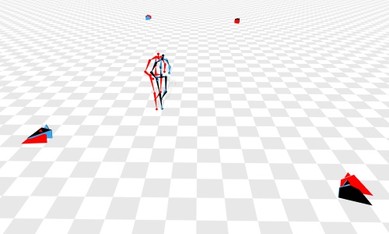}} & 
{0.019\\\includegraphics[width=\linewidth]{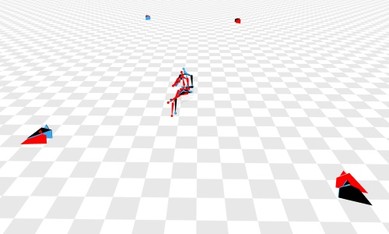}} & 
{0.034\\\includegraphics[width=\linewidth]{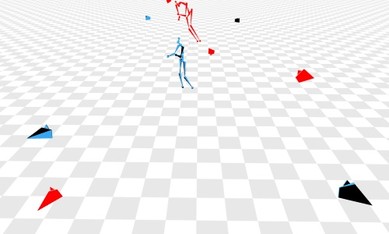}} & 
0.019$\pm$0.003 \\
\hline
\small{S9\_\allowbreak Smoking} & 
{0.071\\\includegraphics[width=\linewidth]{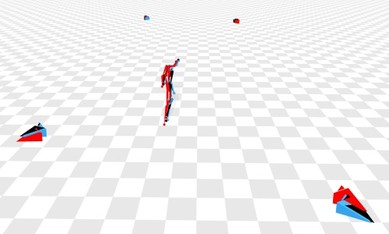}} & 
{0.176\\\includegraphics[width=\linewidth]{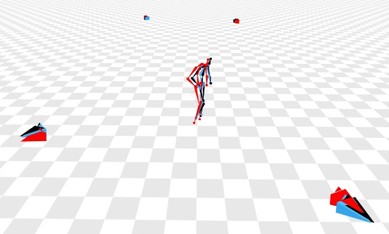}} & 
{2.860\\\includegraphics[width=\linewidth]{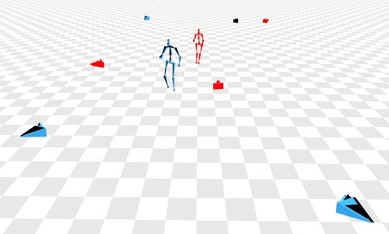}} & 
0.180$\pm$0.110 & 
{0.024\\\includegraphics[width=\linewidth]{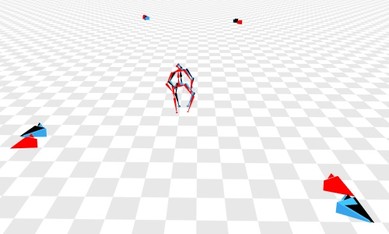}} & 
{0.034\\\includegraphics[width=\linewidth]{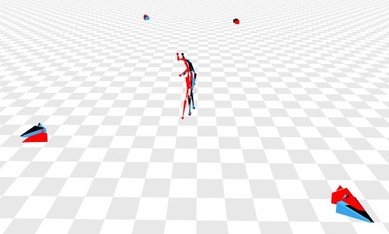}} & 
{0.059\\\includegraphics[width=\linewidth]{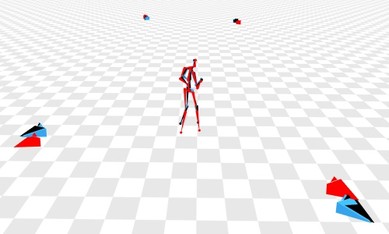}} & 
0.035$\pm$0.006 \\
\hline
\small{S11\_\allowbreak Smoking} & 
{0.057\\\includegraphics[width=\linewidth]{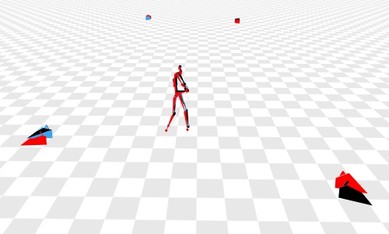}} & 
{0.144\\\includegraphics[width=\linewidth]{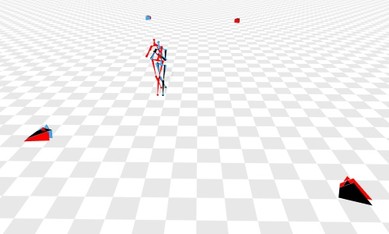}} & 
{4.310\\\includegraphics[width=\linewidth]{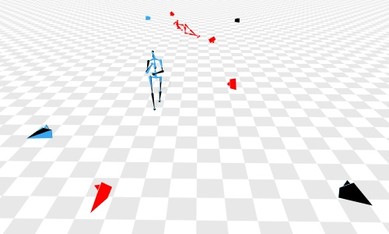}} & 
0.228$\pm$0.451 & 
{0.012\\\includegraphics[width=\linewidth]{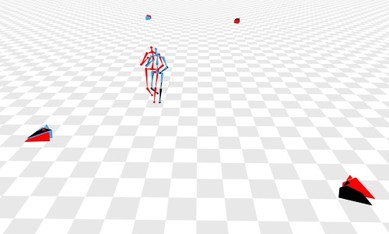}} & 
{0.018\\\includegraphics[width=\linewidth]{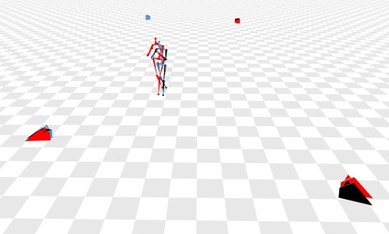}} & 
{0.030\\\includegraphics[width=\linewidth]{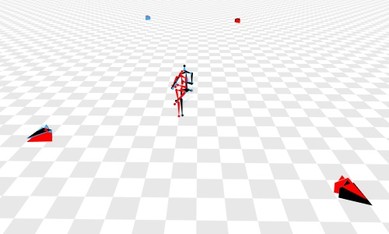}} & 
0.019$\pm$0.003 \\
\hline
\small{S9\_\allowbreak Photo 1} & 
{0.094\\\includegraphics[width=\linewidth]{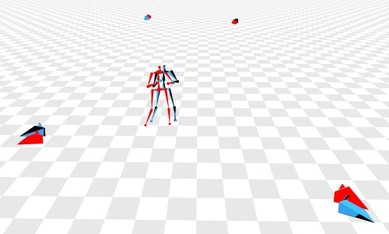}} & 
{0.248\\\includegraphics[width=\linewidth]{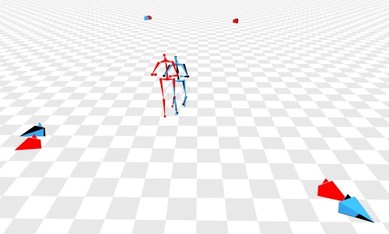}} & 
{11.388\\\includegraphics[width=\linewidth]{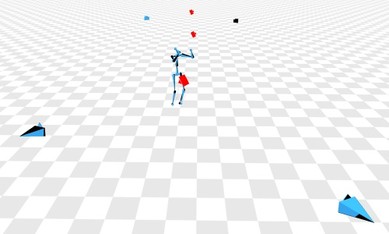}} & 
0.313$\pm$0.730 & 
{0.027\\\includegraphics[width=\linewidth]{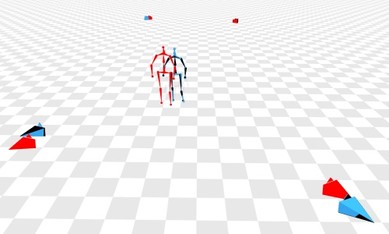}} & 
{0.037\\\includegraphics[width=\linewidth]{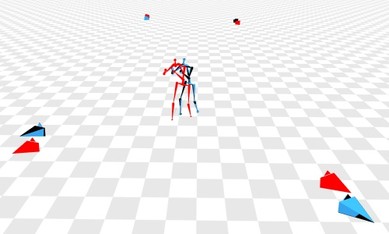}} & 
{0.055\\\includegraphics[width=\linewidth]{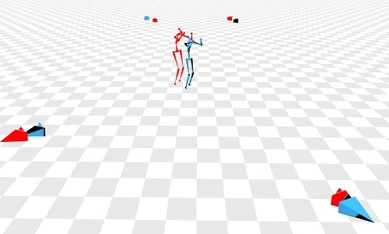}} & 
0.039$\pm$0.008 \\
\hline
\small{S11\_\allowbreak Photo 1} & 
{0.055\\\includegraphics[width=\linewidth]{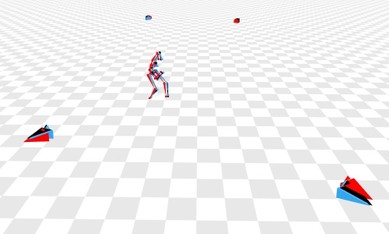}} & 
{0.178\\\includegraphics[width=\linewidth]{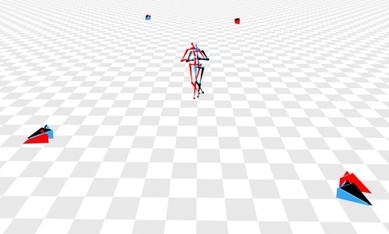}} & 
{13.303\\\includegraphics[width=\linewidth]{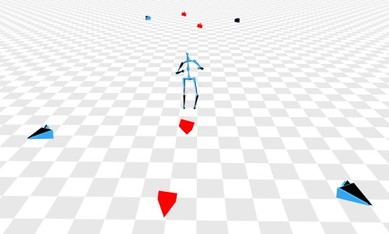}} & 
0.579$\pm$1.532 & 
{0.014\\\includegraphics[width=\linewidth]{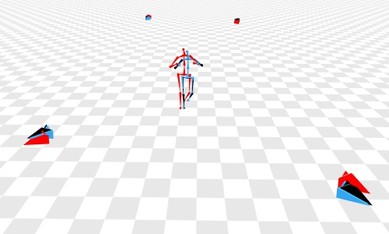}} & 
{0.023\\\includegraphics[width=\linewidth]{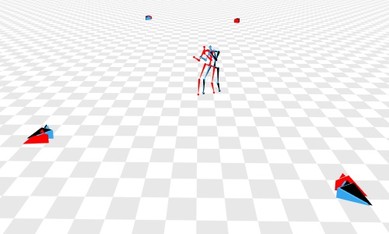}} & 
{0.032\\\includegraphics[width=\linewidth]{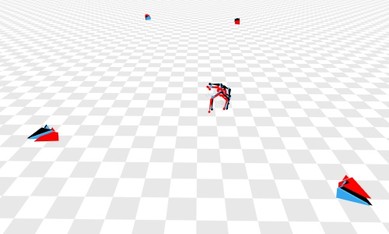}} & 
0.023$\pm$0.004 \\
\hline
\small{S9\_\allowbreak Photo} & 
{0.099\\\includegraphics[width=\linewidth]{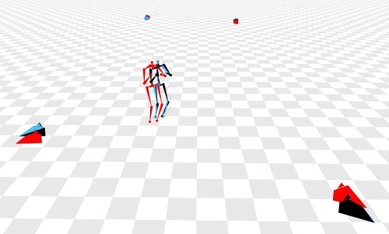}} & 
{0.253\\\includegraphics[width=\linewidth]{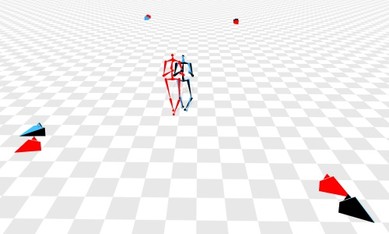}} & 
{13.131\\\includegraphics[width=\linewidth]{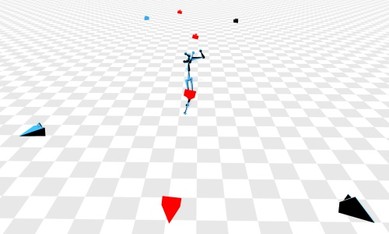}} & 
0.585$\pm$1.762 & 
{0.032\\\includegraphics[width=\linewidth]{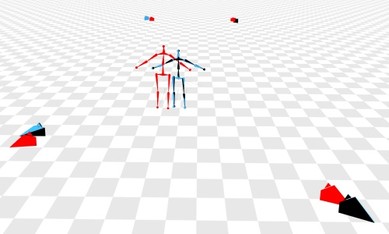}} & 
{0.040\\\includegraphics[width=\linewidth]{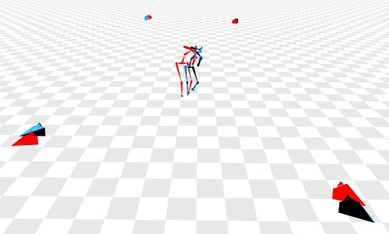}} & 
{0.057\\\includegraphics[width=\linewidth]{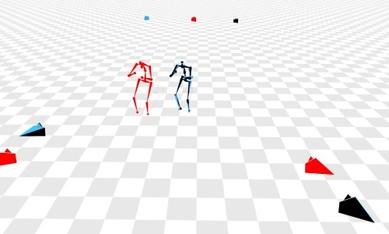}} & 
0.041$\pm$0.005 \\
\hline
\small{S11\_\allowbreak Photo} & 
{0.069\\\includegraphics[width=\linewidth]{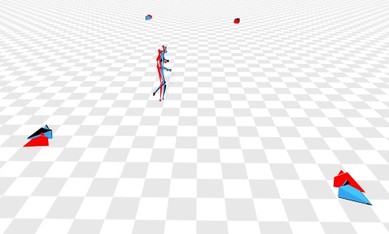}} & 
{0.181\\\includegraphics[width=\linewidth]{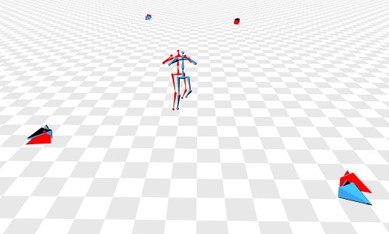}} & 
{14.154\\\includegraphics[width=\linewidth]{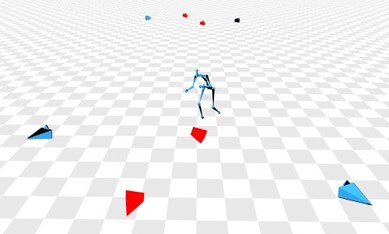}} & 
0.751$\pm$2.131 & 
{0.020\\\includegraphics[width=\linewidth]{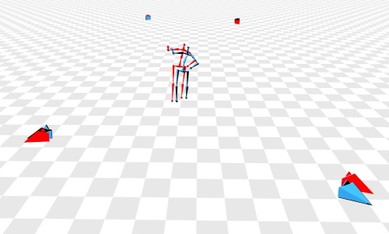}} & 
{0.027\\\includegraphics[width=\linewidth]{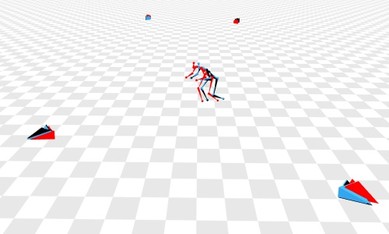}} & 
{0.036\\\includegraphics[width=\linewidth]{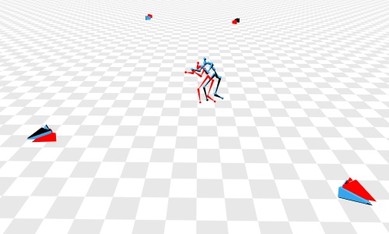}} & 
0.027$\pm$0.003 \\
\hline
\small{S11\_\allowbreak Waiting 1} & 
{0.110\\\includegraphics[width=\linewidth]{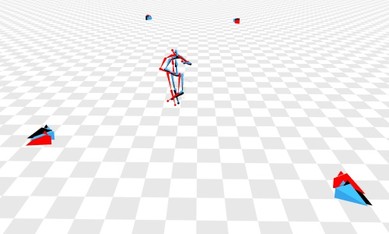}} & 
{0.171\\\includegraphics[width=\linewidth]{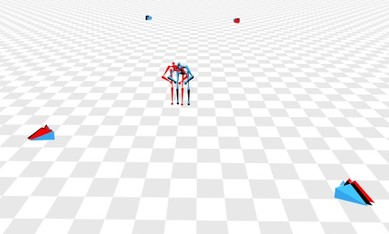}} & 
{6.420\\\includegraphics[width=\linewidth]{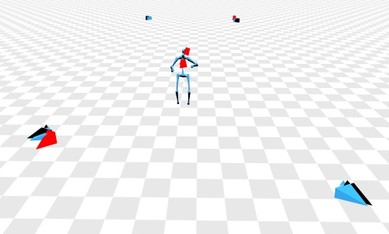}} & 
0.287$\pm$0.637 & 
{0.044\\\includegraphics[width=\linewidth]{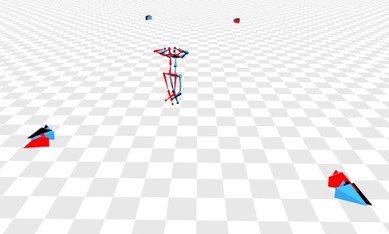}} & 
{0.048\\\includegraphics[width=\linewidth]{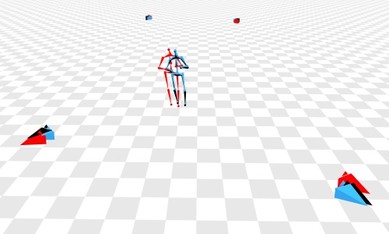}} & 
{0.061\\\includegraphics[width=\linewidth]{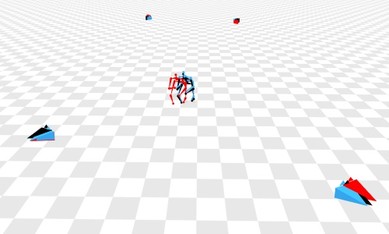}} & 
0.049$\pm$0.002 \\
\hline
\small{S11\_\allowbreak Waiting} & 
{0.104\\\includegraphics[width=\linewidth]{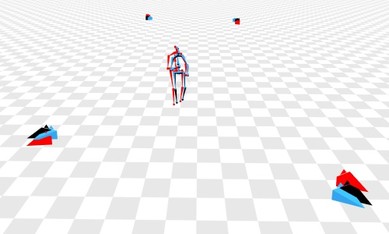}} & 
{0.158\\\includegraphics[width=\linewidth]{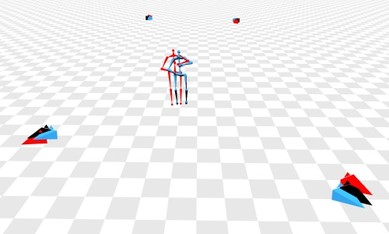}} & 
{11.165\\\includegraphics[width=\linewidth]{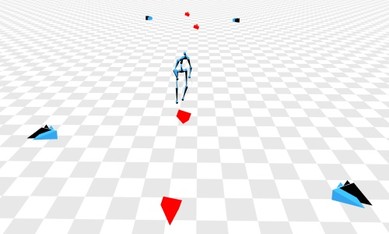}} & 
0.416$\pm$1.457 & 
{0.037\\\includegraphics[width=\linewidth]{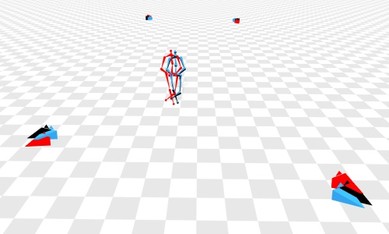}} & 
{0.041\\\includegraphics[width=\linewidth]{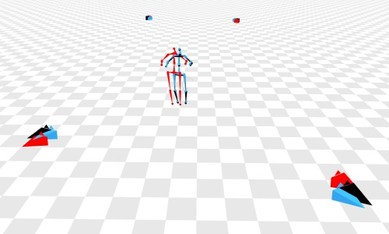}} & 
{0.051\\\includegraphics[width=\linewidth]{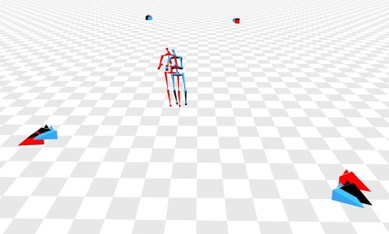}} & 
0.042$\pm$0.002 \\
\hline
\small{S9\_\allowbreak Walking 1} & 
{0.065\\\includegraphics[width=\linewidth]{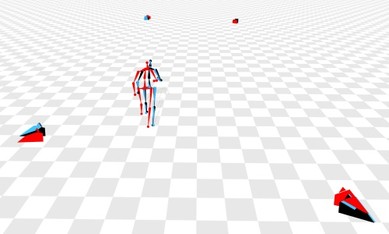}} & 
{0.185\\\includegraphics[width=\linewidth]{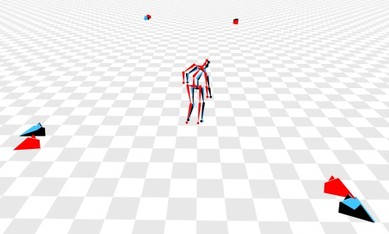}} & 
{10.694\\\includegraphics[width=\linewidth]{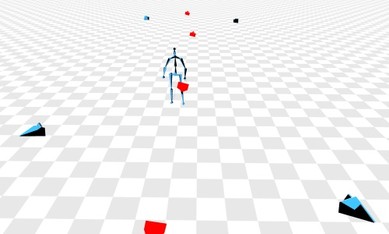}} & 
0.249$\pm$0.570 & 
{0.033\\\includegraphics[width=\linewidth]{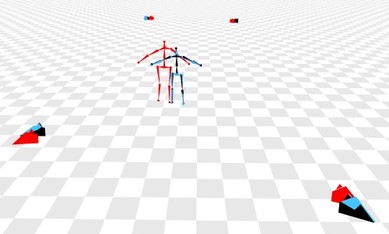}} & 
{0.054\\\includegraphics[width=\linewidth]{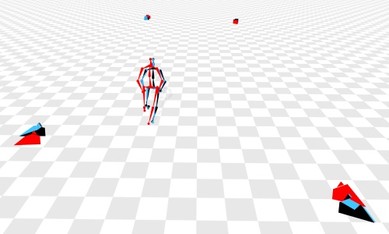}} & 
{0.087\\\includegraphics[width=\linewidth]{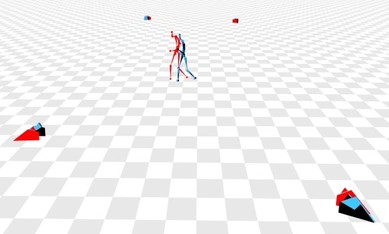}} & 
0.055$\pm$0.012 \\
\hline
\small{S9\_\allowbreak Walking} & 
{0.068\\\includegraphics[width=\linewidth]{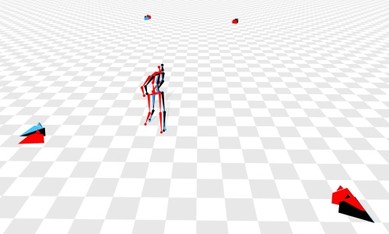}} & 
{0.193\\\includegraphics[width=\linewidth]{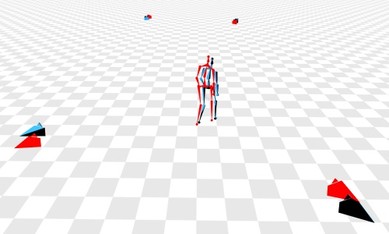}} & 
{14.010\\\includegraphics[width=\linewidth]{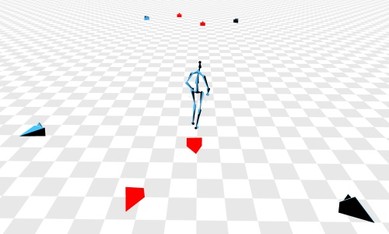}} & 
0.415$\pm$1.505 & 
{0.033\\\includegraphics[width=\linewidth]{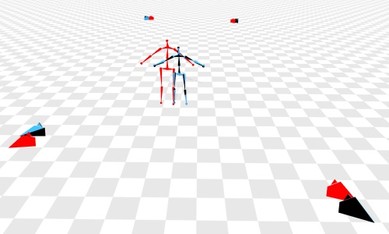}} & 
{0.053\\\includegraphics[width=\linewidth]{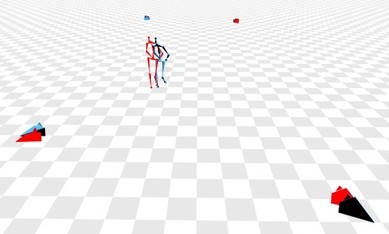}} & 
{0.078\\\includegraphics[width=\linewidth]{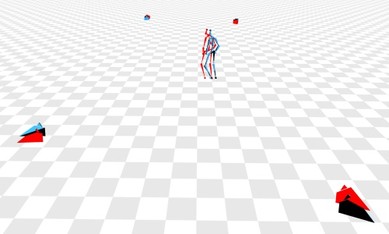}} & 
0.054$\pm$0.011 \\
\hline
\small{S9\_\allowbreak WalkDog 1} & 
{0.065\\\includegraphics[width=\linewidth]{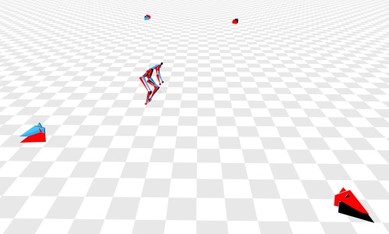}} & 
{0.193\\\includegraphics[width=\linewidth]{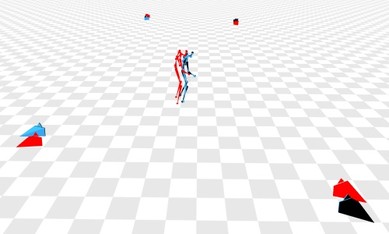}} & 
{12.397\\\includegraphics[width=\linewidth]{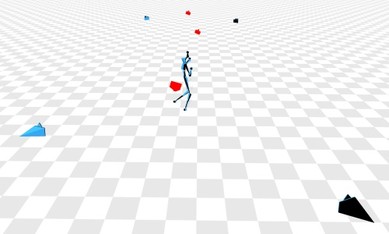}} & 
0.314$\pm$0.947 & 
{0.030\\\includegraphics[width=\linewidth]{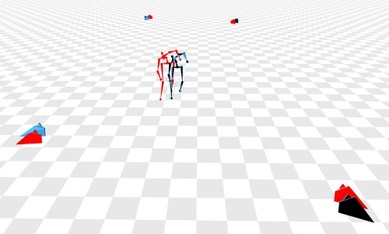}} & 
{0.047\\\includegraphics[width=\linewidth]{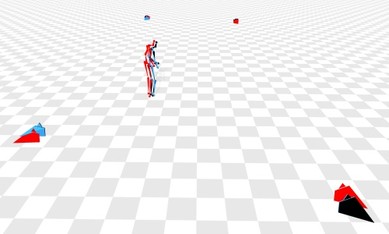}} & 
{0.072\\\includegraphics[width=\linewidth]{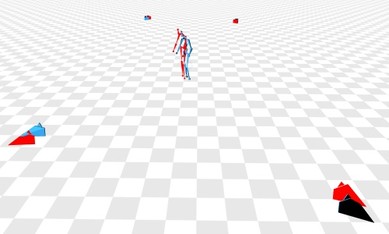}} & 
0.047$\pm$0.007 \\
\hline
\small{S11\_\allowbreak WalkDog 1} & 
{0.044\\\includegraphics[width=\linewidth]{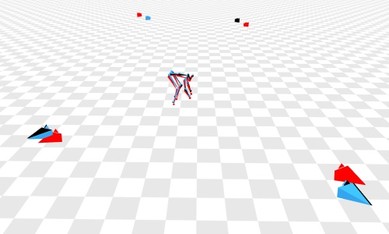}} & 
{0.199\\\includegraphics[width=\linewidth]{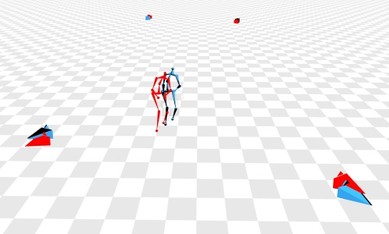}} & 
{11.408\\\includegraphics[width=\linewidth]{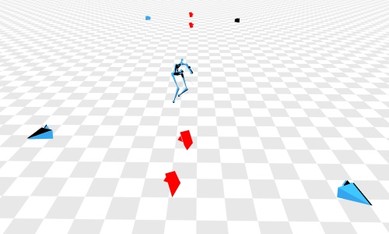}} & 
1.610$\pm$2.684 & 
{0.020\\\includegraphics[width=\linewidth]{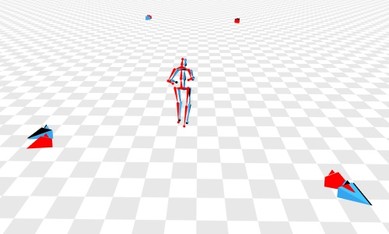}} & 
{0.027\\\includegraphics[width=\linewidth]{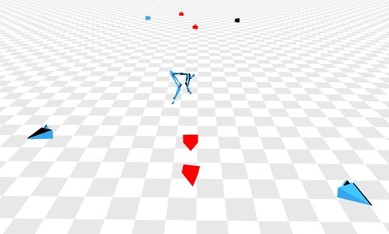}} & 
{0.041\\\includegraphics[width=\linewidth]{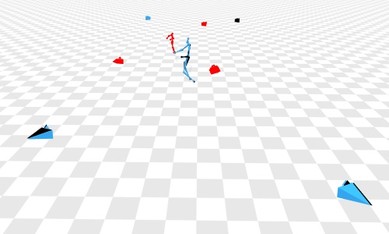}} & 
0.028$\pm$0.004 \\
\hline
\small{S9\_\allowbreak WalkDog} & 
{0.059\\\includegraphics[width=\linewidth]{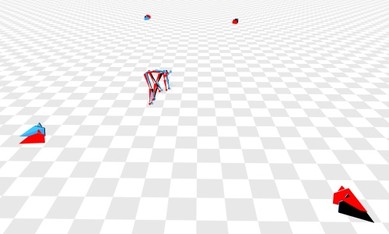}} & 
{0.237\\\includegraphics[width=\linewidth]{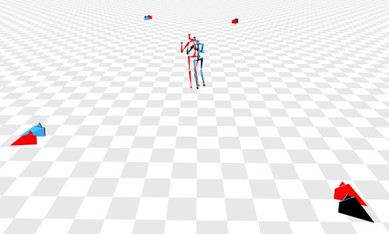}} & 
{11.765\\\includegraphics[width=\linewidth]{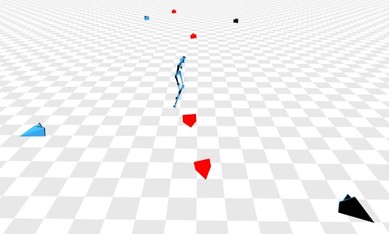}} & 
0.390$\pm$0.954 & 
{0.031\\\includegraphics[width=\linewidth]{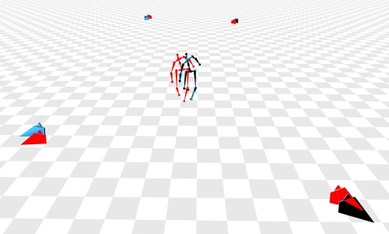}} & 
{0.048\\\includegraphics[width=\linewidth]{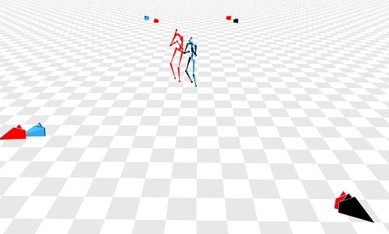}} & 
{0.070\\\includegraphics[width=\linewidth]{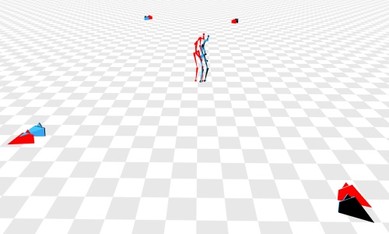}} & 
0.048$\pm$0.008 \\
\hline
\small{S9\_\allowbreak WalkTogether 1} & 
{0.068\\\includegraphics[width=\linewidth]{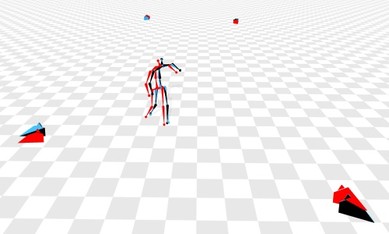}} & 
{0.203\\\includegraphics[width=\linewidth]{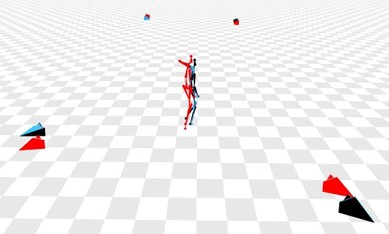}} & 
{4.962\\\includegraphics[width=\linewidth]{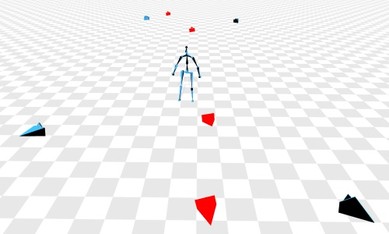}} & 
0.267$\pm$0.508 & 
{0.033\\\includegraphics[width=\linewidth]{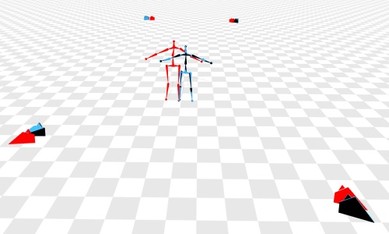}} & 
{0.052\\\includegraphics[width=\linewidth]{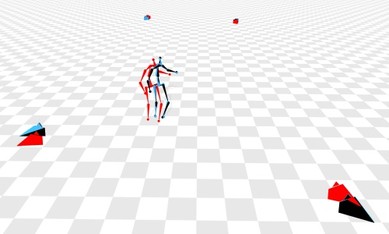}} & 
{0.082\\\includegraphics[width=\linewidth]{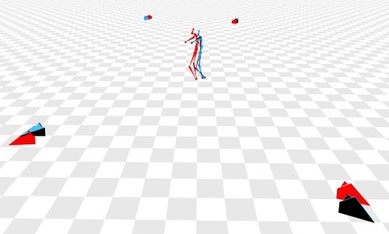}} & 
0.053$\pm$0.012 \\
\hline
\small{S11\_\allowbreak WalkTogether 1} & 
{0.062\\\includegraphics[width=\linewidth]{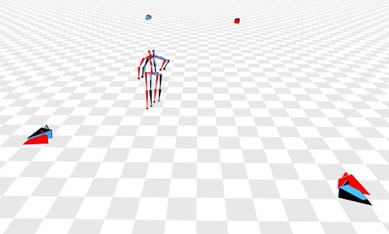}} & 
{0.141\\\includegraphics[width=\linewidth]{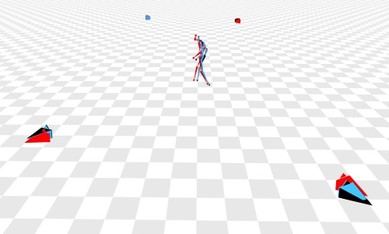}} & 
{13.837\\\includegraphics[width=\linewidth]{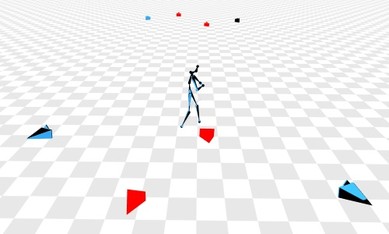}} & 
0.386$\pm$1.263 & 
{0.011\\\includegraphics[width=\linewidth]{figs/min_max_median/S11_WalkTogether_1_kineo_min.jpg}} & 
{0.015\\\includegraphics[width=\linewidth]{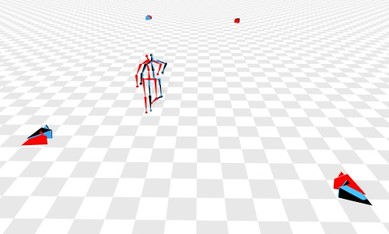}} & 
{0.025\\\includegraphics[width=\linewidth]{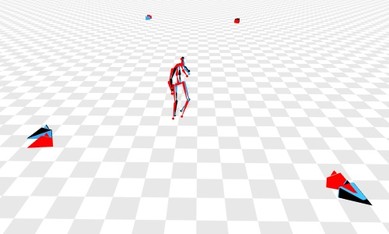}} & 
0.016$\pm$0.003 \\
\hline
\small{S11\_\allowbreak WalkTogether} & 
{0.054\\\includegraphics[width=\linewidth]{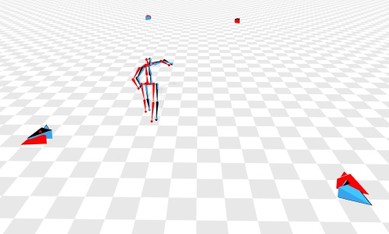}} & 
{0.136\\\includegraphics[width=\linewidth]{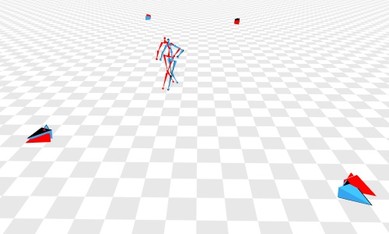}} & 
{12.240\\\includegraphics[width=\linewidth]{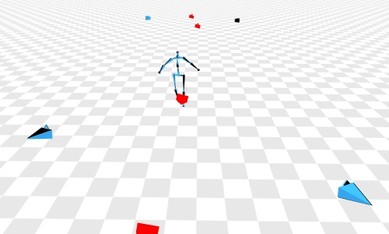}} & 
0.231$\pm$0.817 & 
{0.012\\\includegraphics[width=\linewidth]{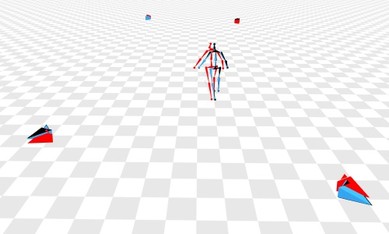}} & 
{0.016\\\includegraphics[width=\linewidth]{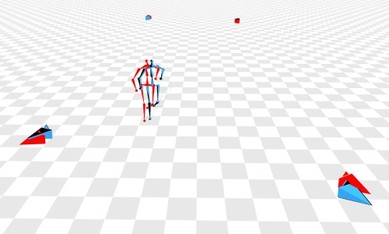}} & 
{0.027\\\includegraphics[width=\linewidth]{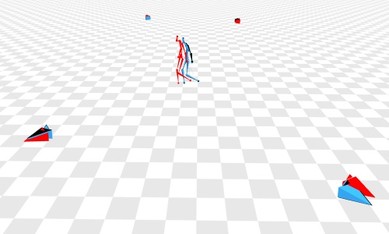}} & 
0.017$\pm$0.003 \\

\end{longtblr}

\section{Additional Files}
\label{supp-sec:additional_files}

Additional files are provided alongside this document, including videos and interactive visualization files presenting reconstructed 3D sequences.

\subsection{Video Files}
\begin{figure}[tbp]
    \centering
    \begin{subfigure}[b]{0.30\linewidth}
        \centering
        \includegraphics[width=\linewidth,height=80pt,keepaspectratio]{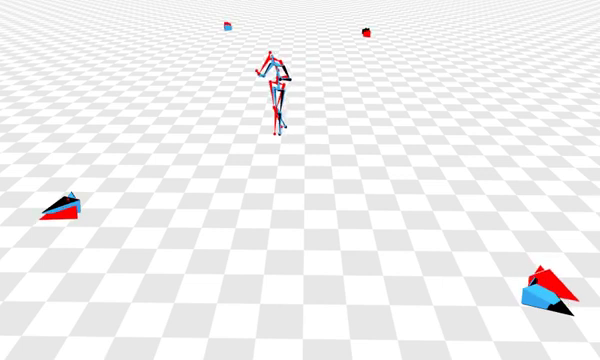}
        \caption{}
        \label{supp-fig:video_hsfm_kineo}
    \end{subfigure}
    \hfill
    \begin{subfigure}[b]{0.30\linewidth}
        \centering
        \includegraphics[width=\linewidth,height=80pt,keepaspectratio]{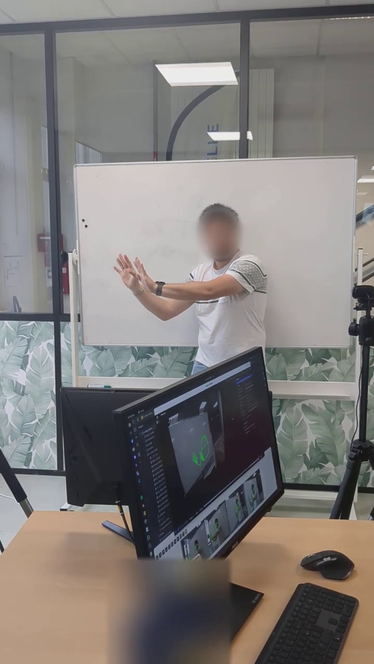}
        \caption{}
        \label{supp-fig:video_realtime}
    \end{subfigure}
    \hfill
    \begin{subfigure}[b]{0.30\linewidth}
        \centering
        \includegraphics[width=\linewidth,height=80pt,keepaspectratio]{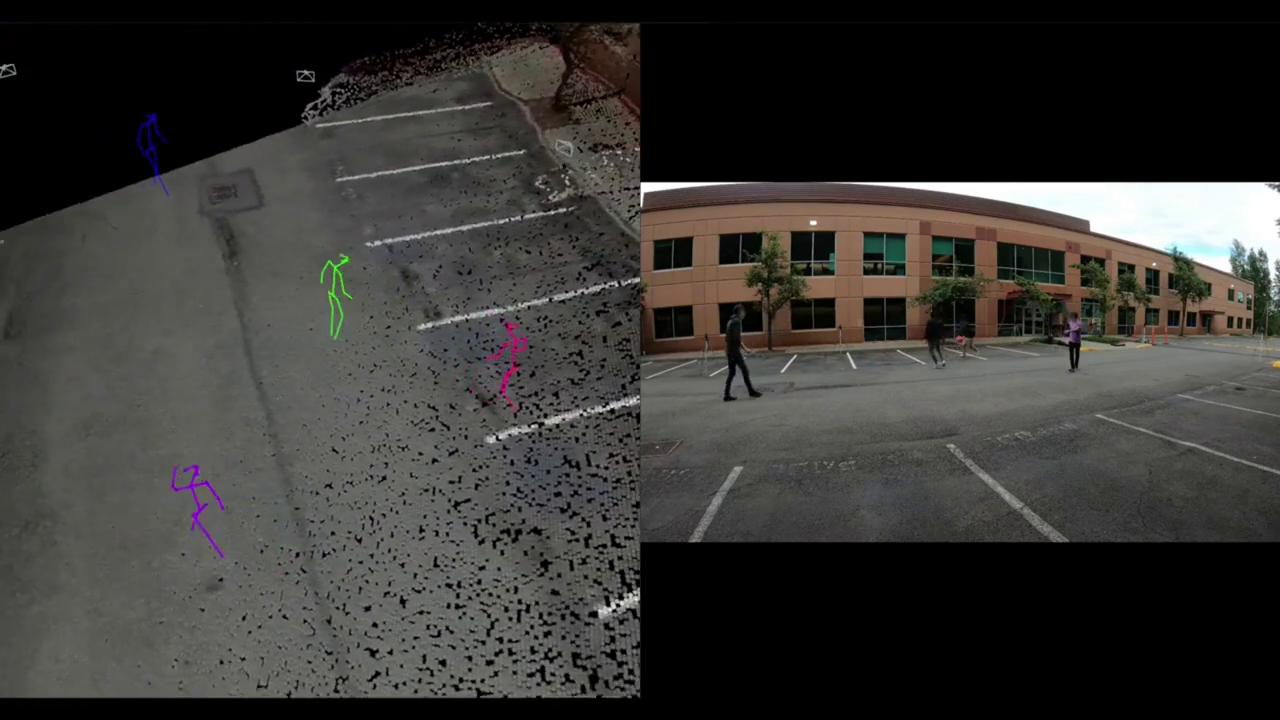}
        \caption{}
        \label{supp-fig:video_basketball}
    \end{subfigure}

    \vspace{2mm} %

    \begin{subfigure}[b]{0.30\linewidth}
        \centering
        \includegraphics[width=\linewidth,height=80pt,keepaspectratio]{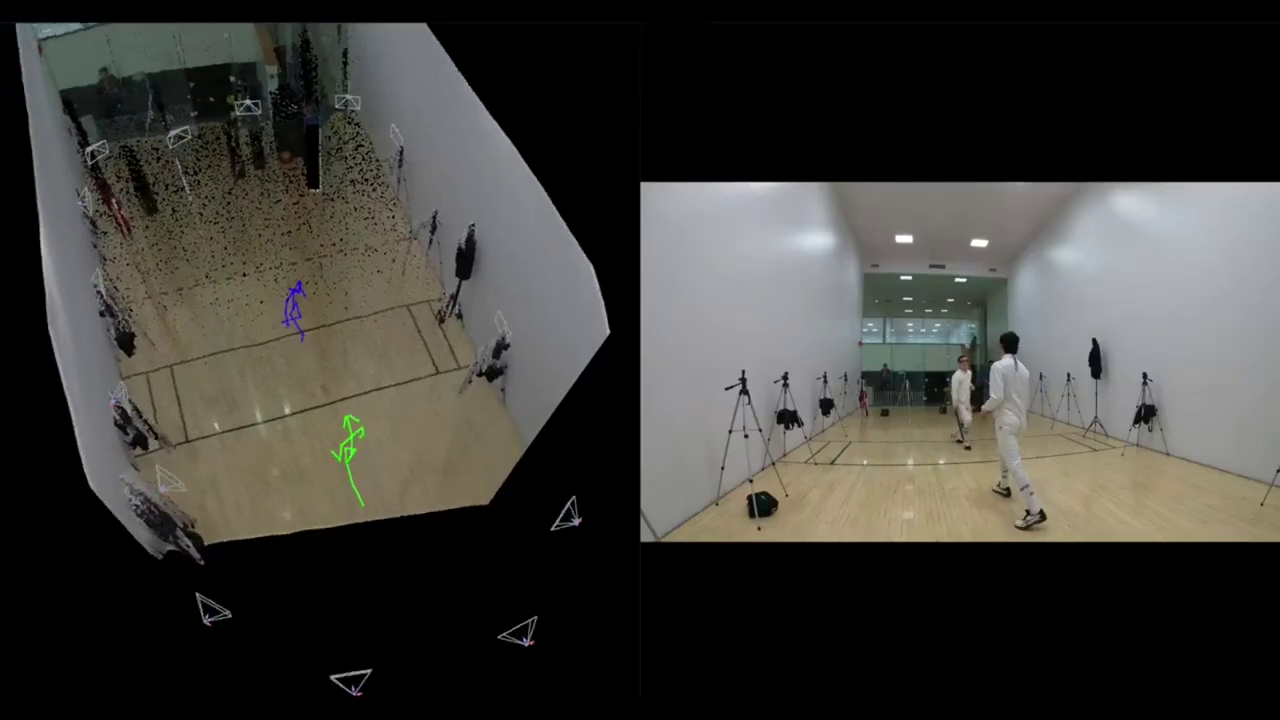}
        \caption{}
        \label{supp-fig:video_fencing_coco}
    \end{subfigure}
    \hfill
    \begin{subfigure}[b]{0.30\linewidth}
        \centering
        \includegraphics[width=\linewidth,height=80pt,keepaspectratio]{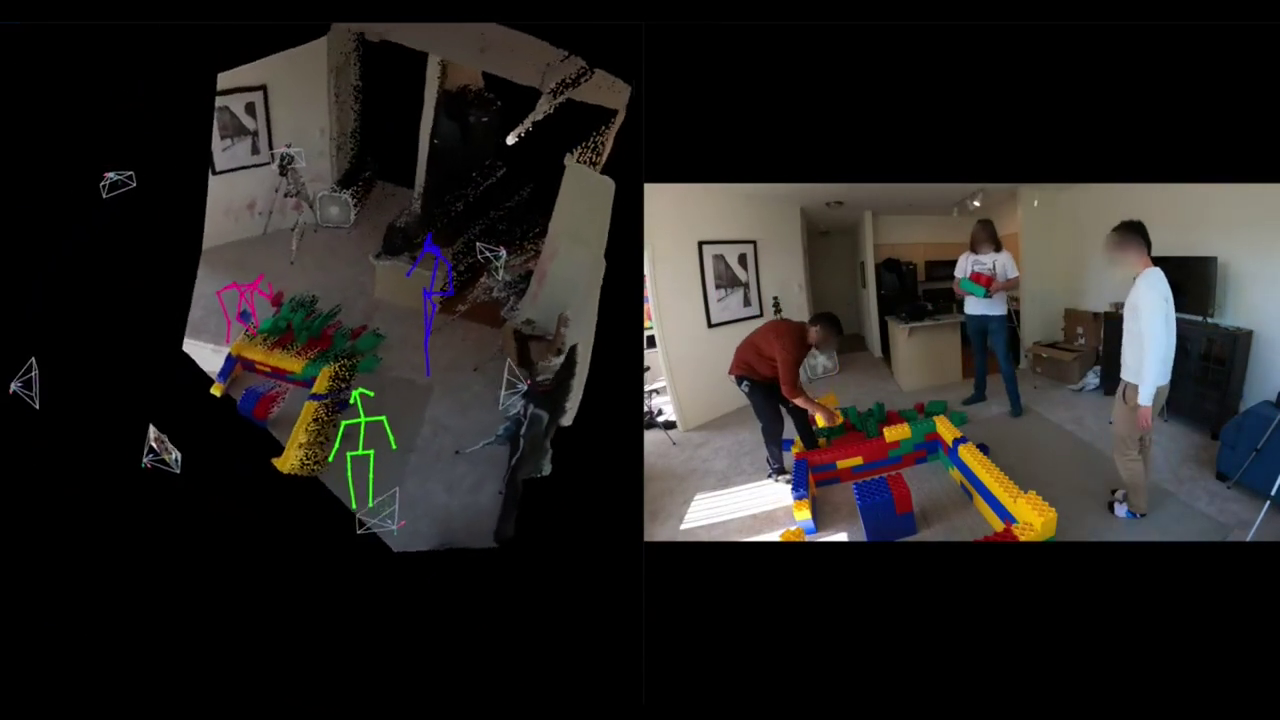}
        \caption{}
        \label{supp-fig:video_legoassemble}
    \end{subfigure}
    \hfill
    \begin{subfigure}[b]{0.30\linewidth}
        \centering
        \includegraphics[width=\linewidth,height=80pt,keepaspectratio]{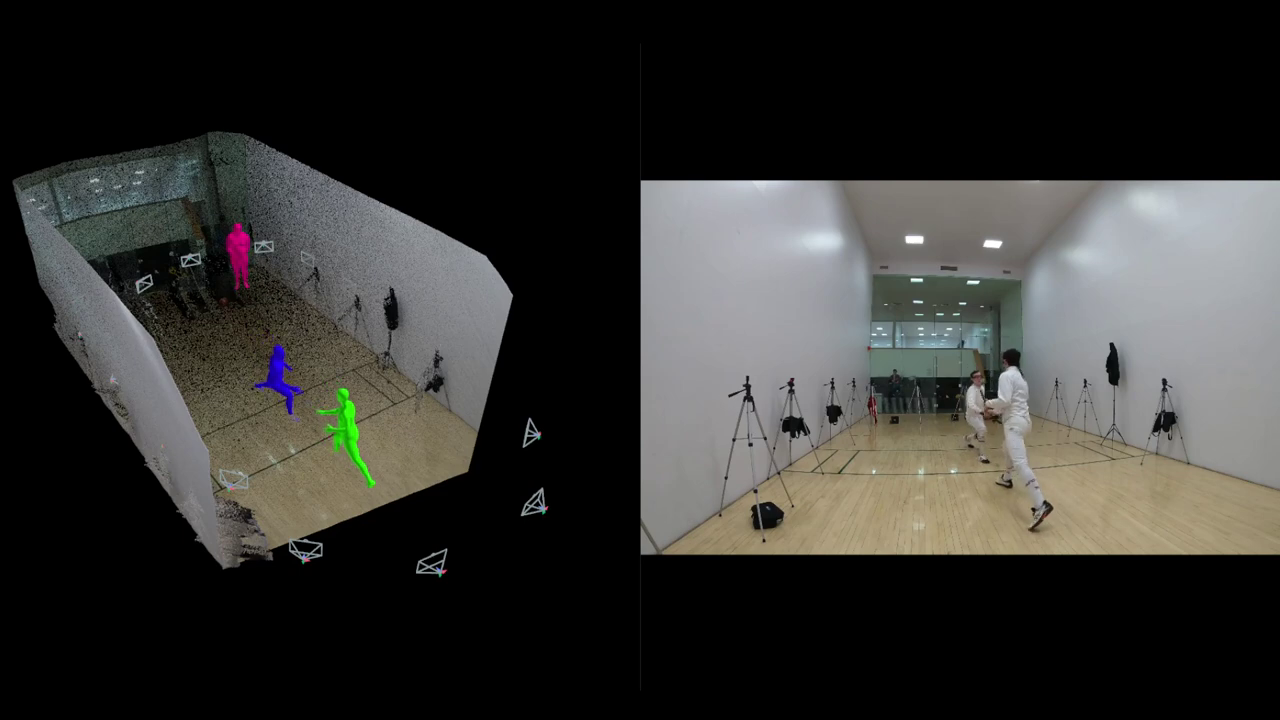}
        \caption{}
        \label{supp-fig:video_fencing_smpl}
    \end{subfigure}

    \vspace{2mm}

    \begin{subfigure}[b]{0.30\linewidth}
        \centering
        \includegraphics[width=\linewidth,height=80pt,keepaspectratio]{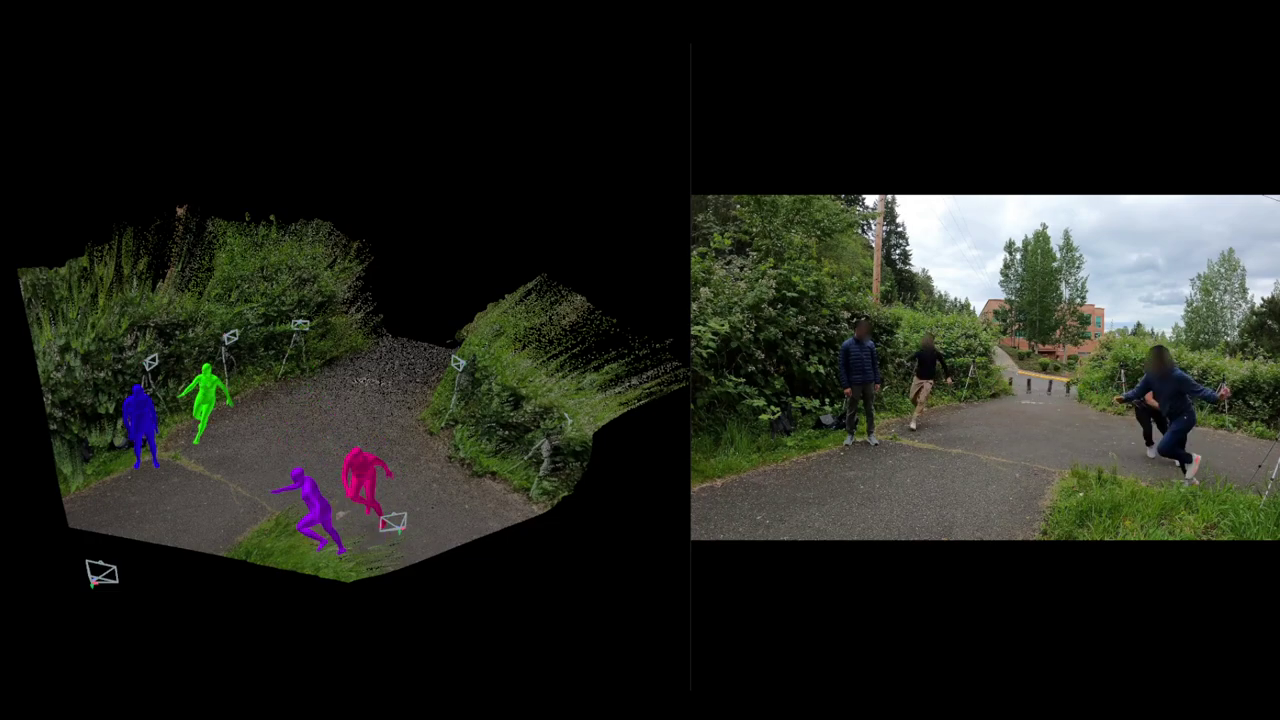}
        \caption{}
        \label{supp-fig:video_tagging}
    \end{subfigure}
    \hfill
    \begin{subfigure}[b]{0.30\linewidth}
        \centering
        \includegraphics[width=\linewidth,height=80pt,keepaspectratio]{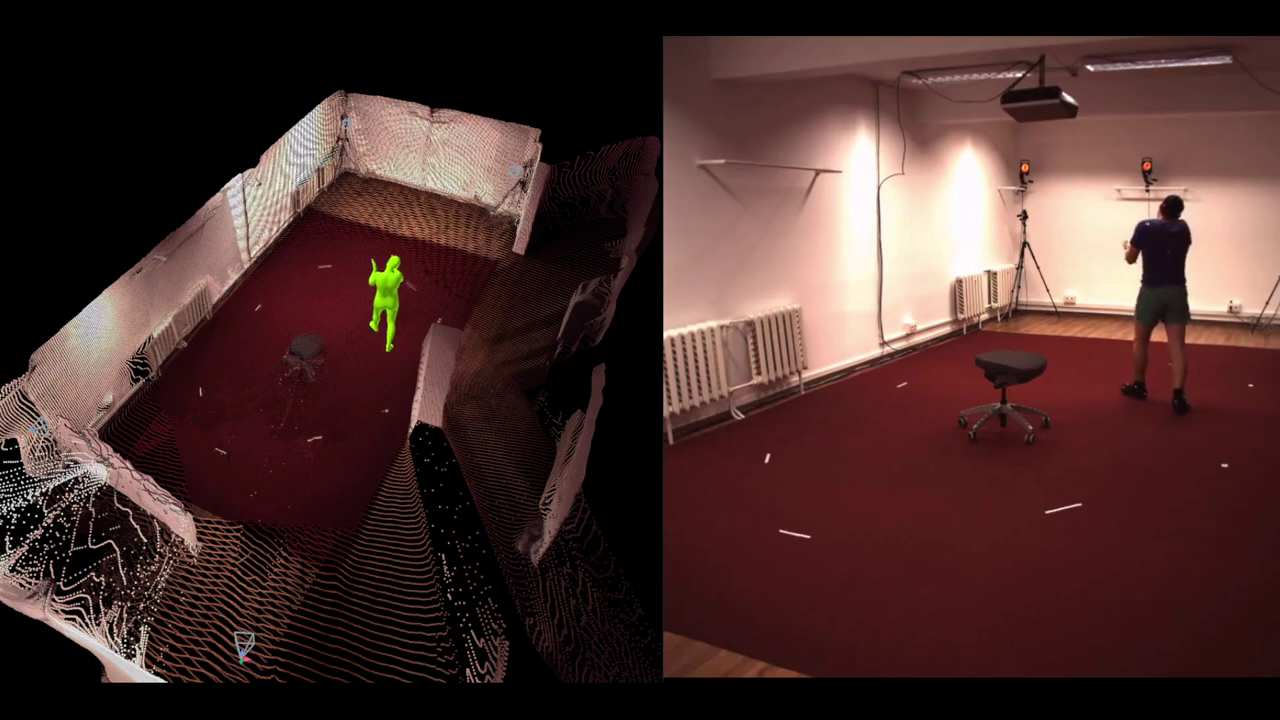}
        \caption{}
        \label{supp-fig:video_phoning}
    \end{subfigure}
    \hfill
    \begin{subfigure}[b]{0.30\linewidth}
        \centering
        \includegraphics[width=\linewidth,height=80pt,keepaspectratio]{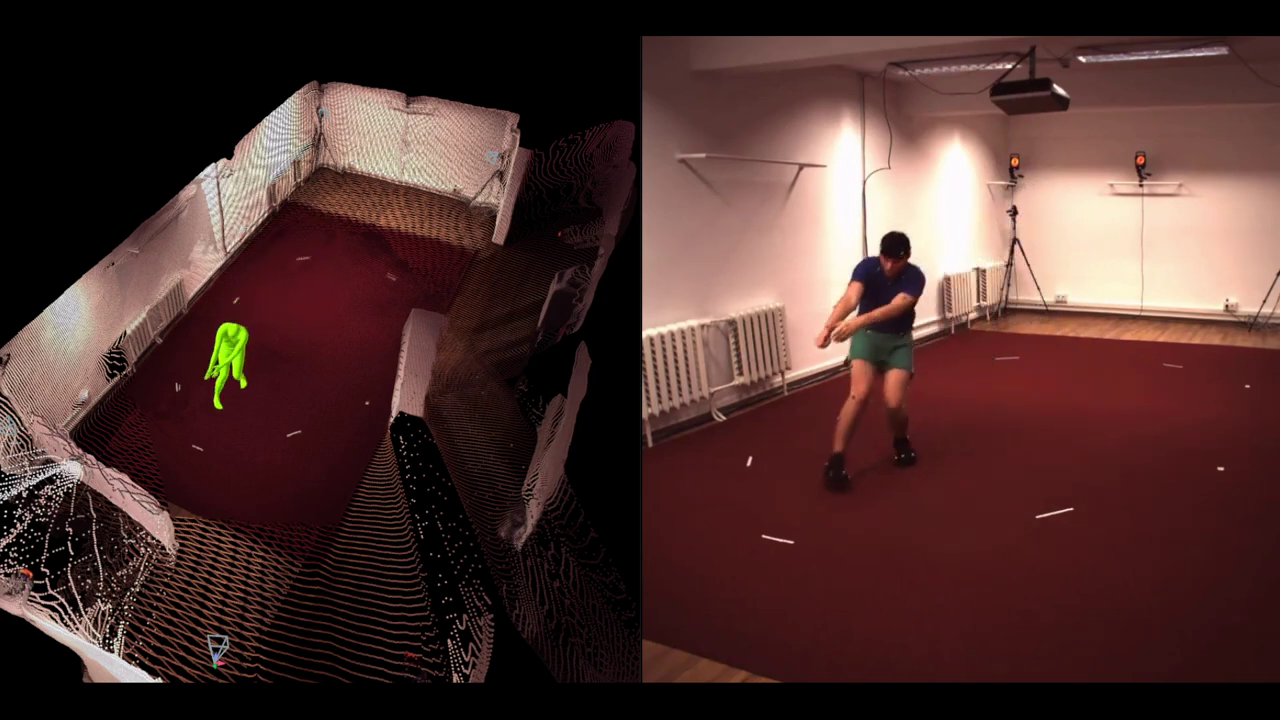}
        \caption{}
        \label{supp-fig:video_walkdog}
    \end{subfigure}

    \caption{Thumbnails of all supplementary video files provided alongside this document. Each thumbnail represents a single frame from the corresponding video; full videos are provided separately.}
    \label{supp-fig:video_thumbnails_all}
\end{figure}

\begin{itemize}
    \item \textbf{hsfm\_kineo\_comparison.mp4}: Qualitative comparison of Kineo and HSfM on a Human3.6M sequence selected as the one where HSfM achieves its best quantitative performance (lowest MPJPE). Kineo reconstructions are shown in blue, HSfM in red, and the ground truth in black, allowing direct visual assessment of reconstruction accuracy.

    \item \textbf{realtime\_use\_case.mp4}: Demonstrates the system operating in real time on live video input using 4 webcams and DWPose for keypoint detection. Human motion is reconstructed on-the-fly and visualized in 3D, highlighting Kineo's capability for interactive applications.
    
    \item \textbf{basketball\_nlf\_coco.mp4}: Presents the reconstructed basketball sequence from the EgoHumans dataset in the COCO output format using NLF.
    
    \item \textbf{fencing\_nlf\_coco.mp4}: Presents the reconstructed fencing sequence from the EgoHumans dataset in the COCO output format using NLF.

    \item \textbf{legoassemble\_nlf\_coco.mp4}: Presents the reconstructed lego assembly sequence from the EgoHumans dataset in the COCO output format using NLF.
    
    \item \textbf{fencing\_nlf\_smpl.mp4}: Presents the reconstructed fencing sequence from the EgoHumans dataset in the SMPL output format using NLF.

    \item \textbf{tagging\_nlf\_smpl.mp4}: Presents the reconstructed tagging sequence from the EgoHumans dataset in the SMPL output format using NLF.

    \item \textbf{phoning\_nlf\_smpl.mp4}: Presents the reconstructed phoning sequence from the Human3.6M dataset in the SMPL output format using NLF.
    
    \item \textbf{walkdog\_nlf\_smpl.mp4}: Presents the reconstructed walk dog sequence from the Human3.6M dataset in the SMPL output format using NLF.
\end{itemize}

\subsection{Interactive Visualization}

\begin{itemize}
    \item \textbf{S11\_Greeting\_h36m.rrd}: This file provides an interactive 3D visualization of the reconstructed greeting sequence from Human3.6M. Using the \texttt{rerun} viewer, users can manipulate the scene and overlay the ground truth (in purple) and the estimated skeleton (in green). The file can be opened directly from the online Rerun viewer available at: \url{https://rerun.io/viewer}.
\end{itemize}

\begin{figure}[H]
    \centering
    \begin{subfigure}[b]{0.49\linewidth}
        \centering
        \includegraphics[width=\linewidth]{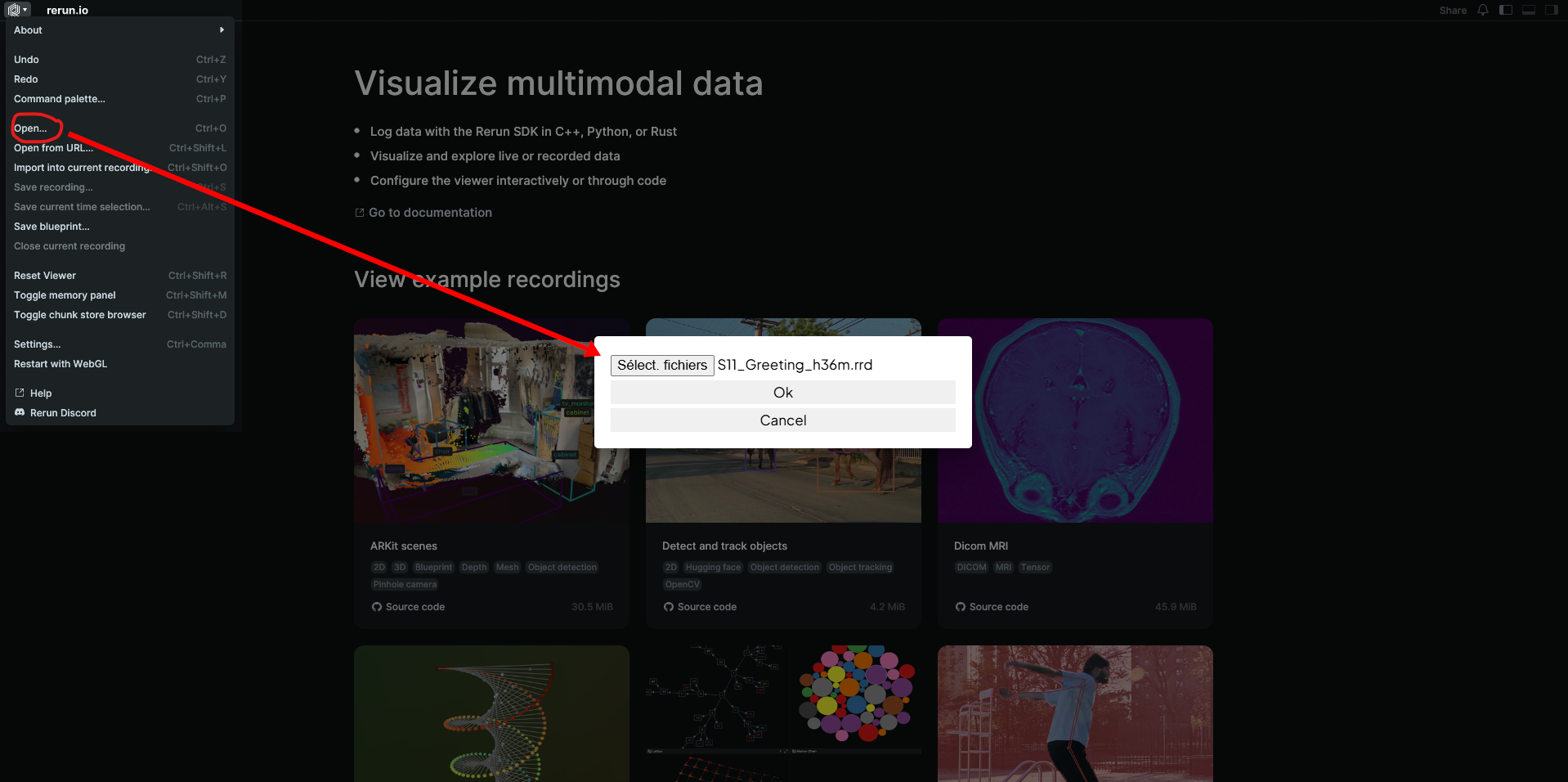}
        \caption{}
        \label{supp-fig:rerun_open_file}
    \end{subfigure}
    \hfill
    \begin{subfigure}[b]{0.49\linewidth}
        \centering
        \includegraphics[width=\linewidth]{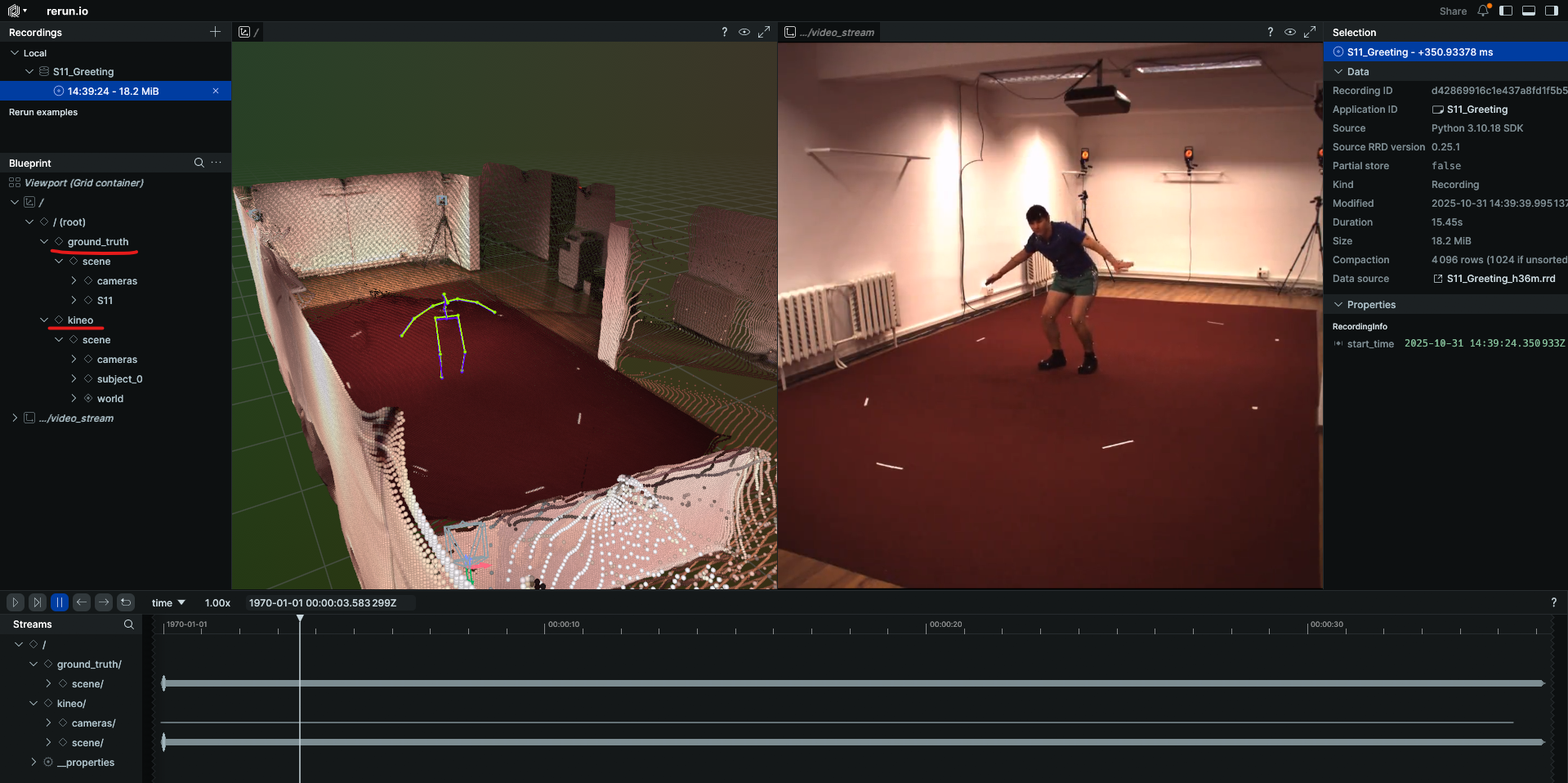}
        \caption{}
        \label{supp-fig:rerun_viz}
    \end{subfigure}
    \caption{Illustration of the procedure to visualize the interactive visualization file for the S11\_Greeting sequence from Human3.6M. (\textbf{a}) Opening the .rrd file using the online \texttt{rerun} viewer available at \url{https://rerun.io/viewer}. (\textbf{b}) Interactive 3D visualization, showing the scene hierarchy with ground-truth and Kineo estimates.}
\end{figure}

\end{document}